%% file: MAIN_iclr2025_conference.tex
\documentclass{article} % For LaTeX2e
\usepackage[main,final]{neurips_2026}
\usepackage{times}

\input{math_commands.tex}

\input{macros}

\usepackage{hyperref}
\usepackage{url}
\usepackage{booktabs}
\usepackage{colortbl}
\usepackage{subcaption}
\usepackage{soul}
\usepackage{array} 
\usepackage{wrapfig}
\usepackage{tabularx}
\usepackage{graphicx}
\usepackage{comment}
\usepackage{multirow}
\usepackage[table]{xcolor} % for row colors
\usepackage{amsmath}
\DeclareMathOperator{\sech}{sech}

\usepackage[skip=0.5\baselineskip]{caption}
\newcolumntype{Y}{>{\centering\arraybackslash}X}

\definecolor{tablegray}{gray}{0.92}
\definecolor{tablegray}{gray}{0.9}
\sethlcolor{tablegray}

\newcolumntype{L}{>{\raggedright\arraybackslash}X}
\newcolumntype{C}{>{\centering\arraybackslash}X}
\newcolumntype{R}{>{\raggedleft\arraybackslash}X}

\makeatletter
\newcommand{\listofappendices}{%
  \section*{Appendix contents}%
  \@starttoc{apx}%
}
\makeatother

\title{Effective Biological Representation Learning by Masking Gene Expression}

\author{\textbf{Kian Kenyon-Dean$^{1}$ \quad Alina Selega$^{1\star}$ \quad Ihab Bendidi$^{1,2,3\star\dagger}$  \quad Jordan M. Sorokin$^{1\dagger}$} \\
\textbf{Luca Bertinetto$^{1}$ \quad David Errington$^{1}$ \quad Hayley Donnella$^{1\dagger}$} \quad \textbf{Oren Kraus}$^{1\dagger}$ \\
$^1$Recursion \quad $^2$Valence Labs \quad $^3$\'{E}cole Normale Sup\'{e}rieure PSL\\
\texttt{kian.kd@recursion.com} \quad \texttt{info@rxrx.ai}
}

\begin{document}

\maketitle

\newcommand{\fix}{\marginpar{FIX}}
\newcommand{\new}{\marginpar{NEW}}
\newcommand{\mytilde}{\raise.17ex\hbox{$\scriptstyle\mathtt{\sim}$}}

% \maketitle

\begin{abstract}

\input{sections/abstract}
\end{abstract}

% Graphical abstract
% TODO: UPDATE FIGURE TO REPLACE WITH TXFM-B and NOT -L
\begin{figure}[h!]
    \centering % trim = lower right upper left
    % \includegraphics[angle=90,trim={4cm 1cm 3.5cm 3cm},clip,width=\linewidth]{imgs/fig1.pdf}
    %%%% left lower right upper
    % \includegraphics[width=\linewidth,trim={0.5cm 10cm 2cm 9cm},clip]{figures/TxFM Architecture.pdf}
    \includegraphics[width=\linewidth,trim={2cm 10cm 1.5cm 2.4cm},clip]{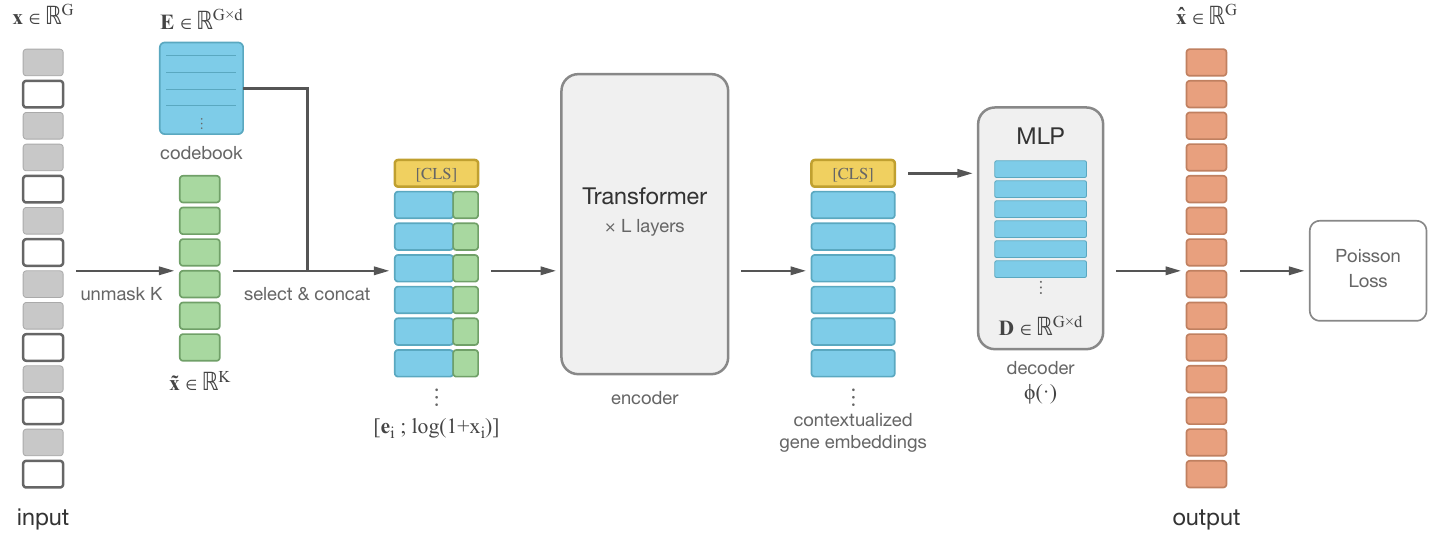}
    \caption{\textbf{Overview of our approach.} TxFM is a self-supervised learning (SSL) model trained for masked reconstruction of partially observed gene expression count data $\textbf{x}$, from bulk and single-cell RNA-seq transcriptomic assays. We evaluate inductive representation quality of genetically perturbed input cells (CLS token), and the functional relationships recalled by the transformer encoder's gene codebook $\textbf{E} \in \mathbb{R}^{G\times d}$, and the MLP decoder's gene reconstruction parameters $\textbf{D} \in \mathbb{R}^{G\times d}$.}%TxFM offers a $\geq$10\% overall relative improvement in zero-shot perturbational benchmarks versus the next-best alternative method.}
    \label{fig:graphical_abstract}
\end{figure}

\section{Introduction}

\input{sections/introduction}

\section{Related Work}
% \todo{REFACTOR THIS SECTION AS NEEDED}
\input{sections/related_work}

\section{Approach} \label{sec:approach}
% \todo{UPDATE THIS SECTION WITH THEORY SUBSECTION}
\input{sections/approach}

% \subsection{Ablation of the main properties of TxFM} \label{sec:ablations}
% \input{sections/ablations}

% \section{Experiments}
% \todo{REFACTOR THIS SECTION}
% \subsection{Models}
% \input{sections/experments}

\section{Evaluation}
\input{sections/results}

%\section{Ablation of the main properties of TxFM} %\label{sec:ablations}
%\input{sections/ablations}

\section{Conclusion} \label{sec:conclusion}
\input{sections/conclusion}

% \section*{Impact Statement}
% This paper advances the field of Machine Learning by improving deep representations of high-dimensional transcriptomic data, including genetic perturbations across diverse cellular contexts. By doing so, this work may accelerate drug discovery and the identification of therapeutic targets for diseases like cancer. While the primary impact is to provide a more efficient and interpretable tool for biological research, the authors recognize that any model used in a biomedical context must be rigorously validated before informing clinical decisions to ensure safety and equity across diverse patient populations.

\bibliography{iclr2025_conference}
\bibliographystyle{iclr2026_conference.bst}

\clearpage
\appendix

% \begingroup
%   \centering
%   {\Large\bfseries Appendix}   % main title text
%   \vspace{0.6cm}                          % gap between title and TOC
% \endgroup

\listofappendices

\let\oldsection\section
\let\oldsubsection\subsection

\let\origsection\section
\renewcommand{\section}[1]{%
  \origsection{#1}%
  \addcontentsline{apx}{section}{\protect\numberline{\thesection}#1}%
}
\let\origsubsection\subsection
\renewcommand{\subsection}[1]{%
  \origsubsection{#1}%
  \addcontentsline{apx}{subsection}{\protect\numberline{\thesubsection}#1}%
}

\input{sections/Z_appendix}

\newpage
\input{checklist.tex}

\end{document}

%% file: math_commands.tex
\usepackage{amsmath,amsfonts,bm}

\def\eqref#1{equation~\ref{#1}}
\def\1{\bm{1}}

\DeclareMathAlphabet{\mathsfit}{\encodingdefault}{\sfdefault}{m}{sl}
\SetMathAlphabet{\mathsfit}{bold}{\encodingdefault}{\sfdefault}{bx}{n}

%% file: macros.tex
\usepackage{xcolor}

%% file: sections/abstract.tex
RNA sequencing produces rich and diverse datasets of gene expression, offering compelling insights into cellular state and function that have many applications in drug discovery. Modeling such data is challenging due to inherent technical noise and experimental batch effects, as evidenced by many existing transcriptomic foundation models (FMs) underperforming relative to linear baselines. Such results raise the question of whether deep representation learning provides a distinct advantage over the direct use of raw transcript counts. Our work explores this by developing a new self-supervised model, TxFM, with a focus on inductive representation learning evaluations. TxFM employs a masked autoencoding approach tailored to diverse RNA-seq count data, and our ablation study empirically identifies crucial architecture configurations required for strong transfer performance. Additionally, we curate a public training corpus, DiverseRNA-1.4M, and find that TxFM trained on this curated dataset yields high-fidelity gene representations that outperform FMs trained on atlas-scale corpora over $100\times$ larger. Overall, our results indicate that inductive self-supervised learning is a viable modeling approach for transcriptomics representation, provided a careful synthesis of model architecture and training data curation.\footnote{$\star$: equal contribution. $\dagger$: work done while still employed at Recursion.}

%% file: sections/introduction.tex
Deep representation learning is now standard in vision and language, yet its application to modeling gene expression has encountered various difficulties. Several recent benchmarking studies find that transcriptomic foundation models (FMs) fail to outperform simple linear baselines such as PCA on held-out tasks \citep{szalata2024transformers, boiarsky2024deeper, bendidi2024benchmarking, AhlmannEltze2025,kedzierska2025zero,csendes2025benchmarking,wenkel2026txpert}. These results admit two readings: either gene expression counts already carry most recoverable biological signal in their raw form, or current pretraining approaches are inadequate for transfer learning to new data.

This work explores the latter hypothesis. We develop \textbf{TxFM}, a masked autoencoder (MAE) for transcriptomics, and systematically investigate the recipe needed to produce representations that can surpass count-based baselines. Our investigation is organized around three axes.

\textbf{Inductive representation learning.} We focus on \emph{inductive} SSL: the goal is to train a model whose representations transfer to cellular contexts unseen during training, rather than to precisely model the training data distribution. We benchmark various count-based baselines and 16 different FMs on three held-out genetic perturbation datasets featuring cell lines not seen during training (\S~\ref{sec:perts}). Furthermore, we evaluate the intrinsic gene representations learned directly within FM parameters as a byproduct of SSL (\S~\ref{sec:bmdb}), testing their ability to group functionally related genes (e.g., those encoding subunits of a protein complex). Unlike work leveraging pretrained protein embeddings \citep{lin2023evolutionary, Adduri2025STATE}, TxFM does not incorporate external biological priors, allowing us to isolate what can be learned strictly from expression data.

\textbf{Training data composition.} We construct \textbf{DiverseRNA-1.4M}, a curated public corpus of 1.4 million bulk and single-cell RNA-seq samples with broad genetic coverage (\autoref{tab:datasets}). We find that TxFM and baselines trained on this data produce higher-signal gene representations than models trained on atlas-scale datasets over $100\times$ larger. This advantage holds true both at inference-time for unseen perturbed cells and for the intrinsic gene features learned in model parameters, suggesting that data curation is a fruitful lever for learning rich representations of perturbations and genes.

\textbf{Architecture and reconstruction objective.} TxFM is an asymmetric MAE. Unmasked input counts are mapped by a gene transformer encoder to a bottlenecked CLS-token representation, which an MLP decoder uses to reconstruct the full expression vector (\autoref{fig:graphical_abstract}, \S~\ref{sec:approach}). Following standard efficiency conventions \citep{he2022masked}, only unmasked tokens pass through the encoder. We introduce a domain-specific reconstruction loss based on the Poisson likelihood, stabilized by a novel \emph{rectified tanh} activation that constrains decoder outputs to the data support while preserving gradient flow.

Our ablation study (\S~\ref{sec:ablations}) empirically identifies each impactful design choice in TxFM, offering practical guidance for building transferable gene expression models. Beyond inference performance on unseen data, TxFM's learned decoder gene weights achieve state-of-the-art recovery of known biological relationships without direct supervision. Interestingly, we find that these relationships concentrate on lower-dimensional manifolds in gene parameter matrices---a phenomenon consistent across pretrained transformers and with the field's understanding of functional gene programs \citep{heimberg2017geneprogram}. Altogether, these results motivate future research into mining pretrained models to discover new mechanisms, biomarkers, and drug targets \citep{schenone2013target}. 

%% file: sections/related_work.tex
% \textit{Other tx models are so bad its crazy and scandalous.}

%... BMDB normally done on perts \cite{kraus2024masked,kenyon2025vitally,bendidi2024benchmarking,ramezani2025genome}. However, running genome-wide perturbational screens is highly expensive, and if rich genomic representations can be obtained non-perturbationally, then new opportunities arise for discovering novel biological gene-gene relationships ...

%... SSL model training on transcriptomics \cite{denadel2024evaluating} also shows that large-scale data is unnecessary for training the current generation of models ...

Existing FMs for transcriptomics use SSL but address the high-dimensional, non-sequential nature of gene expression data in different ways. scBERT \citep{yang2022scbert}, scGPT \citep{cui2024scgpt}, and GeneFormer \citep{theodoris2023transfer} employ masked language modeling. %, and focus on expressed or protein-coding genes only, trading off the ability to represent non-coding RNAs which can have important regulatory functions \citep{statello2021gene}.
% Unlike words in a sentence, gene expression lacks a sequential order, a challenge that requires architectural adaptation.
% scGPT adopted custom causal self-attention masking, while GeneFormer rank-ordered sequences by expression.
% Models also vary in how they represent gene identity and expression.
 scFoundation \citep{hao2024large} embeds raw counts instead of binned or ranked values. UCE \citep{rosen2023universal} uses ESM-2 \citep{lin2022language} embeddings to represent genes. scTab \citep{fischer2024sctab} employs an architecture tailored to tabular data. CellPLM \citep{wen2023cellplm} and scCello \citep{yuan2024cell} jointly model other modalities, and Cell2Sentence \citep{Rizvi2025GOOGLEC2S} casts gene expression as linguistic sentences with genes ordered by decreasing expression.
% The latest transcriptomics FMs are motivated by the scaling laws seen in vision and language, aiming to generate large-scale atlases. 
AIDO.Cell \citep{ho2024scaling}, scPrint \citep{kalfon2025scprint}, and TranscriptFormer \citep{pearce2025cross} scaled pre-training datasets to 50--100 million cell atlases. Tahoe-100M \citep{Gandhi2025TAHOETX1} is a large small-molecule perturbational dataset with multiple recent models pre-trained on it, including the dataset's companion FM Tahoe-x1, GeneJEPA \citep{Litman2025GENEJEPA}, and STATE \citep{Adduri2025STATE}.  %and pushed the scale to 24B parameters with their largest model.

%AIDO.Cell \citep{ho2024scaling} increased the pretraining dataset to 50 million cells across diverse tissues and the context length to \mytilde20k tokens, capturing all protein-coding genes. scPrint \citep{kalfon2025scprint}, with a similarly sized pretraining set, focused on interpretability by separating the embeddings into known cell state covariates. TranscriptFormer \citep{pearce2025cross} introduced a cross-species atlas by pretraining on 100 million cells across 12 species, while CellFM \citep{zeng2025cellfm} scaled to 100 million cells and a 800-million parameter model: the current largest single-species transcriptomics FM.

However, benchmarking studies have found many FMs unable to generalize or outperform simple baselines across a range of tasks despite their scale \citep{boiarsky2024deeper, bendidi2024benchmarking, kedzierska2025zero, wong2025simple, AhlmannEltze2025}. 
% A recent study provides additional evidence by showing that 
Indeed, increasing data size does not necessarily translate to improved performance of SSL models \citep{denadel2024evaluating}.
%This suggests that high-quality data curation might be more advantageous than arbitrary data scaling, relaxing the requirements on the model size and computational resources and motivating our model TxFM.
%In our approach to TxFM, we focused on modeling and architecture design choices suitable to gene expression data and transfer learning tasks, while additionally employing heavy curation to boost data quality.
%Our benchmarking supports the finding that scale alone is not sufficient for generalization (\autoref{tab:summary}). 
%We found that architecture and data quality may be more impactful on model performance than sheer scale, motivating our approach to TxFM.

\textbf{Masked autoencoders (MAEs).} Originally designed for images, MAE \citep{he2022masked} offers a self-supervised learning regime with reliable performance and transfer to downstream biological tasks \citep{kraus2024masked,kenyon2025vitally,kim2025self,richter2025delineating}. 
% Although MAE differs from contrastive learning (CL) in objective formulation, MAE training implicitly aligns multiple masked views of the same input over long training horizons \citep{zhang2022mask,li2024masked}, leading to a degree of functional equivalence between the two regimes \citep{hondru2025masked}. Theoretical analysis suggests that MAE captures both global and local features, whereas CL preferentially emphasizes global structure \citep{huang2025theoretical}; this distinction is particularly relevant for transcriptomics, as both global cell state and localized gene programs are biologically meaningful. 
TxFM belongs to the MAE model family but incorporates several adaptations for count-based gene expression data, in particular an MLP decoder using only the final encoder CLS token (rather than a token-wise transformer decoder), and other design choices described below.

% While there are differences between MAE and contrastive learning (CL), MAE implicitly aligns positive input pairs (each differently augmented by a randomly sampled mask) when trained for many epochs \citep{zhang2022mask,li2024masked}, indicating a certain equivalence between the two regimes \citep{hondru2025masked}. Although, theoretical analysis argues that MAE learns both global and local features while CL favours global features \citep{huang2025theoretical}. TxFM belongs to the model class of MAEs, with several differences we tuned for gene expression data, including only passing the final encoder block's CLS token to an MLP count decoder (rather than a transformer decoder on all latent tokens), reconstruction loss applied on both masked and unmasked expression values, and other architectural contributions we ablate in \S~\ref{sec:ablations}.

%corresponding to the following works, respectively, \citet{Adduri2025STATE,chen2024genept,Rizvi2025GOOGLEC2S,ho2024scaling,pearson1901liii,lopez2018deep,pearce2025cross,Gandhi2025TAHOETX1,cui2024scgpt,kalfon2025scprint,yuan2024cell,rosen2023universal,wen2023cellplm,fischer2024sctab,Litman2025GENEJEPA,theodoris2023transfer}.
%Altogether, we determined that architecture and data quality may be more impactful on current model performance than sheer scale,
% relaxing the requirements on model size and computational resources and 
%motivating our approach to TxFM.

%% file: sections/approach.tex
% Simple SSL based on MAE.
% Curated data.
\input{tables/diverserna_training_data_summary}
Let $\mathbf{x}\in\mathbb{N}^G$ be a gene expression count vector for a single-cell (sc) or bulk RNA-seq sample with $G$ profiled genes. We seek low-dimensional embeddings $\mathbf{e}\in\mathbb{R}^d$ ($d\ll G$) that isolate biological signal from high-dimensional technical noise, enabling downstream analyses such as comparing perturbations and inferring gene relationships.

% \subsection{Our Transcriptomics Foundation Model, TxFM}

We adopt an SSL framework based on masked reconstruction. TxFM consists of two main components: the \textbf{encoder}, a transformer that takes masked gene expression $\mathbf{x}$ as input to produce a sample-level representation $\mathbf{e}$; and the \textbf{decoder}, a lightweight MLP that takes $\mathbf{e}$ as input and predicts a full expression vector $\hat{\mathbf{x}} \in \mathbb{R}^G$. %As gene expression data is provided all at once due to the data-generating processes of mRNA sequencing, we do not model gene expression as a sequence nor include positional embeddings.
Unlike language modeling where word order is critical, a gene expression profile is an unordered set of measurements of the abundance of individual RNA molecules taken simultaneously, akin to a bag-of-words. For this reason, we do not model gene expression as a sequence. 
% In \S~\ref{sec:ablations}, we ablate the impact of each significant architectural decision described below. 

\textbf{Training dataset.}
We train our default model on a curated collection of public transcriptomic datasets with both single-cell and bulk RNA-seq samples, selected with a focus on oncology and perturbational biology. We call this dataset containing $\sim$1.4 million samples \textit{DiverseRNA-1.4M} (\autoref{tab:datasets}). 
% These datasets were curated from the following related works, respectively: \citet{replogle2022mapping,ruiz2022harmonized,mcfarland2020multiplexed,srivatsan2020massively,nowicki2023single,guimaraes2024single,wu2021single,weinstein2013cancer,gtex2020gtex}. 
%
For the large-scale perturbational dataset of CRISPRi-edited K562 cells \citep{replogle2022mapping}, we apply a \textit{phenoprint} curation strategy that filters for perturbations with distinct transcriptional profiles (see \S~\ref{sec:pert_appendix_benchmarks}) to enrich for biologically meaningful variation; a strategy shown by \citet{kenyon2025vitally} to work effectively when training MAEs for microscopy. For the other single-cell datasets, we retain only genes expressed in at least 1,000 cells and cells expressing at least 2,000 genes, yielding an aggregate union of 44,349 unique genes in the training data.

\textit{On dataset heterogeneity.} As different transcriptomic datasets contain different sets of sequenced genes, our DiverseRNA-1.4M has different degrees of gene overlap between its composite datasets. To handle these feature differences, and, importantly, to enable cross-dataset integration, we adopt the following strategy instead of zero-padding: TxFM learns representations for every gene in the vocabulary, but only unmasks and applies reconstruction loss on the genes that are actually measured in an individual sample, as described below.

% \begin{itemize}
%     \item \textbf{Encoder:} A transformer that processes a masked version of the input $\mathbf{x}$ and produces a sample-level representation, $\mathbf{e} \in \mathbb{R}^d$.
%     \item \textbf{Decoder:} A lightweight multilayer perceptron (MLP) that takes $\mathbf{e}$ as input and reconstructs the full expression vector $\hat{\mathbf{x}} \in \mathbb{R}^G$.
% \end{itemize}

\textbf{Preprocessing.} Each input gene expression vector $\mathbf{x}$ is normalized to have a total count of $\sum_{i=1}^G \mathbf{x}_i = L = 10^5$ (library size normalization), then log-transformed after adding a pseudocount to accommodate zero counts: $\mathbf{x}_i = \log(\mathbf{x}_i + 1)$, or log1p. This is a standard normalization procedure for RNA-seq analysis that enables comparisons between samples with varying numbers of total mapped counts \citep{conesa2016survey}.

\textbf{Masking strategy.} To enable self-supervision, we uniformly at random sample a subset $\mathbf{k} \subseteq \{1, \dots, G\}$ of size $K$ and only reveal the values $\{\mathbf{x}_i : i \in \mathbf{k}\}$ to the encoder. The remaining gene expression values are not seen by the encoder. Sampling is uniform across all \textit{observable} genes within each sample; e.g. if $K=2048$, a cell from the K562 dataset \citep{replogle2022mapping} has an effective mask ratio of $\sim$75$\%$ since there are 8,248 possible genes to sample, while a sample from GTEX (\autoref{tab:datasets}) would have an implicit mask ratio of $\sim$94$\%$ ($1 - \frac{2048}{36,695}$).

\textbf{Model.} Each unmasked gene indexed with $i \in \mathbf{k}$ is embedded by \textit{concatenating} a learnable gene embedding $\mathbf{E}_{i} \in \mathbb{R}^{d-1}$ with its preprocessed expression value $\mathbf{x}_i$, forming:
$\text{token}_i = [\mathbf{E}_{i}; \mathbf{x}_i] \in \mathbb{R}^d.$

A learnable CLS token is appended to this set and all $K+1$ tokens are passed through a transformer encoder, relating every unmasked gene to each other through dense self-attention. The transformer's output embedding of the CLS token $\mathbf{e}$ is then passed through an MLP (count) decoder:
$\hat{\mathbf{x}} = \phi(\text{MLP}(\mathbf{e}))$, $\hat{\mathbf{x}} \in \mathbb{R}^G$,
where $\phi$ is a novel \textit{rectified tanh} activation function we introduce and define on the model's predicted gene expression logits $z$ as:
% \begin{equation} \label{eq:activation}
%     \phi(z) = \log(L+1) \text{ReLU}\left(2\sigma\left(\frac{z}{2e}\right) - 1\right),
% \end{equation} %
%
% \begin{equation} \label{eq:activation}
%     \phi(z) = \log(L+1) \text{ReLU}\left(\tanh\left(\frac{z}{4e}\right)\right).
% \end{equation} %
\begin{equation} \label{eq:activation}
    \phi(z) = \alpha  \text{ReLU}\left(\tanh\left(\frac{z}{\alpha}\right)\right), \quad \alpha= \log(L+1) .
\end{equation} %
%
% \begin{equation} \label{eq:activation}
%     \phi(z) = \alpha \text{ReLU}\left(\tanh\left(\frac{z}{\beta}\right)\right), \alpha=\log(L+1), \beta=4e
% \end{equation} %
% \footnote{This can be generalized to $\phi(z) = \alpha \text{ReLU}\left(\tanh(\frac{z}{\beta})\right)$ and unit-size gradient steps can be ensured by $\alpha=\beta$.}
We designed this activation function to respect the non-negative, count-based nature of the data. Unlike pure ReLU, this activation is bounded and asymptotically approaches the normalized upper bound of the data, setting $\alpha=\log(L+1)$, preventing divergence by naturally dampening gradients as the predicted count increases (\S~\ref{sec:activations}, \autoref{fig:activations}).%, providing bounded-influence updates.

\textbf{Loss function.} TxFM can be trained to minimize any reconstruction loss. Our default architecture uses a Poisson-based reconstruction loss function on
% However, the choice should be guided by the data's underlying statistics. Common choices for reconstruction, like the Mean Squared Error (MSE) loss, implicitly assume continuous, normally distributed errors. This assumption is a poor fit for gene expression data, which consists of discrete, non-negative counts. We therefore adopt a Poisson-based loss, as it provides a more statistically appropriate model for this type of data. Accordingly, TxFM is trained to minimize the following Poisson-based reconstruction loss function, which compares 
the model's prediction $\hat{\mathbf{x}}$ and the gene expression in the log-normalized training data $\mathbf{x}$ for each sample (derivation in \S~\ref{sec:appendix_nbloss}):
\begin{equation} \label{eq:loss}
    \mathcal{L}_{\text{Poisson}}(\mathbf{x}, \hat{\mathbf{x}}) = \frac{1}{G}\sum_{i=1}^G{ (e^{\hat{\mathbf{x}}_i} - \hat{\mathbf{x}}_i e^{\mathbf{x}_i})}.
\end{equation} 

By combining this loss with the rectified tanh activation, we constrain the hypothesis class to a library-bounded link function, where the Poisson rate $\lambda \in [1, L+1)$ matches the empirical support of the library-normalized target data with pseudocounts. The rectification in $\phi(z)$ clamps negative gene logits to the minimum rate ($\lambda$=$1$) without an explicit mixture model of zero-inflation.

\textbf{On the match between input data, activation, and loss function.} Library size normalization removes sequencing depth variation while preserving gene rankings. While Poisson is a discrete distribution defined on natural numbers as support, the Poisson likelihood remains a valid, convex, and differentiable objective for continuous targets. Thus, the optimal reconstruction under the Poisson loss corresponds to the same gene expression patterns regardless of whether targets are raw or normalized. %Additionally, we empirically found that having a fixed range for all samples was helpful for model training in conjunction with our novel activation function that constrains model output not to exceed library size. We trained a TxFM variant without applying library size normalization to input data and with evaluating the Poisson loss on integer targets and found that it performed significantly worse than our default approach (\autoref{tab:txfm_ablations}b). 
Unlike MSE or SmoothL1, the Poisson loss generates gradients proportional to the relative prediction error, which prioritizes reconstruction of genes with low-to-moderate expression (which comprise the majority of scRNA-seq data). 

While RNA-seq is often overdispersed \citep{love2014moderated}, Poisson may be a better SSL training objective than negative binomal (NB) and zero-inflated NB (ZINB), as these display vanishing gradients for overestimation errors on high-dispersion genes, while Poisson retains gradient signal (\S~\ref{sec:appendix_gradient_analysis}). We hypothesize that \textit{not} modeling gene dispersion explicitly can force the encoder to absorb biological variation into the CLS token embedding, rather than reducing the loss by increasing the dispersion parameters of a NB model. We interrogated this hypothesis by regressing gene mean and variance against TxFM's encoder parameters, and we found that the model encoded both moments into its latent space (\S~\ref{sec:variance_regression}). Further, as the gradient of the loss passes through our rectified tanh activation function during backpropagation, the $\sech^2$ factor attenuates the gradient for highly expressed genes, focusing the model's capacity on low-to-moderate expression levels where the Poisson approximation is more appropriate (\S~\ref{sec:appendix_nbloss}).

%% file: tables/diverserna_training_data_summary.tex
\begin{table}[t!]
\centering
\caption{Composition of the default curated public TxFM SSL training dataset. sc: single-cell.}
\label{tab:datasets}
\begin{tabular}{llrr}
dataset summary & mode & \# samples & \# genes\\
\toprule
Glioblastoma \citep{ruiz2022harmonized} & sc & 504,929 & 21,310 \\
K562 CRISPRi \textit{phenoprints} \citep{replogle2022mapping} & sc & 502,080 & 8,248 \\
MixSeq cancer cell lines \citep{mcfarland2020multiplexed} & sc & 102,205 & 15,438 \\
sci-Plex compounds  \citep{srivatsan2020massively} & sc & 99,300 & 18,486 \\
Gastric metaplasia \citep{nowicki2023single} & sc & 88,399 & 16,445 \\
Tumor microenvironment \citep{guimaraes2024single} & sc & 71,585 & 17,619 \\
Breast cancer  \citep{wu2021single} & sc & 31,542 & 24,712 \\
TCGA  \citep{weinstein2013cancer} & bulk & 23,733 & 19,594 \\
GTEX  \citep{gtex2020gtex} & bulk & 10,526 & 36,695 \\
\midrule
Total composite dataset: \textbf{DiverseRNA-1.4M} & mixed & 1,434,299 & 44,349\\
\end{tabular}
\end{table}

%% file: sections/results.tex
\input{tables/camera_ready__combined_benchmarks}

We focus evaluation on two primary TxFMs, each using a Base transformer backbone and 768 model dimensions: (1) a TxFM-B trained on our curated DiverseRNA-1.4M (along with several baselines trained on the same data, including a 1024-dimensional scVI \citep{lopez2018deep}, \S~\ref{sec:benchmarked_models}); (2) a TxFM-B trained with 4$\times$ more compute on the 40$\times$ larger atlas-scale TF-Sapiens 57 million single-cell corpus used to train the Transcriptformer model \citep{pearce2025cross}. Hyperparameters in \S~\ref{sec:appendix_hyperparameters}. %Our evaluations span the following comparisons: same architecture, different training data; different architecture, same data; and, different architecture, different data (as many public foundation models are trained on different atlas-scale datasets). %We also include non-embedding input count baselines, and later ablate performance drivers in \S~\ref{sec:ablations}.

\subsection{Inference-time cell perturbation representation learning} \label{sec:perts}

\paragraph{Inductive SSL results.} \autoref{tab:main_overall_results} benchmarks representations of genetically perturbed cells across 16 FMs and strong baselines on three held-out datasets (each with hundreds of thousands of input cells spanning 2,392 CRISPRi genetic perturbations) according to the methodology proposed by \citet{bendidi2024benchmarking}. We focus on the linear probing benchmark, and report performance on the remaining five tasks in \S~\ref{sec:appendix_benchmarks}, where relative performance trends are similar. The linear probing benchmark measures how accurately batch-corrected cell embeddings can be linearly classified into their corresponding perturbations in batches held-out from the linear probe fit. The \textit{overall} aggregates performance across all six tasks (described in \S~\ref{sec:appendix_eval_details}); a higher overall score indicates more biologically meaningful organization of samples and perturbations into discriminative clusters while preserving structural interpretability of underlying gene expression changes. Additional cell representation benchmarks on non-perturbational datasets are reported in \S~\ref{sec:msr_results}.

\textit{Models versus strong baselines.} We first note that standard count-data preprocessing (library size normalization with log transform, or highly variable gene selection) establishes a competitive baseline that many atlas-scale FMs fail to exceed. Against this backdrop, TxFM-B trained on DiverseRNA-1.4M achieves the highest linear probing accuracy and overall score on every held-out dataset, outperforming all competing FMs including STATE-SE \citep{Adduri2025STATE}. Notably, it achieves this despite having nearly $4\times$ fewer parameters, training on $100\times$ less data, and using no auxiliary biological priors such as pretrained ESM2 protein embeddings.

\textit{Factors driving TxFM's performance.} Two controlled comparisons shed light on factors contributing to improved performance. First, TxFM-B trained on the curated DiverseRNA-1.4M substantially outperforms the same architecture trained on the TF-Sapiens 57M cell atlas (35.62 vs 39.11), suggesting that atlas-scale volume alone is not sufficient for high-signal representation of perturbational data inputs. Second, when controlling for training data, TxFM outperforms comparison models on both DiverseRNA-1.4M and TF-Sapiens 57M, indicating that architectural choices are impactful when holding the training data constant. 

\textit{Ablating impact of dataset composition on performance.} Our curation of DiverseRNA-1.4M includes cells from another cell type (K562) sequenced using a similar assay protocol as the evaluation data \citep{replogle2022mapping}, raising a potential concern about data leakage. We address this with a controlled ablation when training TxFM-B (\S\ref{sec:ablations}): removing all K562 cells yields an overall score of 36.73; including only the 72{,}000 K562 control cells (no perturbed cells) improves this to 38.22; the full DiverseRNA-1.4M (including phenoprint-curated perturbed K562 cells) achieves 39.11. All three configurations outperform every competing FM, demonstrating that TxFM's advantage is not attributable to perturbational overlap with the evaluation data, but rather reflects the broader curation of DiverseRNA and the model architecture. We also note that STATE and Tahoe-x1 may benefit from HEPG2 cells in Tahoe training data.

\paragraph{Transductive SSL results.} In \autoref{tab:evaldata}, we evaluate a transductive setting where TxFM is SSL fine-tuned directly on the evaluation dataset, using the same masked reconstruction objective (\S\ref{sec:appendix_sslft}). This reflects a common practitioner workflow where generalization beyond the target data is not the primary goal, instead opting for a bespoke model that learns dataset-specific structure. 

\input{tables/camera_ready_train_on_eval}

SSL fine-tuning the default TxFM-B (pre-trained on DiverseRNA-1.4M) outperforms both PCA and scVI fitted to the same target data, yielding a 5--19\% relative improvement to an overall score of 41.21. Notably, the \emph{inductive} TxFM-B (overall 39.11, \autoref{tab:main_overall_results}) outperforms transductive PCA (overall 38.72), meaning that a forward pass through the pre-trained encoder produces cell representations competitive with a traditional model fit on the evaluation dataset. In \autoref{fig:sslft} (\S\ref{sec:appendix_sslft}), we visualize reconstruction loss epoch-wise when training a TxFM-B from scratch on RPE1 compared to SSL finetuning. Validation loss is already near optimal performance at the start of SSL fine-tuning, thus demonstrating the efficacy of pre-training.

\subsection{Gene representation learning in model parameters} \label{sec:bmdb}

Many transcriptomics models learn static non-perturbational gene representations as a byproduct of training. Transformers often include a learnable codebook of gene token features, while models with full expression decoders (like TxFM and scVI \citep{lopez2018deep}) learn an additional matrix of gene reconstruction weights. Other models, such as UCE \citep{rosen2023universal} and STATE-SE \citep{Adduri2025STATE}, bypass this by leveraging frozen ESM2 protein structure embeddings. As visualized in \autoref{fig:graphical_abstract}, TxFM learns both gene codebook features and decoder gene weights from scratch in matrices \textbf{E} and \textbf{D}, respectively.

\input{tables/codebook_decoder_bmdb}
\input{figures/pca_v2}

\autoref{tab:bmdb} compares 15 different sources of gene representations from the learned parameters of different models. We report gene-gene relationship recall \citep{celik2024building,kraus2024masked,kenyon2025vitally} to determine how well different models intrinsically capture protein-protein interactions, functional connections, and other known relationships annotated in public databases. We compute recall on the model's immediate centered gene matrix, before and after post-processing it with PCA to the optimal performing lower dimensionality (\autoref{fig:pca_dims}). 

\textit{Recall performance comparison.} Our default model, TxFM-B trained on DiverseRNA-1.4M, obtains the highest recall with its decoder's gene parameters both before and after PCA post-processing (42.7\% / 43.9\%); scVI's decoder trained on the same data performs second best (40.4\% / 40.9\%). 
The PCA baseline (fit on the transpose of the same training data) underperforms compared to scVI and TxFM, indicating that non-linear models can capture more relationships from gene expression data. 
Codebooks and decoders from various models trained on atlas-scale data underperform compared to training on DiverseRNA-1.4M, while TxFM trained on TF-Sapiens 57M has higher recall both from its codebook and decoder compared to each other atlas-scale model.

\textit{On the impact of PCA-postprocessing.} Interestingly, we find that PCA-postprocessing the gene embeddings to a lower dimensionality often provides significant improvements, especially for transformer codebook representations (\autoref{fig:pca_dims}). For example, PCA-reducing STATE-SE's 2048-dimensional projection of frozen ESM2 embeddings down to 256 dimensions improves recall by 42\% (22.0$\rightarrow$31.3), and Transcriptformer's gene embeddings similarly improve by 41\% (23.4$\rightarrow$32.9). TxFM's gene representations also benefit from such post-processing, while scVI (a VAE MLP) performs best with full dimensionality. In \S~\ref{sec:appendix_epochwise} (\autoref{fig:epochwise}) we evaluate how relationship recall develops across all 5 databases separately, epoch-wise during training. The TxFM decoder learns gene-gene relationships quickly, within the first 50 epochs, while the codebook features gradually improve up to epoch 150, underperforming the decoder on every database except Signor \citep{perfetto2016signor}, similar to scVI.

% TxFM learns gene representations in both its encoder token embeddings and the decoder reconstruction matrix; cosine similarities between these  gene representations recover known biological relationships (\S~\ref{sec:bmdb}). 

% Taken together, these results show that many fundamental biological relationships emerge in dot products along lower-rank manifolds in high-dimensional transformer-derived gene embedding spaces as a natural byproduct of SSL training on transcriptomics data. 
Taken together, these results show that many fundamental biological relationships emerge naturally in lower-rank manifolds of high-dimensional transformer gene representation spaces, recoverable through simple dot products, as a byproduct of SSL training on transcriptomics data.

% In the next section, we comprehensively ablate TxFM’s architecture and training data to understand the main drivers contributing to performance improvements.

\subsection{Comparing TxFM's learned gene parameters to sequence-based representations}

\autoref{tab:txfm_vs_structure} evaluates gene embeddings made by three sequence-based models, compared to the default TxFM-B decoder gene matrix (\S~\ref{sec:bmdb}). 
% EVO2 \citep{brixi2026genome} is a nucleotide-based sequence model that can create gene embeddings by passing each gene's corresponding nucleotide sequence through the model; we found that the coding region of the DNA sequence (cDNA) worked best for EVO2 on this task. Orthrus \citep{fradkin2026orthrus} is a nucleotide-based sequence model trained on mature RNA transcript-level sequences. ESM2 \citep{lin2023evolutionary} operates on amino-acid sequences to learn evolutionary biology, and can similarly inference on each gene's corresponding amino acid sequence to produce a gene embedding.
EVO2 \citep{brixi2026genome} is a nucleotide-based sequence model that can create 4096-dimensional gene embeddings by passing each gene's corresponding DNA nucleotide sequence through the model; we used the coding region for each gene. Orthrus \citep{fradkin2026orthrus} is a nucleotide-based sequence model trained on mature RNA transcript-level sequences, creating 512-dimensional embeddings. ESM2 \citep{lin2023evolutionary} operates on amino-acid sequences to learn evolutionary biology, and can similarly be evaluated by passing each gene's corresponding amino acid sequence to produce 2560-dimensional gene embeddings.

\input{tables/txfm_vs_structure_bmdb}

The gene-gene relationship benchmarking in \autoref{tab:txfm_vs_structure} shows that these structure-based gene representations \textit{also} benefit substantially from PCA-post processing to a lower dimensionality. TxFM's learned gene parameters substantially outperform each of these models' inference-time gene embeddings on this task. In general, sequence-based models appear to capture more structural relationships such as paralog families, whereas TxFM's decoder matrix captures more pathway biology annotated in the benchmark databases.

\subsection{Ablation of the main properties of TxFM} \label{sec:ablations}
\input{sections/ablations}

%% file: tables/camera_ready__combined_benchmarks.tex
\begin{table*}[tb]
  \centering
  % \setlength{\tabcolsep}{8pt}
  % \small
  \caption{\textbf{Inductive SSL.} \textit{Genetic perturbation representation learning results from model inference.} Benchmarks on RPE1, HEPG2, and Jurkat cell lines, measuring linear separability of perturbations in cellular replicates across batches in an inductive SSL setting (data is unseen during pretraining). We report linear probing performance (\textbf{lin}; mean of top-1 and top-5 accuracy of linear classifier predicting perturbation identity, evaluated on held-out batches) \citep{bendidi2024benchmarking}, mean $\pm$ std across 5 random seeds. \textit{\textbf{Overall}} is the mean of all six downstream tasks (batch effect correction, linear probing, latent space organization, perturbation consistency, biological relationship recall, and latent space interpretability), averaged across the three cell lines. Best is in \textbf{bold}. See \S~\ref{sec:appendix_eval_details} for task descriptions, \S~\ref{sec:appendix_benchmarks} for score breakdown by metric.}
  \begin{tabular}{llcccc}
    % \toprule
    & & \multicolumn{3}{c}{\textbf{Linear probe (lin)}} & \\
    \cmidrule(lr){3-5}
    model & train data, \# samples & \textbf{RPE1} & \textbf{HEPG2} & \textbf{Jurkat} & \textbf{\textit{Overall}}\\
    \toprule
    % \addlinespace
    % \midrule
    \multicolumn{2}{l}{\textbf{Input counts baselines}} & \multicolumn{4}{l}{} \\
      random label shuffle & n/a & 0.9 {\scriptsize $\pm$ .01} & 0.7 {\scriptsize $\pm$ .01} & 0.5 {\scriptsize $\pm$ .00} & \textit{19.43} \\
      (Lib+Log)Norm+1k HVG & evaluation data & 23.7 {\scriptsize $\pm$ .02} & 15.0 {\scriptsize $\pm$ .04} & 9.7 {\scriptsize $\pm$ .01} & \textit{34.46} \\
      (Lib+Log)Norm+5k HVG & evaluation data & 26.7 {\scriptsize $\pm$ .03} & 19.2 {\scriptsize $\pm$ .05} & 13.1 {\scriptsize $\pm$ .01} & \textit{35.08} \\
      Raw data & n/a & 36.7 {\scriptsize $\pm$ .06} & 23.2 {\scriptsize $\pm$ .74} & 20.5 {\scriptsize $\pm$ .05} & \textit{35.27} \\
      (Lib+Log)Norm & n/a & 30.1 {\scriptsize $\pm$ .03} & 21.5 {\scriptsize $\pm$ .05} & 16.7 {\scriptsize $\pm$ .02} & \textit{36.29} \\
    % \addlinespace
    % \midrule
    % \addlinespace
    \midrule
    \multicolumn{2}{l}{\textbf{Atlas-scale SSL} (different data)} & \multicolumn{4}{l}{} \\
      Geneformer-v2 & Genecorpus, 104M & 4.1 {\scriptsize $\pm$ .01} & 2.2 {\scriptsize $\pm$ .00} & 3.0 {\scriptsize $\pm$ .00} & \textit{19.89} \\
      GeneJEPA & Tahoe, 100M & 4.7 {\scriptsize $\pm$ .00} & 2.6 {\scriptsize $\pm$ .00} & 3.1 {\scriptsize $\pm$ .01} & \textit{23.01} \\
      scTab & Atlases, 22M & 6.7 {\scriptsize $\pm$ .01} & 4.6 {\scriptsize $\pm$ .02} & 3.7 {\scriptsize $\pm$ .01} & \textit{24.80} \\
      CellPLM & Tumor atlases, 11M & 7.3 {\scriptsize $\pm$ .00} & 5.1 {\scriptsize $\pm$ .02} & 4.3 {\scriptsize $\pm$ .00} & \textit{24.84} \\
      UCE & Atlases, 36M & 7.4 {\scriptsize $\pm$ .00} & 5.2 {\scriptsize $\pm$ .01} & 5.0 {\scriptsize $\pm$ .00} & \textit{26.44} \\
      scCello & Atlases, 22M & 14.0 {\scriptsize $\pm$ .01} & 8.5 {\scriptsize $\pm$ .03} & 7.3 {\scriptsize $\pm$ .01} & \textit{27.04} \\
      scPrint-L & Atlases, 50M & 16.3 {\scriptsize $\pm$ .02} & 3.0 {\scriptsize $\pm$ .00} & 8.4 {\scriptsize $\pm$ .01} & \textit{27.49} \\
      scGPT & Atlases, 33M & 12.6 {\scriptsize $\pm$ .01} & 7.2 {\scriptsize $\pm$ .01} & 7.7 {\scriptsize $\pm$ .01} & \textit{28.96} \\
      Tahoe-x1-70M & Tahoe+atlases, 266M & 17.2 {\scriptsize $\pm$ .00} & 9.5 {\scriptsize $\pm$ .02} & 9.6 {\scriptsize $\pm$ .01} & \textit{30.99} \\
      AIDO.Cell-100M & Atlases, 50M & 27.7 {\scriptsize $\pm$ .02} & 17.0 {\scriptsize $\pm$ .03} & 15.1 {\scriptsize $\pm$ .00} & \textit{34.50} \\
      Cell2Sentence & Web+atlases, 50M & 26.7 {\scriptsize $\pm$ .00} & 16.1 {\scriptsize $\pm$ .03} & 13.9 {\scriptsize $\pm$ .00} & \textit{35.44} \\
      STATE-SE & Tahoe+atlases, 170M & 29.1 {\scriptsize $\pm$ .00} & 17.2 {\scriptsize $\pm$ .00} & 17.2 {\scriptsize $\pm$ .00} & \textit{36.43} \\
    \midrule
    \multicolumn{2}{l}{\textbf{Atlas-scale SSL} (same data)} & \multicolumn{4}{l}{} \\
      TranscriptFormer-Sapiens & TF-Sapiens, 57M & 18.7 {\scriptsize $\pm$ .01} & 12.2 {\scriptsize $\pm$ .02} & 10.8 {\scriptsize $\pm$ .01} & \textit{32.12} \\
      TxFM-B (ours) & TF-Sapiens, 57M & 27.5 {\scriptsize $\pm$ .00} & 17.8 {\scriptsize $\pm$ .04} & 16.5 {\scriptsize $\pm$ .01} & \textit{35.62} \\
    % \addlinespace
    \midrule
    \multicolumn{2}{l}{\textbf{Curated data SSL} (same data)} & \multicolumn{4}{l}{} \\
      scVI-L (VAE) & DiverseRNA, 1.4M & 22.1 {\scriptsize $\pm$ .01} & 14.2 {\scriptsize $\pm$ .03} & 13.0 {\scriptsize $\pm$ .01} & \textit{32.85} \\
      PCA & DiverseRNA, 1.4M & 16.1 {\scriptsize $\pm$ .01} & 11.1 {\scriptsize $\pm$ .05} & 7.6 {\scriptsize $\pm$ .00} & \textit{33.54} \\
      Linear autoencoder & DiverseRNA, 1.4M & 24.2 {\scriptsize $\pm$ .03} & 16.6 {\scriptsize $\pm$ .07} & 14.1 {\scriptsize $\pm$ .05} & \textit{35.38} \\
      Ridge PCA & DiverseRNA, 1.4M & 28.2 {\scriptsize $\pm$ .07} & 20.2 {\scriptsize $\pm$ .09} & 16.5 {\scriptsize $\pm$ .09} & \textit{36.74} \\
      \rowcolor{tablegray}TxFM-B (ours) & DiverseRNA, 1.4M & \textbf{37.6} {\scriptsize $\pm$ .02} & \textbf{25.9} {\scriptsize $\pm$ .07} & \textbf{23.8} {\scriptsize $\pm$ .01} & \textbf{\textit{39.11}} \\
    % \bottomrule
  \end{tabular}
  \label{tab:main_overall_results}
\end{table*}

%% file: tables/camera_ready_train_on_eval.tex
\begin{wraptable}{r}{0.65\linewidth} %
    \caption{\textbf{Transductive SSL.} Linear probing and overall performance for PCA, scVI, and SSL-fine-tuned (f.t.) TxFM-B models trained separately on RPE1, HEPG2, Jurkat evaluation datasets.}
    \centering
    \begin{tabular}{lcccc}
        & \multicolumn{3}{c}{\textbf{Linear probe (lin)}} & \\
        \cmidrule(lr){2-4}
        model & \textbf{RPE1} & \textbf{HEPG2} & \textbf{Jurkat} & \textbf{\textit{Overall}} \\
        \toprule
        scVI & 19.7 {\scriptsize $\pm$ .02} & 12.6 {\scriptsize $\pm$ .04} & 10.0 {\scriptsize $\pm$ .01} & \textit{35.30} \\
        PCA & 32.0 {\scriptsize $\pm$ .05} & 23.9 {\scriptsize $\pm$ .09} & 15.6 {\scriptsize $\pm$ .02} & \textit{38.72} \\
        TxFM-B f.t. & \textbf{42.1} {\scriptsize $\pm$ .01} & \textbf{29.3} {\scriptsize $\pm$ .04} & \textbf{25.6} {\scriptsize $\pm$ .03} & \textbf{\textit{41.21}} \\
    \end{tabular}
    \label{tab:evaldata}
\end{wraptable}

%% file: tables/codebook_decoder_bmdb.tex
\begin{table}[t]
    \centering
    \caption{\textbf{Whole-genome known gene-gene relationship recall from learned gene parameters} for 155,390 relationships combined over 5 databases (CORUM, hu.MAP, Signor, Reactome-PPI, StringDB). We report the \% of relationships recalled at 95th percentile cosine similarity threshold, $\pm$ recall std from 10 runs with different null distributions sampled from random pairs of protein-coding genes \citep{celik2024building}. Each gene matrix is benchmarked before (raw centered) and after post-processing to its best-performing PCA-reduced \textit{n} components. Best is in \textbf{bold}.}
    \begin{tabular}{lllccr}
        model & layer & train data, \# samples & \textbf{recall} (raw) & \textbf{recall} (PCA) & \textit{n} \\
        \toprule
        \textit{random vectors} & n/a & n/a & 4.9 {\scriptsize $\pm$ .02} & 4.9 {\scriptsize $\pm$ .02} & n/a\\
        \midrule
        Geneformer-v1 & codebook & Genecorpus, 30M     & 18.2 {\scriptsize $\pm$ .16} & 20.2 {\scriptsize $\pm$ .11} & 512 \\
        Geneformer-v2 & codebook & Genecorpus, 104M    & 19.4 {\scriptsize $\pm$ .19} & 20.2 {\scriptsize $\pm$ .11} & 256 \\
        STATE-SE      & projection & Tahoe+atlases, 170M & 22.0 {\scriptsize $\pm$ .12} & 31.3 {\scriptsize $\pm$ .19} & 256 \\
        scGPT         & codebook & Atlases, 33M        & 31.4 {\scriptsize $\pm$ .13} & 31.8 {\scriptsize $\pm$ .16} & 256 \\
        % \midrule
        Tahoe-x1 70M & codebook & Tahoe+atlases, 266M & 29.2 {\scriptsize $\pm$ .10} & 29.6 {\scriptsize $\pm$ .08} & 512 \\
        Tahoe-x1 1B  & codebook & Tahoe+atlases, 266M & 24.0 {\scriptsize $\pm$ .18} & 25.7 {\scriptsize $\pm$ .15} & 256 \\
        Tahoe-x1 3B  & codebook & Tahoe+atlases, 266M & 32.5 {\scriptsize $\pm$ .09} & 35.9 {\scriptsize $\pm$ .15} & 256 \\
        \midrule
        TranscriptFormer & codebook & TF-Sapiens, 57M & 23.4 {\scriptsize $\pm$ .12} & 32.9 {\scriptsize $\pm$ .21} & 256 \\
        TxFM-B (ours)          & codebook & TF-Sapiens, 57M & 36.1 {\scriptsize $\pm$ .09} & 38.3 {\scriptsize $\pm$ .10} & 128  \\
        TxFM-B (ours)          & decoder  & TF-Sapiens, 57M & 40.1 {\scriptsize $\pm$ .13} & 40.6 {\scriptsize $\pm$ .09} & 512  \\ 
        \midrule
        PCA (transpose fit)    & n/a      & DiverseRNA, 1.4M  & 29.2 {\scriptsize $\pm$ .15} & 29.4 {\scriptsize $\pm$ .16} & 1024 \\
        scVI-L & codebook & DiverseRNA, 1.4M  & 23.6 {\scriptsize $\pm$ .15} & 22.3 {\scriptsize $\pm$ .12} & 1024 \\
        \rowcolor{tablegray} TxFM-B (ours) & codebook & DiverseRNA, 1.4M  & 26.3 {\scriptsize $\pm$ .06} & 32.6 {\scriptsize $\pm$ .23} & 64   \\
        scVI-L & decoder  & DiverseRNA, 1.4M  & 40.4 {\scriptsize $\pm$ .24} & 40.9 {\scriptsize $\pm$ .24} & 1024 \\
        \rowcolor{tablegray} TxFM-B (ours) & decoder  & DiverseRNA, 1.4M  & \textbf{42.7} {\scriptsize $\pm$ .15} & \textbf{43.9} {\scriptsize $\pm$ .20} & 256 \\        
    \end{tabular}
    \label{tab:bmdb}
\end{table}

%% file: figures/pca_v2.tex
\begin{figure}
    \centering
    \includegraphics[width=0.9\linewidth]{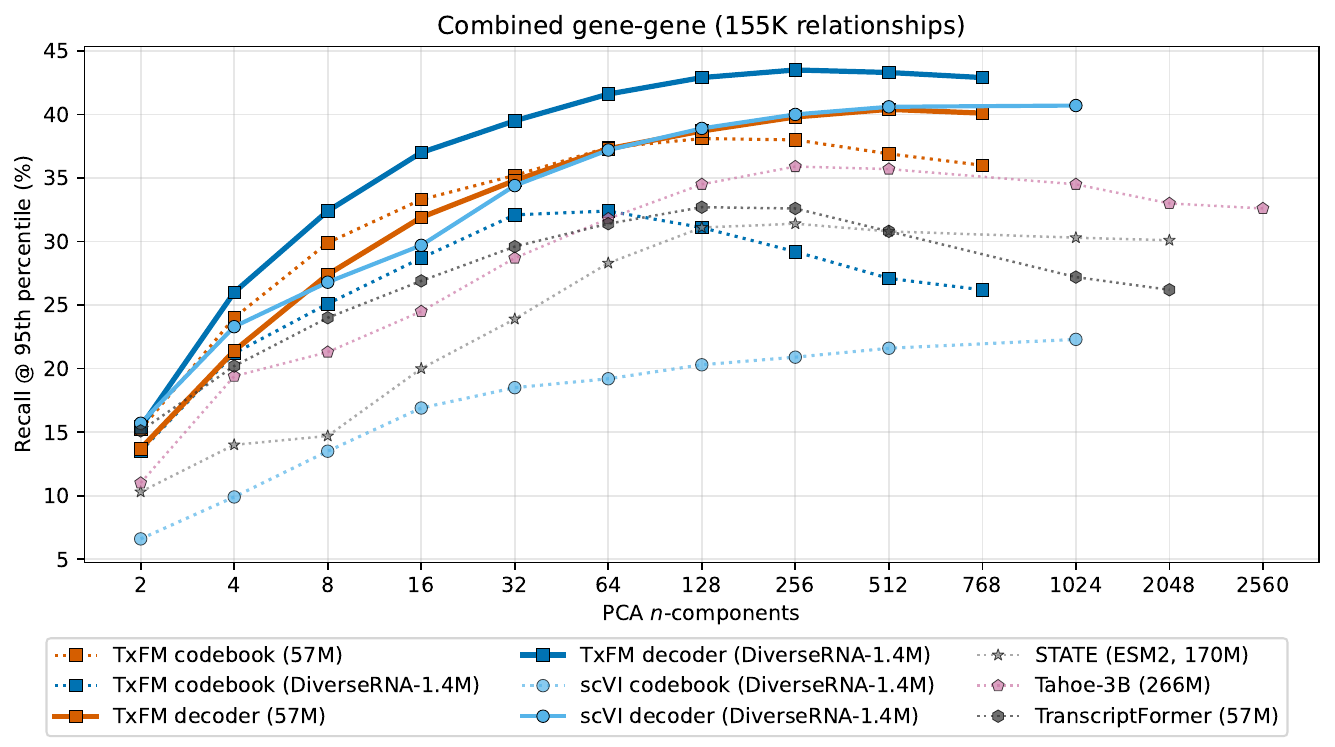}
    \caption{Gene relationship recall as a function of PCA-postprocessing a model's \textit{learned gene embedding parameters} to $n$-components (codebook tokens and/or decoder reconstruction weights).}
    \label{fig:pca_dims}
\end{figure}

%% file: tables/txfm_vs_structure_bmdb.tex
\begin{wraptable}{r}{0.475\linewidth} %
\centering
\begin{tabular}{l c c r}
model & \textbf{rec.} (raw) & \textbf{rec.} (PCA) & \textit{n} \\
\toprule
EVO2 & 12.4 {\scriptsize $\pm$ .06} & 21.3 {\scriptsize $\pm$ .14} & 1024 \\
Orthrus & 14.3 {\scriptsize $\pm$ .07} & 21.9 {\scriptsize $\pm$ .13} & 64 \\
ESM2 & 14.3 {\scriptsize $\pm$ .07} & 29.7 {\scriptsize $\pm$ .15} & 128 \\
\rowcolor{tablegray} TxFM-B & \textbf{42.7} {\scriptsize $\pm$ .15} & \textbf{43.9} {\scriptsize $\pm$ .20} & 256 \\
\end{tabular}
\caption{Gene relationship recall and optimal PCA $n$-components using sequence-based models to create gene representations via inference, compared to the fixed TxFM-B decoder.}
\label{tab:txfm_vs_structure}
\end{wraptable}

%% file: sections/ablations.tex
%\subsection{Design choice ablation of TxFM}
\input{tables/ablations_v1}

\autoref{tab:txfm_ablations} presents our ablation study, inspired by \citet{he2022masked}, on TxFM’s architecture and training data to understand the main drivers contributing to performance improvements. We evaluate TxFM as an encoder by measuring perturbation consistency across the three evaluation datasets unseen during training (one of the six benchmarks used for \S~\ref{sec:perts}), and relationship recall of the post-PCA non-perturbational gene features captured in TxFM's learned parameters (as per \S~\ref{sec:bmdb}).\footnote{If not otherwise specified, TxFM's settings are: library size normalization followed by log1p count preprocessing, $K=2048$ unmasked genes, uniform random mask sampling, a 4-layer MLP decoder, rectified tanh count activation function (\autoref{eq:activation}), Poisson reconstruction loss (\autoref{eq:loss}), Base transformer architecture, trained on DiverseRNA-1.4M (\autoref{tab:datasets}) for 200 epochs, near 1,000 H100 GPU hours per training run. 
% Other hyperparameters are described in \S~\ref{sec:appendix_hyperparameters}, \autoref{tab:model_cards}. 
See \S~\ref{sec:appendix_hyperparameters}, \ref{tab:model_cards}, \ref{sec:appendix_ablations} for hyperparameters and additional details.}

\textbf{(a) Reconstruction loss.} %Each loss is defined with respect to the log1p-transformed ground truth counts $\mathbf{x}$ and the reconstructed count predictions $\hat{\mathbf{x}}$.
We experimented with 6 loss functions (\S~\ref{sec:appendix_nbloss}). Those based on the negative binomial (NB) distribution learn additional dispersion parameters per gene, and the zero-inflated (ZI) ZINB model also modeled gene- and sample-specific ZI probabilities. We also implemented a rank-based cross-entropy objective, similar to scGPT \citep{cui2024scgpt}, with discretized, binned targets. Our Poisson-based loss function (\autoref{eq:loss}) empirically yields significant improvements over the alternatives. While the ZINB variant has an improved codebook, its decoder suffers. %This may be due to the ZI head, creating a new gradient path through the CLS token back to the encoder, thereby relieving the decoder from learning zero-inflation information. %Interestingly, the ZINB model is the only TxFM variant with a codebook that outperforms its own decoder in gene relationship recall. This variant includes a ZI head, creating a new gradient path through the CLS token back to the encoder, thereby relieving the decoder from learning zero-inflation information, which may explain the decline in decoder relationship recall. 
%, as introducing the zero-inflation component caused a surprising decline in decoder relationship recall (41.7 for NB versus 32.1 for ZINB). 
% This suggests that our loss formulation is well-suited to the sparse and count-based nature of gene expression data. 
% One explanation could be that the additional gene-specific parameters lead to a more complex learning task. Alternatively, the Poisson loss could be more amenable to transfer learning in the context of the design choices we made here.
% While a more canonical statistical treatment may lead to more accurate reconstructions of noisy data, our goal is to simply identify a self-supervised loss that yields the highest quality representations for transfer learning.
%While our implementation may deviate from the exact form of the negative log-likelihood, our goal is to simply identify a learning regime that obtains the highest quality representations.

\textbf{(b) Count preprocessing.} Similar to pixel normalization and self-standardization used for image MAEs \citep{he2022masked,kraus2024masked}, we confirm that the standard practice of normalizing inputs by library size \citep{conesa2016survey} is important to achieve good performance. %We library normalize each sample, $\textbf{x}$, to have $\sum_i \textbf{x}_i =$ 100,000, and then apply a log1p transformation $\textbf{x}_i = \log(\textbf{x}_i + 1)$.

\textbf{(c) Mask ratio.} TxFM's effective masking ratio is dataset-dependent on different numbers of sequenced genes in each dataset (\autoref{tab:datasets}), hence we describe this ablation in terms of the exact number of genes \textit{unmasked} during SSL training. Unmasking 2048 genes works most effectively on the DiverseRNA-1.4M dataset, while unmasking 1024 only underperforms slightly but requires significantly less training compute. On this dataset, 2048 unmasked genes entails a nearly 90\% masking ratio, based on the average number of genes per training sample. %This regime outperforms unmasking 4096 genes, despite having significantly less training compute.

\textbf{(d) Masking strategy.} 
We also tested frequency-weighted masking based on a gene's training set sparsity. Let $m_i$ be the number of training samples where gene $i$ has nonzero expression, and define sampling weights $w_i \propto m_i^\tau$ for a temperature $\tau$. Positive $\tau$ preferentially unmasks frequently expressed genes, while negative $\tau$ favors rarely expressed genes. In practice, uniform masking ($\tau=0$) was simplest and gave the best overall trade-off, so we use it by default.

\textbf{(e) Decoder depth.} We assess decoder complexity by comparing a linear mapping to varying-depth MLPs with residual connections. Results indicate minimal sensitivity to depth, confirming that the encoder captures a sufficiently rich representation in the CLS token to enable reconstruction.

\textbf{(f) Decoder count activation.} 
Our proposed \textit{rectified tanh} activation (\autoref{eq:activation}) significantly improves perturbation consistency compared to three alternatives.

\textbf{(g) Encoder backbone.} We ablate encoder capacity by training Small (-S), Base (-B), and Large (-L) variants on DiverseRNA-1.4M. Results indicate a performance plateau on this dataset: moving from -S to -B improves both inference performance and gene token representations, but further scaling to -L did not yield inductive gains despite lower reconstruction loss (\S~\ref{sec:appendix_scaling}).
%, despite -L achieving better validation loss (\S\ref{sec:appendix_scaling}). %We suspect that our relatively small 1.4 million sample dataset could be saturated by TxFM-B despite TxFM-L having overall better reconstruction loss. Additional research in scaling high-quality data is necessary to effectively scale parameters and compute, along with more comprehensive hyperparameter exploration when training larger TxFMs. 
% We therefore also evaluated (\autoref{tab:main_overall_results}) a different TxFM-L pretrained on an expanded dataset with 2.2M samples and then SSL finetuned on patient data (including proprietary data, \S~\ref{sec:appendix_hyperparameters}).

\textbf{(h) Dataset curation.} We vary the curation of DiverseRNA's largest subset, the \citet{replogle2022mapping} K562 CRISPRi dataset (\mytilde1.9M cells), alongside the inclusion of bulk RNA-seq data. Evaluating seven distinct strategies (\S~\ref{sec:datasetcuration}), we find that including K562 data improves overall performance even when restricted to only 72,000 unperturbed control cells. Our default \textit{phenoprint} curation strategy, which prunes $\sim$75\% of the K562 cells with less discriminative signal, effectively improves perturbation representations. Interestingly, including just 33k \textit{bulk} RNA-seq samples from TCGA and GTEX improves both \textit{single-cell} perturbation consistency and decoder parameter recall. Further, removing these 33,000 bulk samples from training degrades performance approximately as much as removing all 450,000 perturbed K562 cells. This highlights a powerful synergy unlocked by using SSL to jointly model single-cell and bulk transcriptomic modalities.
% Doubling the training compute on the larger 2.8M sample dataset does not significantly improve encoder performance compared to our default approach. % which trains with 2x less compute on a 2x smaller dataset. 
% Conversely, scaling to the $40\times$ larger TF-Sapiens dataset (with $4\times$ compute) degrades perturbation representations and decoder structure, despite improving token embeddings.
%, although encoder token relationships are improved. %Future data scaling efforts would likely benefit from intentionally curating additional sources of diverse and rich data, rather than simply including more low-signal samples. %At the same time, future work is necessary to establish precise scaling laws.

%% file: tables/ablations_v1.tex
% Add the following lines to your LaTeX document's preamble:
% \usepackage{booktabs}
% \usepackage{colortbl}
% \usepackage{subcaption}

% Define the gray color used for highlighting default settings in the table

\begin{table*}[t]
% \caption{\textbf{TxFM architecture ablation experiments}. We report total \% of perturbations detected as significant in zero-shot inference over RPE1, Jurkat, and HEPG2 combined, out of 3 $\times$ 2,393 CRISPRi genetic perturbations (perts). We also report gene-gene relationship recall @ 5-95 over 17,073 non-perturbational protein-coding gene representations in the model's learned parameters: (enc)oder gene tokens and (dec)oder gene weights. \hl{Default TxFM architecture is marked in gray.} Performance is significantly different from \hl{Default} when indicated with $\star$ if $p_{adj} < 0.05$, $\star$ if $p_{adj} < 0.01$, and $\star$ if $p_{adj} < 0.001$. Best performance in an ablation is in \textbf{bold}.}
\caption{\textbf{TxFM architecture and data ablation experiments}. We report total perturbation consistency on RPE1, Jurkat, and HEPG2 combined (perts) and whole-genome gene-gene relationship recall captured in a TxFM's learned parameters: (enc)oder gene tokens and (dec)oder gene weights. \hl{Default TxFM architecture is marked in gray.} Performance is significantly different (if $p_{adj} < 0.05$, \S~\ref{sec:appendix_ablations}) from \hl{Default} when indicated with $\star$. Best performance in an ablation is in \textbf{bold}.}

%, according to a 2-proportion z-test for perts and a 2-tailed t-test for enc and dec recall; p-values are adjusted with Benjamini-Hochberg multiple test correction (\S~\ref{sec:appendix_ablations}). Best performance in an ablation is in \textbf{bold}. }

\centering

% --- Row 1 ---
\begin{subtable}[t]{0.48\linewidth}
    \centering
    \begin{tabular}{@{}llll@{}}
        loss & perts & enc & dec \\ \toprule
        MSE         & 27.2$^\star$ & 31.1$^\star$ & 41.8$^\star$ \\
        SmoothL1    & 23.4$^\star$ & 23.2$^\star$ & 38.1$^\star$ \\
        Cross-entropy & 31.2$^\star$ & 30.9$^\star$ & 26.9$^\star$ \\
        \rowcolor{tablegray} Poisson     & \textbf{37.3} & 32.6 & \textbf{43.9}     \\
        NB   & 32.1$^\star$ & 29.1$^\star$ & 41.7$^\star$ \\
        ZINB & 31.1$^\star$ & \textbf{34.2}$^\star$ & 32.1$^\star$\\
    \end{tabular}
    \caption*{(a) \textbf{Reconstruction loss.} Our Poisson-based loss function significantly improves performance.}
\end{subtable}%
\hfill % Adds horizontal space between sub-tables
\begin{subtable}[t]{0.48\linewidth}
    \centering
    \begin{tabular}{@{}llll@{}}
        preprocessing & perts & enc & dec \\ \toprule
        log1p      &   27.9$^\star$    &   25.7$^{\star}$        &  42.2$^{\star}$       \\
        \rowcolor{tablegray} LibNorm, log1p & \textbf{37.3} & \textbf{32.6} &  \textbf{43.9}      \\
    \end{tabular}
    \caption*{(b) \textbf{Count preprocessing.} Normalizing count data for library size helps to learn high-quality perturbational and non-perturbational gene representations.}
\end{subtable}

% \vspace{1em} % Adds vertical space between the rows of sub-tables

% --- Row 2 ---
\begin{subtable}[t]{0.48\linewidth}
    \centering
    \begin{tabular}{@{}llll@{}}
        \# unmasked & perts & enc & dec \\ \toprule
        512           & 31.9$^\star$ &  26.8$^\star$ & 40.2$^\star$ \\
        1024          & 35.2$^\star$ &  31.9$^\star$   & \textbf{44.3} \\
        \rowcolor{tablegray} 2048  & \textbf{37.3} & \textbf{32.6}  & 43.9 \\
        4096          & 35.6$^\star$ &  25.3$^\star$ & 42.8$^\star$ \\
    \end{tabular}
    \caption*{(c) \textbf{Mask ratio.} Training with 2048 unmasked gene tokens ($\sim$90\% mask ratio) yields strong performance, despite 4096 using much more training compute.}
\end{subtable}%
\hfill
\begin{subtable}[t]{0.48\linewidth}
    \centering
    \begin{tabular}{@{}llll@{}}
        temperature $\tau$ & perts & enc & dec \\ \toprule
        -1     & 31.3$^\star$ & 32.5 & 42.7$^\star$\\
        -0.5   & 34.4$^\star$ & \textbf{34.3}$^\star$ & 43.6 \\
        \rowcolor{tablegray} 0 (uniform) &   37.3  &  32.6    &  \textbf{43.9}     \\
        0.5          & \textbf{37.6}  &  28.9$^\star$        &   42.0$^\star$      \\
        1.0          & 36.4 &  25.7$^\star$        &    40.3$^\star$     \\
    \end{tabular}
    \caption*{(d) \textbf{Masking strategy.} Masking genes uniformly at random offers a good performance trade-off vs frequency-weighted masking.}
\end{subtable}

% \vspace{1em} % Adds vertical space between the rows of sub-tables

% --- Row 3 ---
\begin{subtable}[t]{0.48\linewidth}
    \centering
    \begin{tabular}{@{}llll@{}}
        decoder & perts & enc & dec \\ \toprule
        linear &   36.5    &   32.3       &     43.6    \\
        1-layer MLP & 36.0  &  32.3     &    43.8            \\
        \rowcolor{tablegray} 4-layer MLP   & \textbf{37.3} & \textbf{32.6}  & \textbf{43.9}   \\
        8-layer MLP   & 36.5      &    32.0$^\star$      &   43.6      \\
    \end{tabular}
    \caption*{(e) \textbf{Decoder depth.} A linear or MLP decoder works effectively with minimal significant differences.}
\end{subtable}%
\hfill
\begin{subtable}[t]{0.48\linewidth}
    \centering
    \begin{tabular}{@{}llll@{}}
        % activation    & jurkat & enc & dec \\ \midrule
        % ReLU   &   45.2$^\star$    &   21.9$^{\star}$       &   \textbf{42.2}      \\
        % boundedReLU   &  45.2$^\star$      &    \textbf{26.4}      &  \textbf{42.4}       \\
        % \rowcolor{tablegray} rect. sigmoid & \textbf{45.4}     &   \textbf{26.3}       &  \textbf{42.0}  \\
        % softmax       &       &          &         \\
        activation     & perts & enc & dec \\ \toprule
        ReLU (diverges)          & 34.4$^\star$ & 26.7$^{\star}$  & 43.9  \\
        clamped ReLU   & 34.6$^\star$ & 32.4     & \textbf{44.3} \\
        \rowcolor{tablegray} rect. tanh & \textbf{37.3} & \textbf{32.6}  & 43.9 \\
        softmax        &  27.6$^\star$ & 16.2$^\star$ & 15.3$^\star$   \\
    \end{tabular}
    \caption*{(f) \textbf{Decoder count activation.} Our rectified tanh activation function improves inference representations.} 
\end{subtable}

\begin{subtable}[t]{0.38\linewidth}
    \centering
    \begin{tabular}{@{}llll@{}}
        backbone & perts & enc & dec \\ \toprule
        -S (57M) &   33.2$^\star$    &  29.2$^\star$        &  43.9 \\
        \rowcolor{tablegray} -B (159M)   & \textbf{37.3} & \textbf{32.6 } & \textbf{43.9}   \\
        -L (403M)   &  37.2     &   28.2$^\star$       &   43.1$^\star$      \\
    \end{tabular}
    \caption*{(g) \textbf{Encoder backbone.} Scaling TxFM to Base architecture on DiverseRNA-1.4M performed best on these tasks.}
\end{subtable}%
\hfill
\begin{subtable}[t]{0.59\linewidth}
    \centering
    \begin{tabular}{@{}llll@{}}
        train data (\# samples) [compute] & perts & enc & dec \\ \toprule
        Bulk RNA-seq only (33k)   & 21.2$^\star$ & 11.6$^\star$ & 33.5$^\star$\\
        DiRNA w/o bulk RNA (1.4M) & 34.6$^\star$ & 33.1 & 42.4$^\star$\\
        DiRNA w/o K562 cells (932K) & 31.4$^\star$     &  31.5$^\star$    &  42.0$^\star$ \\
        DiRNA w/ K562 controls (1.0M) & 34.3$^\star$ & 32.0$^\star$ & 42.2$^\star$\\
        \rowcolor{tablegray} DiRNA w/ curated K562 (1.4M)   & \textbf{37.3} & 32.6 & 43.9   \\
        DiRNA w/ full K562 (2.8M)   & 35.6$^\star$   &   31.2$^\star$     &  \textbf{44.9}$^\star$   \\
        DiRNA w/ full K562 (2.8M) [2x]  & 36.7   &   29.8$^\star$     &  44.7$^\star$   \\  %%% TODO: UNCOMMENT IF SPACE, or keep appendix
        % TF-Sapiens data (57M) [4x] & 30.3$^{\star}$ & \textbf{35.2}$^{\star}$ & 40.3$^{\star}$ \\   % this is with sub-optimal pca config
        % TF-Sapiens data (57M) [4x] & 30.3$^{\star}$ & \textbf{38.6}$^{\star}$ & 40.6$^{\star}$ \\  % this is with optimal pca config
        % tahoe curated (5.5M) [2x]  & 28.6$^{\star}$ & 24.3$^{\star}$ & 27.5$^{\star}$ \\
    \end{tabular}
    \caption*{(h) \textbf{Dataset curation.} A phenoprints-oriented data curation strategy of DiverseRNA-1.4M is an effective way to train TxFM-B.}
\end{subtable}

% TODO: make training dataset ablations table bigger and more clear
\label{tab:txfm_ablations}
\end{table*}

%% file: sections/conclusion.tex
% We presented TxFM, a data-efficient self-supervised masked autoencoder optimized for the high-dimensional, sparse nature of transcriptomic data. 
% Our results demonstrate that intentional data curation allows TxFM to outperform larger models trained on much larger uncurated cell atlases. This suggests that, in specialized scientific domains, data composition is a more potent driver for effective foundation model transfer learning than sheer volume.
% From an ML architecture standpoint, we empirically isolate our extreme masking strategy (90\%), rectified tanh activation, and Poisson-based loss as the primary drivers of strong transfer learning performance. 
% TxFM establishes a new state-of-the-art for genetic perturbation representation, while learning high-recall gene association maps in its parameters without supervision. Altogether, our results demonstrate TxFM as a practical transcriptomics foundation model with robust transfer across biological contexts.

% \textbf{Limitations and reproducibility.} 
% TxFM was trained and evaluated exclusively on freely accessible public data. While our oncology-oriented curation improves perturbation representation, distributional biases may limit generalization to other contexts.
% Additional research is required to determine scaling laws on transcriptomics data.

% We release the novel modeling components and benchmarking tools here: \url{https://github.com/[redacted]}.

% \textbf{Acknowledgements.} This work was funded and supported by Recursion and Recursion employees.

We present TxFM, a self-supervised masked autoencoder designed for effective transfer learning from transcriptomic count data, and DiverseRNA-1.4M, a curated public dataset designed for training SSL transcriptomics models, enabling faster iteration on fewer computational resources than large atlas-scale data requirements. 

Training TxFM on DiverseRNA-1.4M consistently yields higher-signal gene representations than models trained on much larger atlas-scale datasets, for both inference-time perturbed cell embeddings and the gene features learned within model parameters. 
Architecturally, we find that our high $\sim$90\% masking ratio, Poisson-based reconstruction loss, and a rectified tanh activation were each individually necessary for TxFM's strong transfer performance, resulting in the only model to meaningfully surpass normalized input counts on cellular perturbation representation tasks. 

Lastly, we show that many transcriptomic models generally learn gene parameters that recover many known biological relationships without supervision. TxFM's learned decoder weights achieve the highest gene-gene relationship recall (recovering the most known pathways, protein complexes, etc.) among transcriptomic-based and sequence-based models evaluated, suggesting strong potential for novel biological relationship discovery in service of identifying new drug targets. %\citep{schenone2013target}. 

% We presented TxFM, a self-supervised masked autoencoder carefully designed for effective transfer learning from transcriptomic count data. We also contribute a new curated public dataset, DiverseRNA-1.4M, a  faster iteration and fewer computational resources than the requirements induced by large-scale atlas datasets. consistently yields higher-signal gene representations than training on much larger atlas-scale datasets, for both inference-time perturbed cell embeddings and the gene features learned within model parameters. We found that training on DiverseRNA-1.4M with library normalization, an extreme masking ratio, CLS token bottleneck, Poisson-based reconstruction loss, and our rectified tanh activation were each individually necessary for TxFM's strong transfer performance, resulting in the only model to meaningfully surpass normalized input counts on representation benchmarks.  Lastly, we show that TxFM and other models learn gene parameters that recover many known biological relationships without supervision, suggesting potential for novel relationship discovery in service of identifying new drug targets \citep{schenone2013target}.

\textbf{Limitations and reproducibility.}
Each TxFM checkpoint evaluated in this work was trained and evaluated exclusively on freely accessible public data, including our curated DiverseRNA-1.4M and the uncurated TF-Sapiens 57M dataset. While our oncology-oriented curation of DiverseRNA-1.4M improves both perturbational and non-perturbational gene representation learning, distributional biases may limit generalization to other biological contexts. Additional research is required to establish scaling laws for transcriptomics SSL. We release novel modeling components and benchmarking tools here: \url{https://github.com/recursionpharma/opentxfm}.

\begin{ack}
This work was fully funded and supported by Recursion Pharmaceuticals, which provided computational resources and compensation for the authors. All authors are current or former employees of Recursion Pharmaceuticals and may hold equity, stock, or stock options in the company. The authors declare no other competing financial interests or third-party funding.
\end{ack}

%% file: sections/Z_appendix.tex
\newpage
\section{TxFM training}

\subsection{TxFM training hyperparameters} \label{sec:appendix_hyperparameters}

All TxFM models were trained with data-distributed parallel methods over H100 GPUs on a large-scale compute cluster. Table~\ref{tab:model_cards} describes relevant training hyperparameters. Each model used a one-cycle cosine learning rate decay schedule with 10\% warm-up using the AdamW optimizer with max learning rate 1e-3, betas (0.9, 0.999), epsilon 1e-6, weight decay of 1 divided by the learning rate times the total number of training steps, bfloat16-mixed precision, with additional standard techniques such as LayerScale. The models with 1.4M samples were trained on our curated DiverseRNA-1.4M (\autoref{tab:datasets}). The TxFM-B trained on 57M samples used the much larger CellxGene Transcriptformer-Sapiens dataset \citep{pearce2025cross}; preliminary investigations indicated that $K=1024$ was more suitable for the sparser scRNA-seq data in this larger dataset vs the denser DiverseRNA-1.4M data.
% Our TxFM-L patient finetuned model was trained in two stages on a mix of properietary and public datasets: stage 1 was high-mask ratio pretraining on 2.2M samples and stage 2 was low-mask ratio SSL finetuning on a curated set of 250K samples of single-cell, bulk, and patient data. 
All other model parameter settings, if not otherwise stated, follow the TxFM-B Default model as per the ablations in \S~\ref{sec:ablations}. 
\input{tables/hyperparameters}

\subsubsection{Transductive SSL-finetuning}
\label{sec:appendix_sslft}
In \autoref{tab:evaldata} we evaluate the impact of training unsupervised models directly on the evaluation data to address biological use cases pertaining to individual dataset analysis, where PCA is typically used as the representation. (This is in contrast to the zero-shot setting in \autoref{tab:main_overall_results}, where the evaluation data is entirely unseen during unsupervised training and where we sought to determine how strong pretrained SSL models generalize to new cellular contexts.) Therefore, we evaluated performance when training PCA and scVI models directly on each of the evaluation  datasets (HEPG2, Jurkat, RPE1) and compared to self-supervised finetuning the default TxFM-B on those datasets. Our SSL finetuning performs masked reconstruction on the evaluation dataset with the same setup as pretraining (2048 unmasked genes, 200 epochs). In \autoref{fig:sslft} we visualize the difference in our Poisson reconstruction loss between training a TxFM-B from scratch on RPE1 versus finetuning. We observe that the pretrained model is already very close to high performance on RPE1, and the benefit of finetuning is made clear as it obtains better generalization that the from-scratch TxFM, as the validation set contains held out experimental batches from the assay.

\input{figures/sslft.tex}

\subsubsection{Scaling on DiverseRNA-1.4M}
\label{sec:appendix_scaling}
\autoref{fig:scaling} shows training loss curves for the three different TxFM backbones trained on DiverseRNA-1.4M (\autoref{tab:txfm_ablations} (g), training hyperparameters in \autoref{tab:model_cards}). As we can see, TxFM-L obtains the best reconstruction loss on the validation set during training. However, this unfortunately did not translate to better downstream transfer learning performance. It is possible that the parameter count for TxFM-B saturates DiverseRNA-1.4M, and therefore more curated data should be obtained to effectively scale to TxFM-L. While we would have liked to explore doing so in this work, we sought to instead prioritize our compute budget to comprehensive architecture ablations on the TxFM-B backbone, which already required considerable training compute per standard run. Future work should explore scaling laws and data curation further.
\input{figures/scaling}

\input{figures/epochwise}
\input{figures/layerwise}

\subsection{Epochwise and layerwise analysis of TxFM} \label{sec:appendix_epochwise}

TxFM learns gene representations in both its encoder token embeddings and the decoder reconstruction matrix; cosine similarities between these  gene representations recover known biological relationships. Figure~\ref{fig:epochwise} tracks relationship recall during training of TxFM-B, compared to scVI and PCA trained on the same data. Decoder-based relationships emerge early, while codebook relationships improve more gradually. On Signor and Reactome \citep{perfetto2016signor,fabregat2018reactome}, decoder recall peaks in the first quarter of training and then declines. Interestingly, Signor is the only database where the final epoch's codebook exceeds the decoder, a trend also seen for scVI. TxFM's learned gene weights outperform those of scVI, indicating potential for novel biological relationship discovery for the purpose of finding new drug targets \citep{schenone2013target}. 
% Future work should explore the impact of tying the codebook and decoder parameters \citep{press2017using}.

Figure~\ref{fig:layerwise} evaluates layerwise inference by aggregating perturbation (CLS) representations for three unseen cell lines, extracted from each encoder layer and decoder MLP layer. Representation quality improves after the first encoder layer and continues to increase through the 12th (default) layer, with little change in the decoder MLP. TxFM-B's perturbation representations outperform those of scVI and PCA. The best layer is the final transformer block, unlike some large vision MAEs which find optimal performance at intermediate blocks \citep{alkin2024mimrefiner,kenyon2025vitally}.

\subsection{Loss functions}
\label{sec:appendix_nbloss}

\paragraph{Preprocessing.} We preprocess raw gene expression counts by first performing library size normalization. Library size is the total count summed across all genes in a sample. As library size varies between e.g. single cells, it is a common step to normalize all cells to have equal library size $L$ so that individual gene counts are comparable to each other. In this work, we set $L=10^5$. We then log-transform normalized gene counts, with a prior addition of a pseudocount. This is another common step in RNA-seq analysis, which enables log-transform in the presence of zero gene counts (which are very often found in scRNA-seq data).
\begin{align}
    \tilde{x}_{ng} = \frac{x_{ng}}{\sum_{i=1}^G x_{ni}} \cdot L \\
    \tilde{x}_{ng} = \log (\tilde{x}_{ng} + 1)
\end{align}

$x_{ng}$ is the raw gene expression count for gene $g$ in sample $n$, $G$ is the total number of genes profiled in that sample, and $\log$ is the natural logarithm.

\paragraph{Poisson-based loss.} For an expression count $x_{ng}$, the Poisson negative log-likelihood (NLL) is defined as follows, while we drop the constant below:
\begin{align}
    -\log p(x_{ng}|\lambda) &= -\log \left( \frac{\lambda^{x_{ng}} e^{-\lambda}} {x_{ng}!} \right) \nonumber \\
    &= -x_{ng} \log \lambda + \lambda
\end{align}

Setting the learned rate parameter to our model's prediction as $\hat{\mathbf{x}}_g = \log \lambda$ for the target log-count $\mathbf{x}_g$ (preprocessed as explained above), we recover the loss:
\begin{align}
    \mathcal{L}_{\text{Poisson}}(\mathbf{x}_g, \hat{\mathbf{x}}_g) = 
    e^{\hat{\mathbf{x}}_g} - \hat{\mathbf{x}}_g \cdot e^{\mathbf{x}_g}
\end{align}

We model genes in a sample as i.i.d. and thus sum loss values over genes to obtain $\mathcal{L}_{\text{Poisson}}(\mathbf{x}, \hat{\mathbf{x}})$ for an input gene expression vector $\mathbf{x}$. We used library size-normalized counts with pseudocounts as targets.

\paragraph{Gradient of the loss.} Putting together our rectified tanh activation function $\phi$ and the Poisson-based loss, we can write our loss for a gene count $\mathbf{x}_i$ and its model-generated logit $z$ as:
\begin{align}
    \mathcal{L} = e^{\phi(z)} - \phi(z) e^{\mathbf{x}_i}
\end{align}

Taking the gradient, we obtain:
\begin{align}
    \frac{\delta \mathcal{L}}{\delta z} &= e^{\phi(z)} \phi'(z) - e^{\mathbf{x}_i} \phi'(z) \\
    &= \phi'(z) \left(e^{\phi(z)} - e^{\mathbf{x}_i} \right)
\end{align}

The derivative of the rectified tanh (\S~\ref{sec:activations}) is given by:
\begin{align}
\frac{\delta \phi}{\delta z} &=  \frac{\delta}{\delta z} \left(\log(L+1) \text{ReLU}\left(\tanh\left(\frac{z}{4e}\right)\right) \right) \\
&=
    \begin{cases} 
      \frac{\log(L+1)}{4e} \sech^2\left(\frac{z}{4e}\right) & z>0 \\
      0 & z\leq 0
   \end{cases}
\end{align}

Putting this together and setting $\alpha=\log(L+1), \beta=4e$ for brevity, we obtain the gradient of the loss, where the hyperbolic secant function acts as a saturating gate on the Poisson error term:
\begin{align}
   \frac{\delta \mathcal{L}}{\delta z} &= 
   \begin{cases} 
      \frac{\alpha}{\beta} \sech^2\left(\frac{z}{\beta}\right) \left( e^{\alpha \tanh\left( \frac{z}{\beta} \right)} - e^{\mathbf{x}_i} \right)  & z>0 \\
      0 & z\leq 0
   \end{cases}
\end{align}

\paragraph{Negative binomial-based loss.} For the ablation experiments, we implemented a negative binomial loss function using the Pytorch \texttt{distributions} package. We followed the parameterization by \citet{risso2018general} that models the mean and inverse dispersion as parameters of the distribution and allowed gene-specific inverse dispersion parameters  $\theta_g$ shared across samples. The distribution is then defined as follows for a gene count $x_{ng}$:
\begin{align}
    p(x_{ng} \mid \mu_{ng}, \theta_g) = \nonumber \\
    \binom{x_{ng} + \theta_g - 1}{x_{ng}} \left(\frac{\theta_g}{\theta_g + \mu_{ng}}\right)^{\theta_g} \left(\frac{\mu_{ng}}{\theta_g + \mu_{ng}}\right)^{x_{ng}}
\end{align}

We scaled the learned mean $\mu_{ng}$ for each gene count by the cell size factor defined as $s_n = \frac{\sum_g{x_{ng}}}{L}$. This implies the following reparameterization: $r = \theta_g$ for the number of failures $r$ and $p = \frac{s_n \mu_{ng}}{s_n \mu_{ng} + \theta_g}$ for the probability of success $p$. The loss for a given gene count is given by the negative log-likelihood of $\text{NB}(r,p)$ and loss values are again summed over genes for a given sample. We used raw integer counts as targets as library size is accounted for by the size factor $s_n$.

\paragraph{Zero-inflated negative binomial-based loss.} For the ablation experiments, we implemented a zero-inflated negative binomial loss function building on the NB loss described above. We modeled gene- and cell-specific zero inflation (or dropout) probabilities following \citet{lopez2018deep}. We implemented them as a separate linear layer from the CLS token to $G$ zero-inflation logits for each cell. The ZINB distribution is defined as:

\begin{align}
p(x_{ng} \mid \mu_{ng}, \theta_g, \pi_{ng}) = 
\begin{cases} 
    \pi_{ng} + (1 - \pi_{ng}) \cdot \text{NB}(0 \mid \mu_{ng}, \theta_g) & \text{if } x_{ng} = 0 \\ 
    (1 - \pi_{ng}) \cdot \text{NB}(x_{ng} \mid \mu_{ng}, \theta_g) & \text{if } x_{ng} > 0 
 \end{cases}     
\end{align}

We used raw integer counts as targets as library size is accounted for by the size factor $s_n$.

\paragraph{Cross-entropy loss.} For the ablation experiments, we implemented a rank-based cross-entropy loss. We discretized the log1p-normalized targets into 51 equal-width bins (one dedicated bin for zero expression plus 50 other bins, following the scGPT approach) when computing the loss. We extended our default decoder architecture (a 4-layer MLP generating gene-specific predictions) with a 2-layer nonlinear shared projection from each scalar to 51 bin logits. This modification added negligible number of model parameters ($\sim$3k). We kept the encoder, input representation, masking strategy, and all hyperparameters the same as the default model.

We considered avoiding the scalar bottleneck and experimented with a decoder architecture outputting 51-dimensional logit vector per gene, with a narrower final residual block to control for parameter explosion. However, we found this implementation of the cross-entropy loss to perform worse than the results we report in the ablation study.

%\subsection{Dataset curation} %\label{sec:appendix_curation}

\subsection{Gradient analysis of loss functions}
\label{sec:appendix_gradient_analysis}

\begin{figure*}
    \centering
    \includegraphics[width=\linewidth]{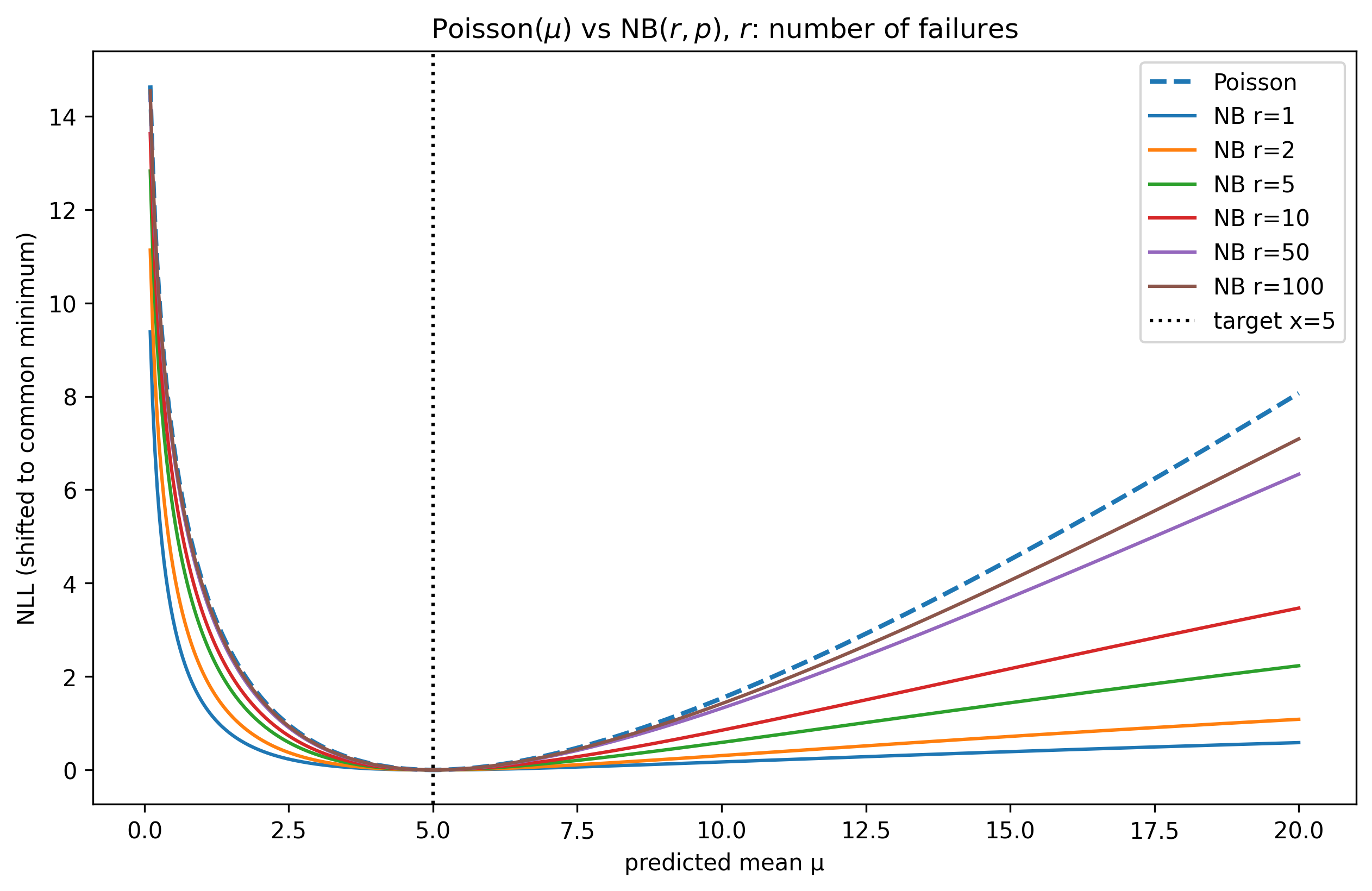}
    \caption{\textbf{Poisson vs NB NLL for a single observation and a range of $r$ values for NB.} The mean of each distribution is plotted on the x-axis and the target is shown with a dashed line. Each curve is shifted on the y-axis so all minima line up at 0 as the gradients are invariant to this shift and for easier comparison of the slopes.}
    \label{fig:pois_nb_gradient_analysis}
\end{figure*}

\paragraph{NB vs. Poisson.} \autoref{fig:pois_nb_gradient_analysis} shows the Poisson and NB NLLs for a single example observation. $\text{NB}(r,p)$ is parameterized with the number of failures $r$ and probability of success $p$, where $r = \theta_g$ for inverse dispersion $\theta_g$. Thus, lower $r$ values correspond to genes with higher dispersion. The loss curves for high-dispersion genes have increasingly small gradients when the mean predictions overestimate the target. In contrast, the loss curves for low-dispersion genes (high $r$) closely approximate the Poisson curve, as expected. For those curves, the gradients are larger for overestimated predictions, providing signal for the optimizer.

We can also see this analytically by considering the NLLs (with constants w.r.t. to the predicted mean dropped):

\begin{align}
    \mathcal{L}_{\text{Poisson}} &= \lambda - x \log \lambda \\
    \mathcal{L}_{\text{NB}} &= - r\log\frac{r}{\mu+r} - x \log\frac{\mu}{\mu+r}
\end{align}

The gradient of each loss w.r.t. to the mean ($\lambda$ for Poisson and $\mu$ for NB) are:

\begin{align}
    \frac{\partial \mathcal{L}}{\partial \lambda} &= 1 - \frac{x}{\lambda} \\
    \frac{\partial \mathcal{L}_{\text{NB}}}{\partial \mu} &= \frac{r}{\mu + r} - \frac{xr}{\mu(\mu + r)} = \frac{r(\mu - x)}{\mu(\mu + r)}
\end{align}

We observe that the NB gradient is the Poisson gradient scaled by $\frac{r}{r + \mu}$. Let's consider two cases: when the model underestimates ($\mu \ll x$) and overestimates the target ($\mu \gg x$).

\begin{itemize}
    \item Overestimation under Poisson: $\lim_{\lambda \to \infty}  \left(1 - \frac{x}{\lambda}\right) = 1$ (gradient is constant, never vanishes)
    \item Overestimation under NB: $\lim_{\mu \to \infty} \frac{r(\mu - x)}{\mu(\mu + r)} = \lim_{\mu \to \infty} \frac{r/\mu - rx/\mu^2}{1 + r/\mu} = \frac{0 - 0}{1 + 0} = 0$ (gradient vanishes)
    \item Underestimation under Poisson: $\lim_{\lambda \to 0^+} \left(1 - \frac{x}{\lambda}\right) = -\infty$ (strong negative gradient)
    \item Underestimation under NB: $\lim_{\mu \to 0^+} \frac{r(\mu - x)}{\mu(\mu + r)} = -\infty$ (strong negative gradient)
\end{itemize}

We can also see that for large $\mu$ in the overestimation case, the gradient of the NB loss $\frac{r(\mu - x)}{\mu(\mu + r)} \approx \frac{r}{\mu}$, meaning that for small $r$ (high dispersion genes), the rate of gradient vanishing is more severe. As a consequence, predictions that overestimate the mean for high-dispersion genes persist under the NB loss. In contrast, the Poisson loss corrects overestimation with a gradient that approaches 1 regardless of dispersion. Both losses can correct underestimation of the mean with strong gradient signal.

\paragraph{ZINB vs NB.} For non-zero targets $x > 0$, the ZINB likelihood equals to the NB likelihood scaled by $1-\pi$ factor. Thus, the same gradient analysis applies as described for the NB loss above, as the zero inflation parameter $\pi$ does not affect the gradient w.r.t.\ $\mu$.

For $x = 0$, the ZINB NLL is:

\begin{align}
    \mathcal{L}_{\text{ZINB}} &= -\log\left(\pi + (1-\pi) \cdot \text{NB}(0; \mu, r) \right)= -\log\left(\pi + (1-\pi) \left( \frac{r}{r+\mu}\right)^r \right)
\end{align}

When $\mu$ is small, it is close to the zero target so the prediction is good and there is no error to correct. 

As $\mu$ increases (the overestimation case) and if $\pi$ is large, the NB component tends to 0 and the NLL saturates at at $-\log(\pi)$. The gradient vanishes because the zero-inflation component $\pi$ explains the observed zero regardless of predicted mean or dispersion. However, when $\pi$ is small, the NB component dominates, though still subject to the same ceiling of $-\log(\pi)$. 

\begin{figure*}
    \centering
    \includegraphics[width=0.48\linewidth]{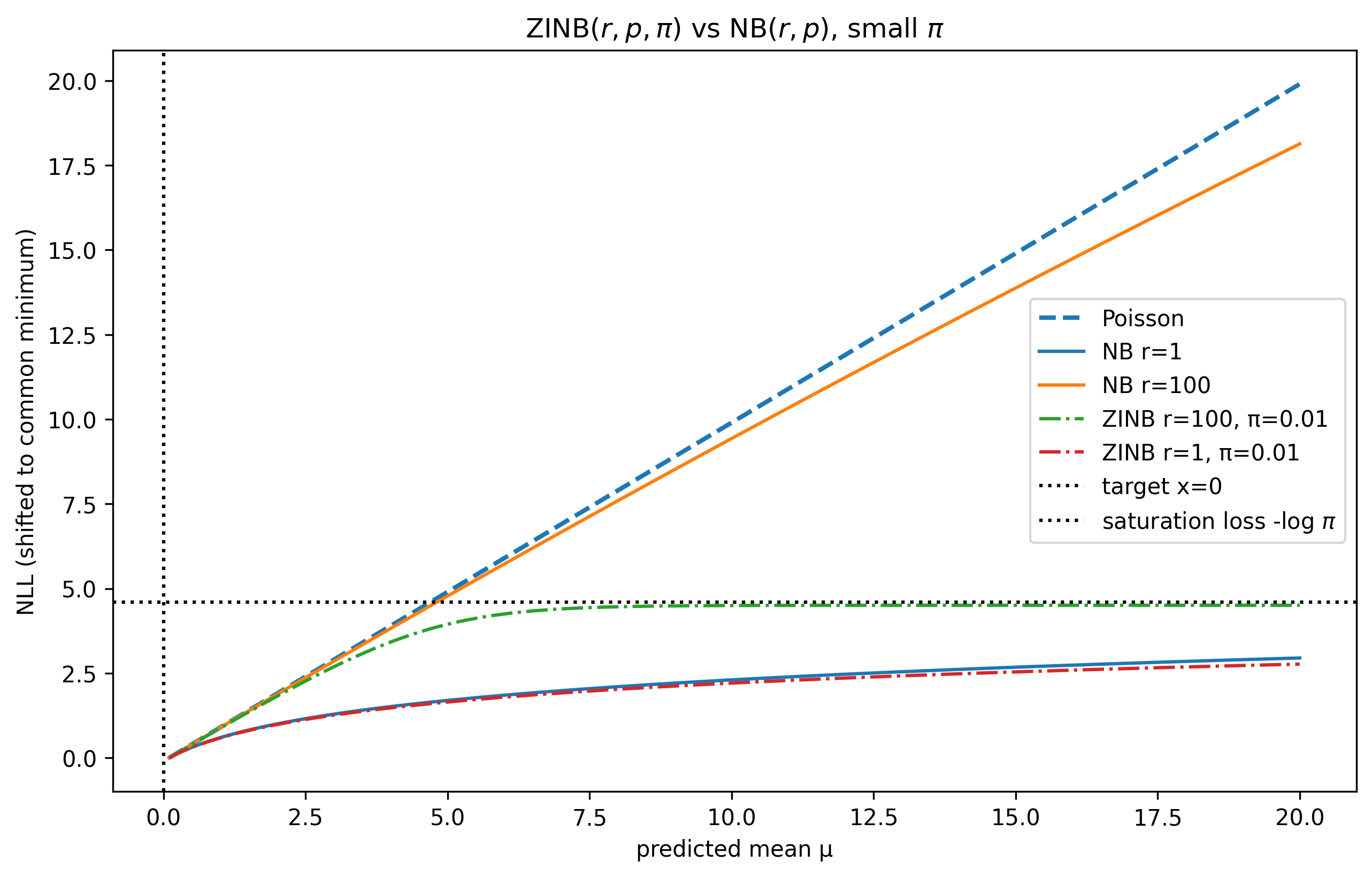}
    \hfill
    \includegraphics[width=0.48\linewidth]{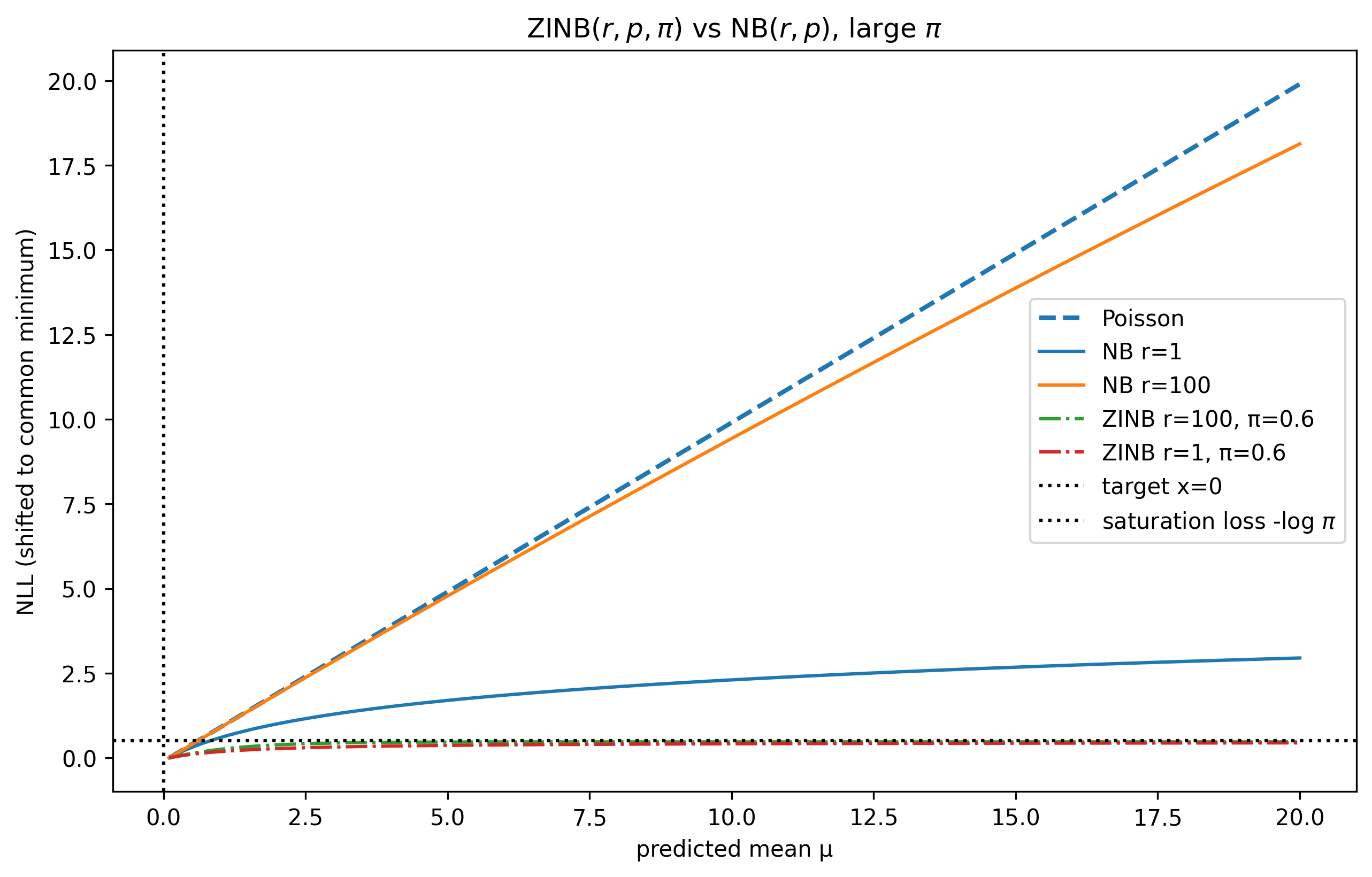}
    \caption{\textbf{ZINB vs NB NLL for a single observation and a range of $r$ values for NB and $\pi$ values for ZINB.} Left: small $\pi$, right: large $\pi$. The mean of each distribution is plotted on the x-axis and the target is shown with a dashed line. Each curve is shifted on the y-axis so all minima line up at 0 as the gradients are invariant to this shift and for easier comparison of the slopes.}
    \label{fig:zinb_nb_gradient_analysis}
\end{figure*}

\autoref{fig:zinb_nb_gradient_analysis} shows that for large $\pi$, the loss curves saturate to $-\log(\pi)$ very quickly for both low- and high-dispersion genes, leading to vanishing gradients. However, for small $\pi$, the high-dispersion ($r=1$) ZINB loss saturates less quickly than the low-dispersion one, and closely approximates the NB loss with the corresponding $r$. On the other hand, the low-dispersion ($r=100$) ZINB loss reaches the saturating ceiling quicker and diverges from the corresponding NB loss, which in turn approximates the Poisson loss. 

In summary, the ZINB loss inherits the NB vanishing gradient behaviours and adds an additional saturation mechanism for zero-count genes via the $-\log(\pi)$ ceiling on the loss.

\subsection{Gene count statistics are encoded into TxFM embedding space}
\label{sec:variance_regression}

\begin{figure*}
    \centering
    \includegraphics[width=0.5\linewidth]{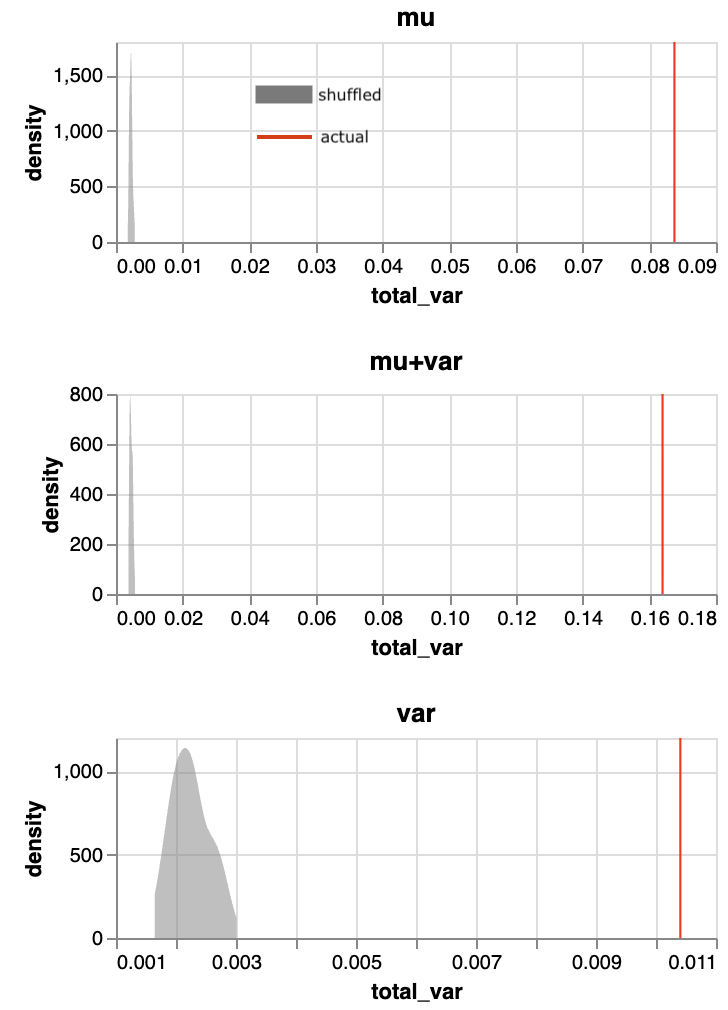}
    \caption{\textbf{Variance explained captured by gene mean and gene variance against principal components of gene representations from TxFM codebook parameters.} Top: variance explained captured by gene mean. Bottom: variance explained captured by gene variance. Middle: variance explained captured by gene mean and variance jointly. Grey densities represent estimates obtained using shuffled gene statistics, red line shows explained variance using actual gene statistics.}
    \label{fig:var_regr}
\end{figure*}

We regressed gene count mean (mu) and variance (var) independently and jointly against projections of TxFM's gene codebook embeddings onto their principal components to obtain estimates of the explained variance captured by these statistics. We repeated this procedure 100 times using shuffled embeddings to obtain a reference null (shown with grey distributions). The actual variance explained is significantly higher than the null for each statistic independently as well as jointly (\autoref{fig:var_regr}). Moreover, TxFM encodes information from each statistic, as evidenced by the increased variance explained when fitted jointly compared to each independently.

\section{Evaluation methodology}
\label{sec:appendix_eval}

% In \autoref{tab:summary} the models correspond to the following works, respectively: \citet{Adduri2025STATE},\citet{chen2024genept},\citet{Rizvi2025GOOGLEC2S},\citet{ho2024scaling},\citet{pearson1901liii},\citet{lopez2018deep},\citet{pearce2025cross},\citet{Gandhi2025TAHOETX1},\citet{cui2024scgpt},\citet{kalfon2025scprint},\citet{yuan2024cell},\citet{rosen2023universal},\citet{wen2023cellplm},\citet{fischer2024sctab},\citet{Litman2025GENEJEPA},\citet{theodoris2023transfer}.

\subsection{Perturbation representation benchmarks}
\label{sec:pert_appendix_benchmarks}

\input{sections/appendix_perturbation_benchmarks}

\input{sections/msr_bench_appendix}

\section{Ablation experiments}
\label{sec:appendix_ablations}

When testing differences in perturbation consistency (perts), recall for encoder gene tokens (enc), and recall for decoder gene weights (dec) between different versions of TxFM, we compared the metric corresponding to each ablation against the Default setting. For perts comparisons, we used a 2-proportion z-test. For enc and dec comparisons, we used a 2-tailed t-test. All resulting p-values were adjusted with Benjamini-Hochberg multiple test correction at 0.05 error rate across all pairs of ablations within each of the three metrics.
%We do FDR correction to have real significant p-values.

\subsection{Activation functions} \label{sec:activations}

\input{figures/activations}

We use a novel activation function called \emph{rectified tanh} (\autoref{eq:activation}) in the decoder to safeguard against predicting counts outside of the training distribution, by leveraging the normalization procedure we employ in preprocessing. Our activation function asymptotically tends to the library size, i.e. the largest possible value an entire expression vector of a preprocessed sample can sum up to. We add 1 to the library size $L$ to mirror the pseudocounts added to input data in our preprocessing. (The activation can be generalized to $\phi(z) = \alpha \text{ReLU}(\tanh(\frac{z}{\beta}))$ and unit-size gradient steps can be ensured by setting $\alpha=\beta$; we recommend setting $\alpha = \log(L+1)$ where $L$ is the library size the data is pre-normalized to ($L=10^5$). In practice, our original implementation set $\alpha=\log(L+1)$ and $\beta = 4e$.)

Indeed, given that the count data is always non-negative, we first tried ReLU as the count activation function $\phi$, but, halfway through training, the model diverges. While a softmax activation struggled to train effectively, we found that simply clamping the ReLU activation based on the maximum possible library size could prevent divergence: 
\begin{equation*}
    \phi_{ClampedReLU}(z) = \min(\text{ReLU}(z), \log(L+1)).
\end{equation*} \autoref{fig:activations} compares rectified tanh to a ReLU and clamped ReLU. Our novel activation function achieved the best zero-shot performance in perturbation representation (\autoref{tab:txfm_ablations}).

% \begin{equation*}
%         \mathcal{L}_{poisson} = e^{\phi(z)} - \phi(z) e^\textbf{x}
% \end{equation*}

% \begin{equation*}
% \begin{split}
%     \frac{d\mathcal{L}}{dz} = e^{\phi(z)} \phi'(z) - e^x \phi'(z)\\
%     \frac{d\mathcal{L}}{dz} =\frac{\alpha}{\beta} \text{sech}^2(\frac{z}{\beta}) \left(e^{\alpha\tanh(\frac{z}{\beta})} - e^\textbf{x} \right)\\
%     \frac{d\mathcal{L}}{dz} = \text{sech}^2(z) \big(e^{\tanh(z)} - e^\textbf{x}\big)
% \end{split}
% \end{equation*}

% \begin{equation*}
%     \phi'(z) = \frac{\alpha}{\beta} \text{sech}^2\left(\frac{z}{\beta}\right)
% \end{equation*}

% \begin{equation*}
%     \phi(x) = \text{ReLU}(\tanh(z))
% \end{equation*}

% \subsection{Masking strategy} \label{sec:masking}

\subsection{Dataset curation} \label{sec:datasetcuration}
We evaluate these dataset curation strategies in \autoref{tab:txfm_ablations} (h):
\begin{enumerate}
    \item Removing all K562 cells from the training data, yielding 932K training samples.
    \item Adding only the 72,000 non-targeting control K562 cells from the \citet{replogle2022mapping} dataset, in addition to the 932K non-K562 samples in the DiverseRNA dataset.
    \item Our default approach, which follows an established strategy shown to improve the performance of MAEs on biological experimental data \citep{kenyon2025vitally} by filtering the perturbational K562 training data to only include cells from perturbations distinguishable from the rest. Specifically, we selected \emph{phenoprint} perturbations by using an earlier version of TxFM to inference the data, computed perturbation consistency as described above (\S~\ref{sec:appendix_eval_details}), and kept perturbations with $p < 0.1$. This more permissive threshold allows us to increase the number of training samples while still keeping perturbations with reasonably distinct transcriptional profiles.
    \item Training with the full uncurated dataset of K562 cells, yielding 2.8M training samples, and holding the compute budget constant by adjusting the number of training epochs according to dataset size; we also evaluate doubling the training compute on the 2.8M dataset to 200 epochs [2x], to about 2,000 H100 GPU hours.
    \item Training TxFM-B with 4x more compute on a different dataset of 57 million cells sampled from 72 large-scale atlases from \cite{czi2025cz}.
\end{enumerate}

% \subsection{Loss curves for encoder backbones} \label{sec:appendix_loss}

\section{Dataset-specific benchmarking results}
\label{sec:appendix_benchmarks}

\autoref{fig:pca_ablation} and \autoref{tab:existing_models_choice} show ablations of the best backbone for each existing model and the best performing number of principal components for PCA on this benchmark. Tables \ref{tab:main_rpe1_results}, \ref{tab:main_hepg2_results}, \ref{tab:main_jurkat_results} show the results on the \citet{bendidi2024benchmarking} benchmark for the RPE-1, HEPG2, and Jurkat datasets, correspondingly.

\input{tables/appendix_model_choice}

\input{tables/rpe1_main_results}

\input{tables/hepg2_main_results}

\input{tables/jurkat_main_results}

\input{figures/pca_ablation}

%% file: tables/hyperparameters.tex
% \begin{table*}[ht]
%     \centering
%     \caption{\textbf{Training hyperparameters} for TxFM variations. Each model used a one-cycle cosine learning rate decay schedule with 10\% warm-up using the AdamW optimizer with max learning rate 1e-3, betas (0.9, 0.999), epsilon 1e-6, weight decay of 1 divided by the learning rate times the total number of training steps, bfloat16-mixed precision, with additional standard techniques such as LayerScale. The models with 1.4M samples were trained on our carefully curated DiverseRNA-1.4M (\autoref{tab:datasets}), and the model trained on 60M samples was on the much larger CellxGene dataset.}
%     \begin{tabular}{lcccc}
%     \toprule
%     \textbf{Hyperparameter} & TxFM-S & TxFM-B Default & TxFM-L & TxFM-B CxG \\
%     \midrule
%     training dataset \# of samples & 1.4M & 1.4M & 1.4M & 60M \\
%     training epochs   & 200    & 200  & 200 & 50   \\
%     K, \# unmasked tokens) & 2048 & 2048 & 2048 & 1024  \\
%     global batch size & 1536   & 1536 & 1536 & 6144   \\
%     \# encoder blocks & 6 & 12& 24 & 12  \\
%     \# MHSA heads & 6 & 12  & 16 & 12 \\
%     model embedding dimension & 384 & 768 & 1024 & 768    \\
%     stochastic depth rate  & 0.1    & 0.1   & 0.3 & 0.1    \\
%     total learnable parameters & 57M  & 159M & 403M & 185M   \\
%     total training GPU-hours  & 400    & 864 & 2560 & 4096  \\
%     \bottomrule
%     \end{tabular}
%     \label{tab:model_cards}
% \end{table*}

\begin{table*}[h]
    \centering
    \caption{Training hyperparameters for TxFM variations.}
    \begin{tabular}{lcccc}
    % \toprule
    hyperparameter& TxFM-S  & \hl{TxFM-B Default} & TxFM-B CxG & TxFM-L\\
    \toprule
    training dataset \# samples &1.4M& 1.4M & 57M & 1.4M\\
    training epochs     &200& 200 & 50  & 200  \\
    K (\# unmasked tokens) &2048& 2048 & 1024 & 2048 \\
    global batch size   &1536& 1536 & 6144  & 1536 \\
    total training GPU-hours    &400& 864 & 4096 & 2500  \\
    \# encoder blocks &6& 12& 12  & 24\\
    \# MHSA heads &6& 12  & 12 & 16 \\
    model embedding dim.  &384& 768 & 768  &1024   \\
    stochastic depth rate     &0.1& 0.1   & 0.1  & 0.3  \\
    total learnable parameters   &57M& 159M  & 185M & 403M  \\

    % \bottomrule
    \end{tabular}
    \label{tab:model_cards}
\end{table*}

%% file: figures/sslft.tex
\begin{figure}
    \centering
    \includegraphics[width=0.7\linewidth]{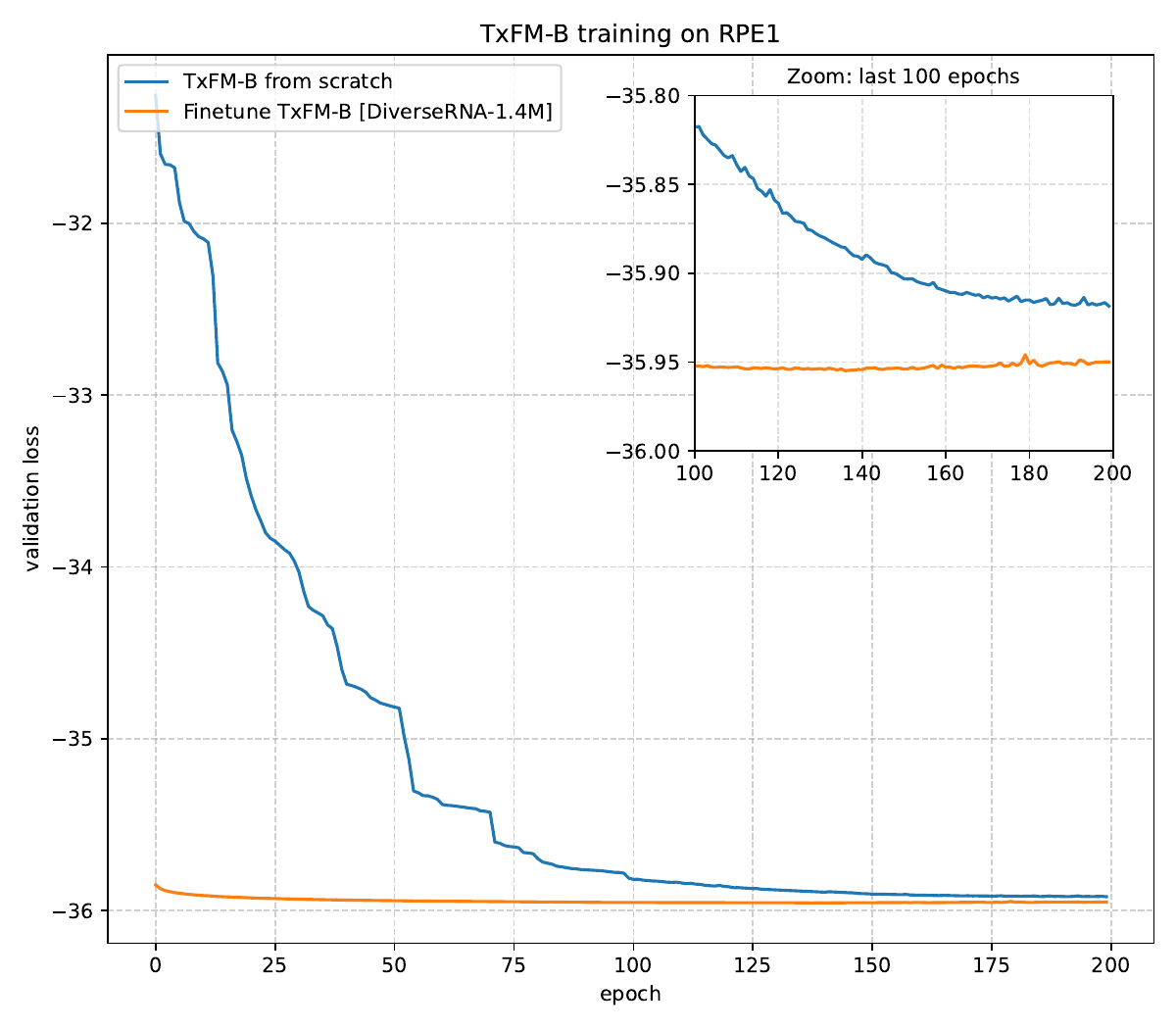}
    \caption{Example of SSL training a TxFM-B on the RPE1 dataset from scratch versus SSL finetuning \hl{TxFM-B Default} (pretrained on DiverseRNA-1.4M) on the RPE1 dataset.}
    \label{fig:sslft}
\end{figure}

%% file: figures/scaling.tex
\begin{figure}
    \centering
    \includegraphics[width=0.7\linewidth]{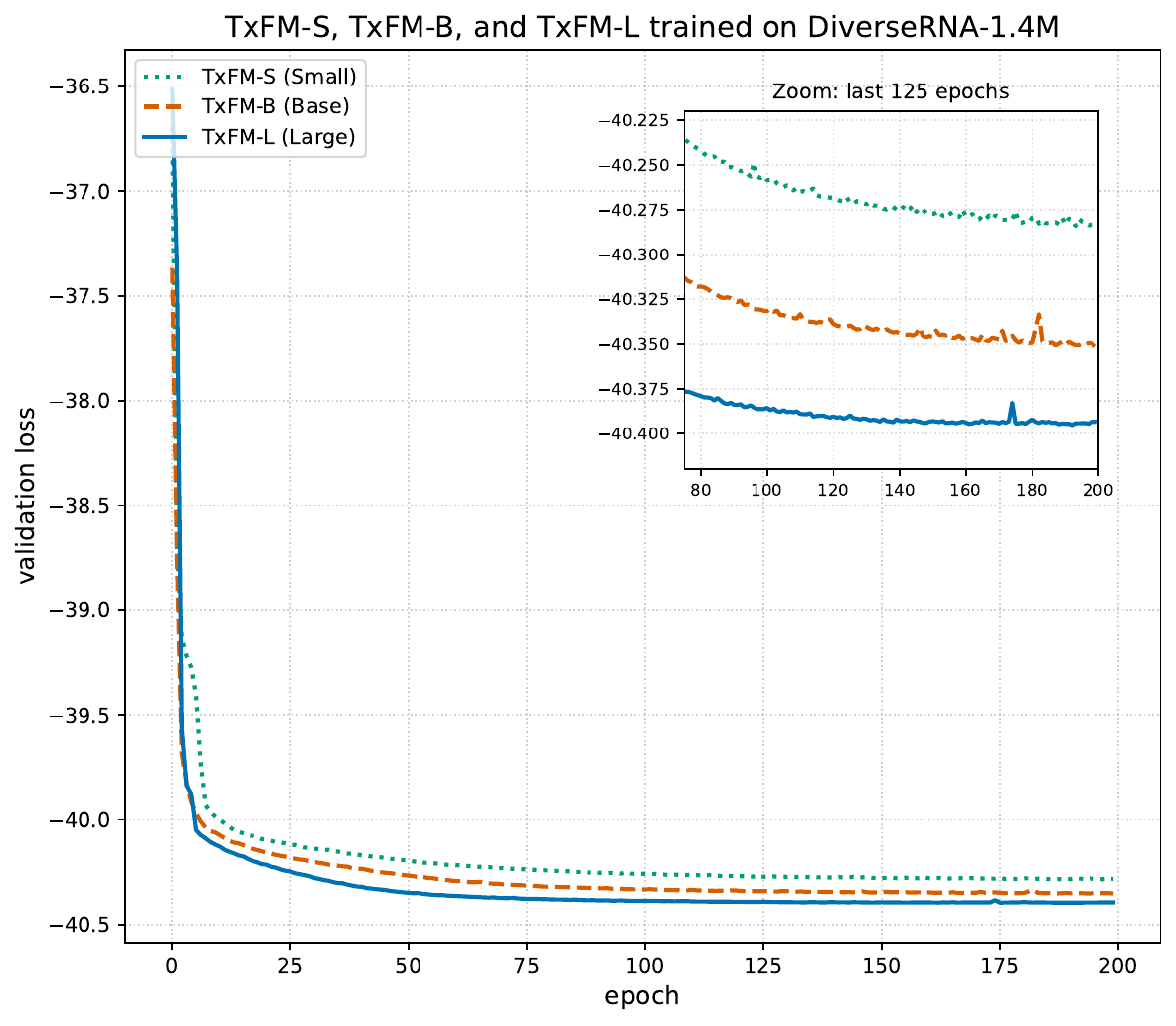}
    \caption{Validation set (held out experimental batches from each of the composite datasets) reconstruction loss curves for the three different TxFM backbones evaluated in \autoref{tab:txfm_ablations} (g).}
    \label{fig:scaling}
\end{figure}

%% file: figures/epochwise.tex
\begin{figure*}
    \centering
    \includegraphics[width=\linewidth]{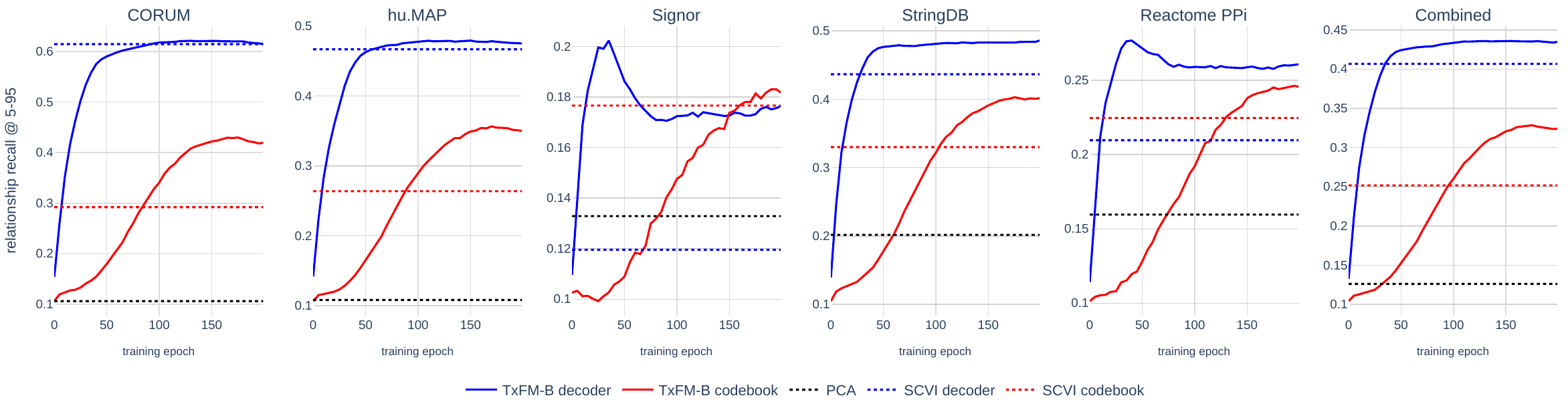}
    \caption{\textbf{Epochwise analysis of TxFM during training.} Gene-gene relationship recall \citep{kraus2024masked} performance on 5 databases (plus all combined) of gene representations extracted from the \hl{Default TxFM-B} codebook and decoder, as a function of training epoch. Dashed lines indicate the recall for the equivalent layers of earlier versions of the scVI and PCA encoder baselines trained on DiverseRNA-1.4M.}
    \label{fig:epochwise}
\end{figure*}

%% file: figures/layerwise.tex
\begin{figure*}
    \centering
    \includegraphics[width=\linewidth]{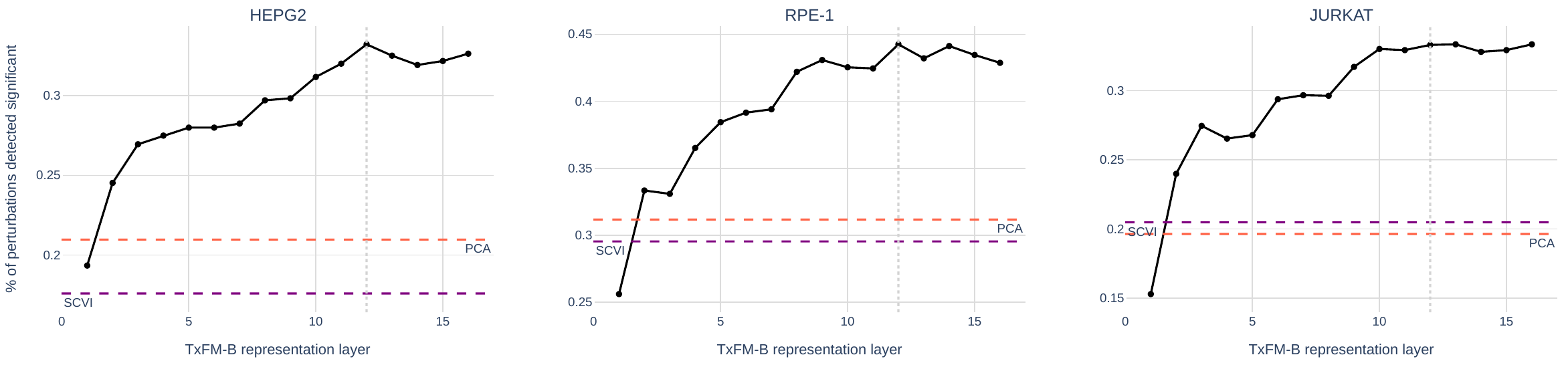}
    \caption{\textbf{Layerwise inference-time analysis of TxFM.} Perturbation consistency on HEPG2, RPE-1, and Jurkat datasets for each layer of our \hl{Default TxFM-B} (encoder layers followed by decoder layers) compared to scVI and PCA (each trained on DiverseRNA-1.4M).}
    \label{fig:layerwise}
\end{figure*}

%% file: sections/appendix_perturbation_benchmarks.tex
\subsubsection{Evaluation setup}
\label{sec:appendix_eval_details}

We adopt the hierarchical evaluation framework introduced in \cite{bendidi2024benchmarking}, which assesses model performance on gene perturbation tasks using a biologically motivated metric suite. Each metric evaluates a different aspect of model utility, with aggregation occurring across runs, metrics, and tasks: results of evaluations performed at 5 different seeds are first averaged together, then groups of metrics constituting a downstream task are averaged together for one final metric per downstream task. We separately apply different post-processing techniques to the embeddings and count baselines before evaluation: centering on negative controls, standardization, and TVN \citep{TVNAndo161422}. For each downstream task and evaluated model pair, we pick the post-processing technique resulting in the best performing average score.

\paragraph{Batch Effect Correction (ilisi):} To quantify batch mixing in the latent space, we use the Integration Local Inverse Simpson’s Index (iLISI). A higher iLISI indicates better mixing of batches and hence more effective batch effect reduction. The final iLISI score for each model is averaged across all batches.

\paragraph{Linear Separability (lin):} This metric assesses how linearly distinguishable different perturbations are in the latent space using a linear classifier trained on known perturbation labels. Top-1 and Top-5 accuracy using unseen batches for evaluation are reported after being averaged together.

\paragraph{Latent Space Organization (knn):} Local structure is assessed via k-nearest neighbor retrieval of perturbation-matched samples across disjoint biological batches. Accuracy is reported for an average of Top-1 and Top-5 retrieval.

\paragraph{Perturbation Consistency (p.cst):} 
%This metric measures whether embeddings of the same perturbation are more similar to each other than expected by chance. For each perturbation, we compute the mean intra-perturbation cosine similarity and compare it to a null distribution derived from random cells. A perturbation is considered consistent if its similarity exceeds the null at an empirical p-value of $p<$ 0.01 (further details in \S~\ref{sec:appendix_benchmark_explanation}). The final score is the fraction of consistent perturbations.

This metric measures whether embeddings of the same perturbation are more similar to each other than expected by chance. We compute perturbation consistency by comparing a perturbation's similarity to a null distribution. Here, we refer to all cells that received a given gene knockout (regardless of the guide used) when we say ``perturbation''. To assemble a null distribution of cosine similarities, we sample $N$ random cells and compute cosine similarity between their embeddings and the mean embedding of each gene perturbation (computed across corresponding cells). We remove pairwise similarities that correspond to the random cell's perturbation to exclude self-self similarities. In our experiments, we set $N=10,000$. Then, for each perturbation, we compute mean leave-one-out cosine similarity across the cell embeddings. Finally, for each perturbation, we compute empirical p-value by comparing its similarity to the null distribution. Perturbations with $p < 0.01$ are considered significantly consistent. The final score is the fraction of consistent perturbations.

Perturbation consistency is computed post-alignment consisting of a PCA transform fit on the non-targeting control samples and followed by standardizing the transformed embeddings within each batch.

\paragraph{Biological Relationship Recall (bmdb):} To test zero-shot biological reasoning, cosine similarities between perturbation embeddings are used to predict gene-gene relationships. Predicted links (top 5\% most similar/dissimilar pairs) are compared to curated databases (e.g., CORUM, HuMAP). Recall is computed per database and averaged across databases.

\paragraph{Latent Space Interpretability (inv):} This is composed of an average of two metrics: Spearman correlation and Structural Integrity. Spearman correlation measures how well latent embeddings can be linearly decoded back into gene expression values. This tests biological interpretability of learned representations. Structural Integrity assesses how well perturbation-induced gene expression changes are preserved in reconstructed profiles. For each batch, we compute the Frobenius norm of the difference between the negative control-centered predicted and the negative control-centered actual gene expression matrices:

\[
\text{Structural Distance} = \frac{1}{B} \sum_{b=1}^{B} \frac{1}{n_b} \left\| \tilde{Y}^{(b)}_{\text{pred}} - \tilde{Y}^{(b)}_{\text{actual}} \right\|_F
\]

where $B$ is the number of batches, $n_b$ the number of samples in batch $b$, and $\tilde{Y}$ denotes gene expression centered by the batch-specific control. The Structural Integrity is then defined as:

\[
\text{Structural Integrity} = 1 - \frac{\text{Structural Distance}}{\text{Structural Distance}_{\text{max}}}
\]

where $\text{Structural Distance}_{\text{max}}$ is an upper bound estimated from the actual data. Higher scores indicate better preservation of perturbation-induced gene expression structure.

\subsubsection{Benchmarked approaches}
\label{sec:benchmarked_models}

We evaluate TxFM across a range of baselines including: commonly used existing methods (PCA and scVI \citep{lopez2018deep}); simple transformations applied directly to the test data (raw counts, library size + log normalization, and its variants with 5000 or 1024 highly variable genes); a baseline of 1024 highly variable genes with randomly shuffled labels (Random label shuffle); models using ChatGPT-derived embeddings (GenePT Large, \cite{chen2024genept}); and large pretrained models trained on CellxGene or similar datasets (Geneformer \citep{theodoris2023transfer}, CellPLM \citep{wen2023cellplm}, scTab \citep{fischer2024sctab}, UCE \citep{rosen2023universal}, scCello \citep{yuan2024cell}, scPrint-L \citep{kalfon2025scprint}, TranscriptFormer-Sapiens \citep{pearce2025cross}, scGPT \citep{cui2024scgpt}, AIDO.Cell-100M \citep{ho2024scaling}), GeneJEPA \citep{Litman2025GENEJEPA}, Tahoe-x1 \citep{Gandhi2025TAHOETX1}, Cell2Sentence \citep{Rizvi2025GOOGLEC2S}, STATE \citep{Adduri2025STATE}. 

We report results in two settings: (i) zero-shot, where models are trained on data disjoint from the test set and (ii) fit-on-evaluation-data, where models are trained or fine-tuned on the test set. Raw counts baseline uses unprocessed gene counts, Lib+Log applies library size normalization followed by a log-transform, HVG variants subset the genes to either 5000 or 1024 highly variable ones, and Random label shuffle keeps gene inputs fixed while randomizing the perturbation labels. PCA and scVI are evaluated in both settings using the standard benchmark splits.

Our TxFM models include TxFM-B DiverseRNA-1.4M, trained on a curated 1.4 million sample dataset, and TxFM-B TF-Sapiens, trained on a larger \mytilde50 million sample dataset. These models allow to isolate the effects of data curation versus training architecture and enable a direct comparison with FMs trained on the same data such as TranscriptFormer-Sapiens. 

For the zero-shot setting, we trained multiple 232M-parameter scVI models on DiverseRNA-1.4M with the following configuration: latent dimensionality of 768, a single hidden layer with 1024 units, AdamW optimizer, lr=0.0005, three variants of likelihoods (negative binomial, zero-inflated negative binomial, Poisson), two settings for weight decay (default value of 1e-06 or 0.03), and other settings set to default values. For the Poisson model, we adjusted the learning rate to 0.0001 and used default weight decay to help with training stability. We compared these scVI models for inductive performance on the held-out data using gene-gene relationship recall and perturbation consistency benchmarks (described in \S~\ref{sec:appendix_eval_details} and under \textit{Perturbation Consistency} above). The model trained with the negative binomial likelihood and default weight decay performed best on both benchmarks for all three datasets; we used this model to report scVI inductive performance and evaluate its codebook and decoder representations. For the transductive SSL setting, we trained scVI on each dataset following the procedure used in \citet{bendidi2024benchmarking}: using 8000 highly variable genes, latent dimensionality of 256, a single hidden layer with 512 units, no dropout, a zero-inflated negative binomial likelihood and gene-level dispersion parameters.

For Cell2Sentence \citep{Rizvi2025GOOGLEC2S}, we use the 2 billion parameter pretrained version and tried several different prompting strategies. We found that their default recommended prompting strategy performed better than prompting it to predict the perturbation, arriving at the following template: 

``The following is a list of $N$ gene names ordered by descending expression level in a Homo sapiens cell. Your task is to give the cell type which this cell belongs to based on its gene expression. Cell sentence: $g_1, g_2,\ldots,g_N$. The cell type corresponding to these genes is:'', 

such that $N$ is the number of genes (sorted by expression value in a given sample) we include in the prompt. The sample-level embedding taken is the last token of the model's hidden state of the sequence. We found that $N=1000$ seemed to perform best in perturbation consistency on the HEPG2 data (i.e., $N=200$ yielded 23.3\% p. cst, $N=500$ yielded 24.7\%, $N=1000$ yielded 24.9\%, but $N=2000$ yielded 23.5\%).

%% file: sections/msr_bench_appendix.tex
\subsection{Additional analysis: cell type clustering and classification with scIB} \label{sec:appendix_msr_bench}

We evaluate embeddings on clustering and batch integration tasks on the five single-cell datasets\footnote{We note that the \emph{Pancreas} dataset contains non-integer values, implying prior normalization. As the original raw counts are not available and we expect negligible impact on performance, these values were used as input for all models.} used in~\citet{kedzierska2025zero}. Additionally, we assess representation quality through classification probing using both linear classifiers and 2-layer MLPs trained on the learned embeddings. For clustering and batch integration, we employ metrics introduced by \cite{luecken2022benchmarking} and implemented in the \texttt{scib} package\footnote{\url{https://scib.readthedocs.io/}}.

\input{tables/msr_benchmark_per_dataset}

\subsubsection{Clustering task}
The clustering task assesses whether cell embeddings preserve meaningful biological structure by measuring how well cells cluster according to their true cell types. We employ Louvain clustering with resolution optimization: we test 10 different resolution parameters and select the resolution that maximizes $\mathrm{NMI}$ between discovered clusters and ground truth labels. Results are obtained by subsampling 10,000 cells with 10 random seeds and reporting the average. Three complementary metrics evaluate different aspects of biological preservation:

\textbf{Normalized Mutual Information ($\mathrm{NMI}$):} Measures information overlap between discovered clusters $\mathcal{C}$ and true cell type labels $\mathcal{L}$:
\begin{align}
\mathrm{NMI}(\mathcal{C}, \mathcal{L}) = \frac{2 \cdot I(\mathcal{C}, \mathcal{L})}{H(\mathcal{C}) + H(\mathcal{L})}
\end{align}
where $I(\mathcal{C}, \mathcal{L}) = \sum_{c,\ell} p(c,\ell) \log\frac{p(c,\ell)}{p(c)p(\ell)}$ is mutual information and $H(\cdot)$ denotes entropy. High $\mathrm{NMI}$ indicates that knowing cluster assignments provides substantial information about true cell types.

\textbf{Adjusted Rand Index ($\mathrm{ARI}$):} Measures pairwise agreement between clusterings, corrected for chance:
\begin{align}
\mathrm{ARI} = \frac{\mathrm{RI} - \mathbb{E}[\mathrm{RI}]}{\max(\mathrm{RI}) - \mathbb{E}[\mathrm{RI}]}
\end{align}
where $\mathrm{RI} = \frac{n_{S} + n_{D}}{\binom{n}{2}}$ counts pairs that are consistently assigned (same cluster in both clusterings: $n_{S}$; different clusters in both: $n_{D}$). $\mathrm{ARI}$ is more conservative than $\mathrm{NMI}$, requiring precise boundary correspondence for high scores.

\textbf{Average Silhouette Width ($\mathrm{ASW}$):} Measures separation of true cell types in embedding space:
\begin{align}
\mathrm{ASW} = \frac{1}{n} \sum_{i=1}^{n} \frac{b_i^{(l)} - a_i^{(l)}}{\max(a_i^{(l)}, b_i^{(l)})}
\end{align}
For each cell $i$ with true label $\ell_i$, $a_i^{(l)}$ is the mean distance to other cells with the same label, and $b_i^{(l)}$ is the mean distance to cells from the nearest different label. High $\mathrm{ASW}$ indicates well-separated cell types.

\subsubsection{Batch integration task}
Batch integration evaluates the dual challenge of removing technical artifacts (batch effects) while preserving biological signal. Batches represent different experimental conditions that should ideally contain identical cell type distributions. Results are obtained by subsampling 10,000 cells with 10 random seeds and reporting the average. Two metrics assess different aspects of this balance:

% \textbf{Batch Mixing ($\mathrm{ASW}_b$):} Measures how well batches are integrated by computing silhouette scores with respect to batch labels:
% \begin{align}
% \mathrm{ASW}_b = 1 - \left|\frac{1}{n} \sum_{i=1}^{n} \frac{b_i^{(b)} - a_i^{(b)}}{\max(a_i^{(b)}, b_i^{(b)})}\right|
% \end{align}
% For each cell $i$ with batch $\beta_i$, $a_i^{(b)}$ is the mean distance to other cells in the same batch, and $b_i^{(b)}$ is the mean distance to cells in the nearest different batch. The absolute value and subtraction from 1 transforms the metric so that high values indicate good batch mixing.

\textbf{Batch-Corrected Biological Preservation ($\mathrm{ASW}_{l/b}$):} Measures whether cell types remain distinguishable within individual batches:
\begin{align}
\mathrm{ASW}_{l/b} = \frac{1}{n} \sum_{i=1}^{n} \frac{b_i^{(l/b)} - a_i^{(l/b)}}{\max(a_i^{(l/b)}, b_i^{(l/b)})}
\end{align}
For each cell $i$ with label $\ell_i$ and batch $\beta_i$, $a_i^{(l/b)}$ is the mean distance to other cells with the same label within the same batch, and $b_i^{(l/b)}$ is the mean distance to cells with different labels but within the same batch. This metric restricts comparisons to within-batch only, testing whether biological structure generalizes across experimental conditions.

\textbf{Principal Component Regression comparison ($\mathrm{PCR_c}$):} Measures batch effect reduction by comparing variance explained by batch labels before and after integration
\footnote{While the original benchmark implementation used a single call to \texttt{pcr()}, which measures absolute remaining batch effects, the paper explicitly describes PCR as comparing ``the proportion of the variance that is explained by the batch variable between the original dataset and the embeddings of the model'', which would instead correspond to \texttt{pcr\_comparison()}.
With either implementation, PCR values are highly variable and inconsistent between models and datasets.
Due to this ambiguity, we opted not to report this metric.
% Therefore, we use \texttt{pcr\_comparison()}, which directly implements this comparative measurement and provides more meaningful relative improvement scores for benchmarking purposes.
}
:
\begin{align}
\mathrm{PCR_c} = \max\left(0, \frac{R^2_{\text{pre}} - R^2_{\text{post}}}{R^2_{\text{pre}}}\right)
\end{align}
where $R^2_{\text{pre}}$ and $R^2_{\text{post}}$ are the variance in the first 50 principal components explained by batch labels in the log-normalized data and embedding spaces, respectively. The $\max(0, \cdot)$ operation ensures the score is non-negative when integration increases batch effects. Values closer to 1 indicate higher batch effect removal.

\subsubsection{Classification probing task}
While not part of the original benchmark by \citet{kedzierska2025zero}, classification probing provides a complementary assessment of representation quality.
Linear probes evaluate whether cell type information is linearly separable in the embedding space, while MLP probes test if this information can be recovered through simple non-linear transformations.
The probe's performance reflects how well the model has organized biologically meaningful structure within its learned representations.

We follow an evaluation protocol with batch-aware data splitting to prevent information leakage.
Data is split at the batch level with 75\% for training and 25\% for testing, ensuring no batch appears in both sets.
Within each training fold, we further reserve 20\% of training data for validation using random sampling (not batch-aware).

For preprocessing, we apply StandardScaler fit on training data and applied to test data.
The training protocol employs early stopping based on validation loss and ReduceLROnPlateau scheduling. 
Model selection is performed through hyperparameter grid search over learning rates $\{1\times10^{-4}, 5\times10^{-4}, 1\times10^{-5}\}$ with fixed weight decay $1\times10^{-4}$, using AdamW optimizer and batch size 2048.
The configuration maximizing the combined metric $(\mathrm{Acc} + \mathrm{F1\,macro})/2$ is selected.

We evaluate two architectures: linear probes and 2-layer MLPs with hidden dimensions $[512, 256]$ and dropout rate $0.5$.
Performance is measured using two complementary metrics:

\textbf{Accuracy ($\mathrm{Acc}$):} 
$$\mathrm{Acc} = \frac{\text{Number of correct predictions}}{\text{Total number of predictions}}$$

\textbf{Macro-averaged F1 Score ($\mathrm{F1\,macro}$):}
$$\mathrm{F1\,macro} = \frac{1}{C} \sum_{c=1}^{C} \mathrm{F1}_c$$
where $\mathrm{F1}_c = \frac{2 \cdot \mathrm{Precision}_c \cdot \mathrm{Recall}_c}{\mathrm{Precision}_c + \mathrm{Recall}_c}$ for class $c$, and $C$ is the number of classes. Unlike accuracy, $\mathrm{F1\,macro}$ provides equal weight to all cell types regardless of frequency, making it particularly suitable for imbalanced single-cell datasets where rare cell types are as important as abundant ones.

\subsubsection{Cell type representation benchmarking, scIB results} \label{sec:msr_results}
\input{tables/msr_benchmark}

\autoref{tab:all-datasets-mean} presents an aggregate assessment of inference-time model performance on cell-type representation learning tasks, revealing a tension between geometric and informational objectives.
While unsupervised clustering rewards models that compress cells into dense, local neighborhoods, this can sometimes come at the cost of global separability.
Our results show that a representation can appear geometrically scattered (lower clustering scores) while still encoding robust, linearly separable cell identities (high probing accuracy), necessitating a holistic view of performance.

\textbf{Aggregate performance and zero-shot significance.}
Navigating this trade-off, TxFM-B trained on TF-Sapiens emerges as the most capable generalist, delivering consistent top-tier performance across both clustering and classification tasks.
This highlights the scaling efficiency of our architecture: when trained on large-scale atlases, TxFM-B outperforms STATE-SE~\cite{Adduri2025STATE}, a model with nearly $4\times$ the learnable parameters trained on a dataset roughly $3\times$ larger.
It is crucial to contextualize these results within the training distributions.
Models trained on large atlases can benefit from partial test leakage, as evaluation datasets like Tabula Sapiens \citep{the2022tabula} are frequently subsets of their self-supervised training corpora.
In contrast, the $\star$ marked versions of TxFM operate in a strict zero-shot setting, yet perform near the top of the ranking.
This underscores that representations learned from curated perturbational data can generalize very effectively.

\textbf{Task-specific inductive biases.}
Some models show performance skewed toward specific metrics, revealing distinct design trade-offs.
scTab~\cite{fischer2024sctab} excels at clustering. Its TabNet backbone \cite{arik2021tabnet} uses sequential attention which, similarly to decision trees, can create sharp partitions ideal for discrete clustering metrics (ASW/NMI/ARI).
However, this can result in a fragmented geometry which is not conducive to strong classification.
AIDO.Cell~\citep{ho2024scaling}, conversely, excels at classification but struggles with clustering, suggesting its embeddings capture rich signal but distribute it with high intra-class variance (e.g., retaining library size noise). Supervised probes can learn to ignore this noise, but unsupervised clustering algorithms cannot.

Ultimately, no single model dominates every metric, but TxFM offers an effective compromise: retaining the local geometry needed for clustering and the invariance required for batch correction, while ensuring cell types remain extractable for downstream classification.

%% file: tables/msr_benchmark_per_dataset.tex
\begin{table*}[t]
  \centering
  \setlength{\tabcolsep}{4.40pt}
  \tiny
  \renewcommand{\arraystretch}{0.99}
  \caption{Per-dataset results for the \citet{kedzierska2025zero} benchmarks in \autoref{tab:all-datasets-mean}.}
  \label{tab:msr_bench_all}

  \begin{tabularx}{\textwidth}{lYYYYYYYY}
     & \multicolumn{3}{c}{\scriptsize\textbf{Clustering}} & \multicolumn{1}{c}{\scriptsize\textbf{Batch Int.}} & \multicolumn{4}{c}{\scriptsize\textbf{Classification}} \\
    \cmidrule(lr){2-4} \cmidrule(lr){5-5} \cmidrule(lr){6-9}
     & \multicolumn{3}{c}{} & \multicolumn{1}{c}{} & \multicolumn{2}{c}{\tiny\textbf{Linear Probe}} & \multicolumn{2}{c}{\tiny\textbf{2-layer MLP}} \\
    \cmidrule(lr){6-7} \cmidrule(lr){8-9}
     & \tiny\textbf{$\mathrm{ASW}\uparrow$} & \tiny\textbf{$\mathrm{NMI}\uparrow$} & \tiny\textbf{$\mathrm{ARI}\uparrow$} & \tiny\textbf{$\mathrm{ASW}_{l/b}\uparrow$} & \tiny\textbf{$\mathrm{Acc}\uparrow$} & \tiny\textbf{$\mathrm{F1}\uparrow$} & \tiny\textbf{$\mathrm{Acc}\uparrow$} & \tiny\textbf{$\mathrm{F1}\uparrow$} \\
    \toprule
    \multicolumn{9}{l}{\footnotesize\textbf{Pan-immune}} \\
    \specialrule{0.3pt}{0pt}{0pt}
    (Lib+Log)Norm+2000HVG & $50.7 \pm 0.1$ & $58.0 \pm 0.6$ & $33.2 \pm 1.0$ & $92.0 \pm 0.2$ & 86.0 & 78.6 & 87.2 & 80.3 \\
    PCA [DiverseRNA-1.4M] & $50.4 \pm 0.1$ & $53.4 \pm 0.4$ & $30.5 \pm 1.5$ & $93.5 \pm 0.2$ & 82.9 & 75.1 & 84.3 & 79.2 \\
    scGPT & $52.7 \pm 0.1$ & $62.8 \pm 0.5$ & $40.6 \pm 0.8$ & $88.3 \pm 0.2$ & 85.1 & 81.7 & 85.9 & 83.0 \\
    AIDO.Cell & $49.5 \pm 0.1$ & $56.8 \pm 0.5$ & $32.0 \pm 0.9$ & $82.6 \pm 0.4$ & 90.3 & 86.4 & 90.7 & 87.6 \\
    scTab & $58.5 \pm 0.2$ & $83.6 \pm 0.7$ & $76.8 \pm 1.7$ & $88.4 \pm 0.4$ & 89.9 & 80.2 & 90.4 & 81.1 \\
    scVI [DiverseRNA-1.4M] & $50.5 \pm 0.2$ & $53.5 \pm 0.4$ & $29.7 \pm 1.7$ & $84.7 \pm 0.4$ & 81.9 & 78.0 & 83.6 & 81.1 \\
    STATE & $50.8 \pm 0.1$ & $60.4 \pm 0.5$ & $34.2 \pm 1.3$ & $96.4 \pm 0.2$ & 88.1 & 81.4 & 88.6 & 83.3 \\
    Transcriptformer & $52.6 \pm 0.1$ & $72.0 \pm 0.5$ & $55.1 \pm 1.6$ & $84.9 \pm 0.4$ & 90.0 & 86.3 & 90.3 & 86.8 \\
    TxFM-B [TF-Sapiens data] & $55.7 \pm 0.2$ & $66.0 \pm 0.5$ & $36.5 \pm 0.9$ & $81.4 \pm 0.3$ & 90.2 & 86.7 & 90.7 & 87.6 \\
    TxFM-S [DiverseRNA-1.4M] & $52.9 \pm 0.2$ & $61.7 \pm 0.7$ & $36.1 \pm 1.9$ & $84.5 \pm 0.3$ & 87.4 & 83.0 & 88.6 & 85.1 \\
    TxFM-B [DiverseRNA-1.4M] & $53.1 \pm 0.1$ & $62.0 \pm 0.6$ & $35.8 \pm 1.6$ & $86.3 \pm 0.4$ & 88.3 & 83.9 & 89.4 & 85.9 \\
    \addlinespace[1.2em]
    \multicolumn{9}{l}{\footnotesize\textbf{Pancreas}} \\
    \specialrule{0.3pt}{0pt}{0pt}
    (Lib+Log)Norm+2000HVG & $55.4 \pm 0.0$ & $75.0 \pm 0.4$ & $56.0 \pm 0.4$ & $93.0 \pm 0.1$ & 71.9 & 60.8 & 75.4 & 70.4 \\
    PCA [DiverseRNA-1.4M] & $49.8 \pm 0.0$ & $39.9 \pm 0.6$ & $17.4 \pm 2.0$ & $83.7 \pm 0.6$ & 95.0 & 85.7 & 94.8 & 83.8 \\
    scGPT & $50.9 \pm 0.2$ & $55.3 \pm 0.6$ & $36.3 \pm 1.9$ & $79.9 \pm 1.4$ & 90.3 & 75.8 & 87.8 & 77.8 \\
    AIDO.Cell & $45.3 \pm 0.3$ & $48.4 \pm 0.8$ & $22.3 \pm 1.8$ & $71.7 \pm 0.5$ & 95.6 & 83.2 & 93.8 & 77.7 \\
    scTab & $51.5 \pm 0.1$ & $54.4 \pm 0.8$ & $50.2 \pm 4.8$ & $90.6 \pm 0.2$ & 75.2 & 50.0 & 78.5 & 60.0 \\
    scVI [DiverseRNA-1.4M] & $45.6 \pm 0.2$ & $41.6 \pm 0.2$ & $17.2 \pm 2.0$ & $70.3 \pm 1.5$ & 73.2 & 61.0 & 76.2 & 67.6 \\
    STATE & $51.6 \pm 0.0$ & $60.7 \pm 0.5$ & $26.8 \pm 0.8$ & $93.4 \pm 0.2$ & 93.4 & 82.4 & 94.3 & 82.6 \\
    Transcriptformer & $48.1 \pm 0.2$ & $56.0 \pm 0.5$ & $34.3 \pm 1.5$ & $73.5 \pm 0.8$ & 75.7 & 57.2 & 73.7 & 52.4 \\
    TxFM-B [TF-Sapiens data] & $51.8 \pm 0.1$ & $50.9 \pm 0.4$ & $25.4 \pm 2.5$ & $75.2 \pm 1.1$ & 89.4 & 82.2 & 89.9 & 83.2 \\
    TxFM-S [DiverseRNA-1.4M] & $51.1 \pm 0.2$ & $50.6 \pm 0.3$ & $22.8 \pm 1.2$ & $80.0 \pm 0.9$ & 90.9 & 71.4 & 91.3 & 72.5 \\
    TxFM-B [DiverseRNA-1.4M] & $51.1 \pm 0.2$ & $50.6 \pm 0.3$ & $22.8 \pm 1.2$ & $80.0 \pm 0.9$ & 90.3 & 73.0 & 90.7 & 72.6 \\
    \addlinespace[1.2em]
    \multicolumn{9}{l}{\footnotesize\textbf{PBMC 12k}} \\
    \specialrule{0.3pt}{0pt}{0pt}
    (Lib+Log)Norm+2000HVG & $51.4 \pm 0.0$ & $69.1 \pm 0.4$ & $59.9 \pm 0.7$ & $99.3 \pm 0.0$ & 94.8 & 90.9 & 95.5 & 92.2 \\
    PCA [DiverseRNA-1.4M] & $51.6 \pm 0.0$ & $68.4 \pm 0.7$ & $61.4 \pm 3.6$ & $99.0 \pm 0.0$ & 93.7 & 87.4 & 94.4 & 88.4 \\
    scGPT & $62.9 \pm 0.1$ & $83.8 \pm 0.8$ & $85.8 \pm 3.7$ & $96.8 \pm 0.2$ & 97.2 & 95.9 & 97.1 & 95.8 \\
    AIDO.Cell & $58.7 \pm 0.1$ & $76.7 \pm 0.9$ & $65.1 \pm 1.6$ & $93.1 \pm 0.2$ & 97.5 & 96.2 & 97.5 & 96.6 \\
    scTab & $57.8 \pm 0.0$ & $79.4 \pm 0.4$ & $82.5 \pm 0.6$ & $98.3 \pm 0.1$ & 95.0 & 90.2 & 95.1 & 91.7 \\
    scVI [DiverseRNA-1.4M] & $59.0 \pm 0.1$ & $74.4 \pm 0.5$ & $66.8 \pm 2.0$ & $96.0 \pm 0.1$ & 95.5 & 91.0 & 96.1 & 92.3 \\
    STATE & $51.2 \pm 0.1$ & $74.4 \pm 0.7$ & $65.0 \pm 1.3$ & $99.2 \pm 0.1$ & 95.7 & 92.2 & 96.0 & 92.3 \\
    Transcriptformer & $61.0 \pm 0.1$ & $79.8 \pm 2.1$ & $74.5 \pm 9.6$ & $94.3 \pm 0.2$ & 97.2 & 96.1 & 97.4 & 96.5 \\
    TxFM-B [TF-Sapiens data] & $59.3 \pm 0.0$ & $86.6 \pm 0.9$ & $89.1 \pm 2.5$ & $96.5 \pm 0.1$ & 97.5 & 96.2 & 97.5 & 96.2 \\
    TxFM-S [DiverseRNA-1.4M] & $66.0 \pm 0.0$ & $87.7 \pm 0.7$ & $92.3 \pm 0.6$ & $96.0 \pm 0.2$ & 97.1 & 95.1 & 97.2 & 95.0 \\
    TxFM-B [DiverseRNA-1.4M] & $66.0 \pm 0.0$ & $87.7 \pm 0.7$ & $92.3 \pm 0.6$ & $96.0 \pm 0.2$ & 97.1 & 94.2 & 97.1 & 95.1 \\
    \addlinespace[1.2em]
    \multicolumn{9}{l}{\footnotesize\textbf{PBMC 68k}} \\
    \specialrule{0.3pt}{0pt}{0pt}
    (Lib+Log)Norm+2000HVG & $48.5 \pm 0.1$ & $46.1 \pm 0.6$ & $27.5 \pm 1.5$ & $96.9 \pm 0.2$ & 78.2 & 68.9 & 82.2 & 73.0 \\
    PCA [DiverseRNA-1.4M] & $49.8 \pm 0.1$ & $43.8 \pm 2.3$ & $21.4 \pm 0.4$ & $96.5 \pm 0.5$ & 65.7 & 55.0 & 67.0 & 54.6 \\
    scGPT & $51.0 \pm 0.2$ & $50.5 \pm 0.6$ & $32.8 \pm 0.9$ & $91.6 \pm 0.7$ & 70.3 & 58.7 & 70.3 & 57.1 \\
    AIDO.Cell & $47.7 \pm 0.2$ & $44.0 \pm 1.2$ & $23.5 \pm 2.9$ & $88.3 \pm 0.7$ & 72.6 & 61.6 & 73.1 & 61.9 \\
    scTab & $50.2 \pm 0.1$ & $49.0 \pm 0.4$ & $19.9 \pm 0.5$ & $94.2 \pm 0.6$ & 68.5 & 56.7 & 69.0 & 56.8 \\
    scVI [DiverseRNA-1.4M] & $49.9 \pm 0.2$ & $41.9 \pm 1.1$ & $22.2 \pm 0.4$ & $91.2 \pm 0.9$ & 68.3 & 56.2 & 68.4 & 57.1 \\
    STATE & $50.1 \pm 0.0$ & $45.8 \pm 1.1$ & $24.4 \pm 1.7$ & $98.6 \pm 0.2$ & 68.0 & 58.3 & 69.0 & 59.3 \\
    Transcriptformer & $49.9 \pm 0.2$ & $49.0 \pm 0.4$ & $21.6 \pm 1.6$ & $90.1 \pm 1.0$ & 73.5 & 62.8 & 72.8 & 61.6 \\
    TxFM-B [TF-Sapiens data] & $50.1 \pm 0.2$ & $49.8 \pm 0.5$ & $28.4 \pm 1.3$ & $89.7 \pm 0.8$ & 74.5 & 64.2 & 74.6 & 63.2 \\
    TxFM-S [DiverseRNA-1.4M] & $50.5 \pm 0.2$ & $50.1 \pm 0.6$ & $24.5 \pm 0.6$ & $91.2 \pm 0.8$ & 70.4 & 58.3 & 70.7 & 59.0 \\
    TxFM-B [DiverseRNA-1.4M] & $50.0 \pm 0.2$ & $46.0 \pm 1.3$ & $23.4 \pm 0.5$ & $92.6 \pm 0.6$ & 69.1 & 58.3 & 69.8 & 57.5 \\
    \addlinespace[1.2em]
    \multicolumn{9}{l}{\footnotesize\textbf{Tabula Sapiens}} \\
    \specialrule{0.3pt}{0pt}{0pt}
    (Lib+Log)Norm+2000HVG & $46.1 \pm 1.0$ & $65.7 \pm 0.5$ & $37.8 \pm 0.9$ & $86.7 \pm 0.6$ & 64.7 & 13.7 & 67.1 & 20.4 \\
    PCA [DiverseRNA-1.4M] & $48.4 \pm 0.2$ & $70.9 \pm 0.4$ & $52.1 \pm 1.1$ & $83.2 \pm 0.5$ & 76.6 & 27.2 & 79.1 & 30.9 \\
    scGPT & $50.2 \pm 0.3$ & $75.1 \pm 0.4$ & $58.8 \pm 0.7$ & $78.3 \pm 0.7$ & 77.2 & 30.5 & 78.3 & 31.4 \\
    AIDO.Cell & $36.8 \pm 0.5$ & $65.0 \pm 0.5$ & $38.4 \pm 1.9$ & $65.7 \pm 1.1$ & 76.4 & 30.0 & 78.2 & 36.6 \\
    scTab & $49.8 \pm 0.4$ & $75.2 \pm 0.3$ & $56.8 \pm 0.5$ & $84.1 \pm 0.7$ & 75.4 & 17.7 & 75.7 & 21.2 \\
    scVI [DiverseRNA-1.4M] & $47.0 \pm 0.4$ & $72.4 \pm 0.4$ & $51.7 \pm 1.1$ & $60.4 \pm 1.4$ & 74.2 & 20.9 & 76.3 & 25.5 \\
    STATE & $50.8 \pm 0.1$ & $78.2 \pm 0.2$ & $59.2 \pm 1.6$ & $89.2 \pm 0.4$ & 80.1 & 38.2 & 81.7 & 40.2 \\
    Transcriptformer & $49.7 \pm 0.3$ & $73.4 \pm 0.2$ & $51.1 \pm 1.4$ & $71.2 \pm 0.9$ & 78.5 & 33.3 & 79.8 & 34.1 \\
    TxFM-B [TF-Sapiens data] & $52.6 \pm 0.2$ & $75.8 \pm 0.4$ & $53.4 \pm 2.1$ & $69.4 \pm 1.0$ & 76.6 & 28.9 & 79.8 & 33.2 \\
    TxFM-S [DiverseRNA-1.4M] & $52.1 \pm 0.3$ & $75.2 \pm 0.5$ & $55.1 \pm 1.1$ & $76.2 \pm 0.9$ & 77.4 & 29.2 & 77.8 & 32.4 \\
    TxFM-B [DiverseRNA-1.4M] & $51.6 \pm 0.2$ & $74.9 \pm 0.3$ & $55.6 \pm 0.7$ & $73.5 \pm 1.0$ & 78.6 & 32.0 & 79.4 & 33.8 \\
  \end{tabularx}

\end{table*}

%% file: tables/msr_benchmark.tex
\begin{table*}[t]
  \centering
  \setlength{\tabcolsep}{4pt}
  \scriptsize
  \renewcommand{\arraystretch}{0.9}
  \caption{\textbf{Cell type representation performance averaged across the five datasets from the~\cite{kedzierska2025zero} benchmark.} We evaluate clustering ($NMI$, $ARI$, $ASW$), batch integration ($ASW_{l/b}$), and classification probing accuracy. $\diamondsuit$ and $\star$ denote, respectively, simple count baseline (no training) and models pretrained without test distribution overlap (zero-shot). 
  For each metric, the best performance is in \textbf{bold}, second best is \underline{underlined}.
  The rank column expresses the average rank of a method across metrics and is used to sort methods from best to worst.
  \autoref{tab:msr_bench_all} shows per-dataset results.}

  \begin{tabularx}{\textwidth}{clYYYYYYYY}
     &  & \multicolumn{3}{c}{\scriptsize\textbf{Clustering}} & \multicolumn{1}{c}{\scriptsize\textbf{Batch Int.}} & \multicolumn{4}{c}{\scriptsize\textbf{Classification}} \\
    \cmidrule(lr){3-5} \cmidrule(lr){6-6} \cmidrule(lr){7-10}
     &  & \multicolumn{3}{c}{} & \multicolumn{1}{c}{} & \multicolumn{2}{c}{\tiny\textbf{Linear Probe}} & \multicolumn{2}{c}{\tiny\textbf{2-layer MLP}} \\
    \cmidrule(lr){7-8} \cmidrule(lr){9-10}
    \tiny\textbf{Rank$\downarrow$} & model & \tiny\textbf{$\mathrm{ASW}\uparrow$} & \tiny\textbf{$\mathrm{NMI}\uparrow$} & \tiny\textbf{$\mathrm{ARI}\uparrow$} & \tiny\textbf{$\mathrm{ASW}_{l/b}\uparrow$} & \tiny\textbf{$\mathrm{Acc}\uparrow$} & \tiny\textbf{$\mathrm{F1}\uparrow$} & \tiny\textbf{$\mathrm{Acc}\uparrow$} & \tiny\textbf{$\mathrm{F1}\uparrow$} \\
    \toprule
3.1 & TxFM-B [TF-Sapiens data] & $\underline{53.9} {\scriptscriptstyle \pm 0.2}$ & $\underline{65.8}      
  {\scriptscriptstyle \pm 0.6}$ & $46.6 {\scriptscriptstyle \pm 1.8}$ & $82.4 {\scriptscriptstyle \pm 0.6}$ & $\underline{85.6}$ & $\mathbf{71.6}$ & $\underline{86.5}$ & $\mathbf{72.7}$ \\                  
      4.4 & STATE-SE & $50.9 {\scriptscriptstyle \pm 0.1}$ & $63.9 {\scriptscriptstyle \pm 0.6}$ & $41.9 {\scriptscriptstyle \pm 1.3}$ & $\mathbf{95.4} {\scriptscriptstyle \pm 0.2}$ & $85.1$ & $70.5$ & $85.9$ & $71.5$ \\       
      4.6 & scGPT & $\underline{53.5} {\scriptscriptstyle \pm 0.2}$ & $\underline{65.5} {\scriptscriptstyle \pm 0.6}$ & $\underline{50.9} {\scriptscriptstyle \pm 1.6}$ & $87.0 {\scriptscriptstyle \pm 0.6}$ & $84.0$ &           
  $68.5$ & $83.9$ & $69.0$ \\                                                                                                                  
      4.7 & TxFM-B [DiverseRNA-1.4M] $\star$ & $\mathbf{54.4} {\scriptscriptstyle \pm 0.1}$ & $64.2 {\scriptscriptstyle \pm 0.7}$ & $46.0 {\scriptscriptstyle \pm 0.9}$ & $85.7 {\scriptscriptstyle \pm 0.6}$ & $84.7$ &           
  $68.3$ & $85.3$ & $69.0$ \\                                                                                                                  
      5.0 & TxFM-S [DiverseRNA-1.4M] $\star$ & $\mathbf{54.5} {\scriptscriptstyle \pm 0.2}$ & $\underline{65.1} {\scriptscriptstyle \pm 0.5}$ & $46.2 {\scriptscriptstyle \pm 1.1}$ & $85.6 {\scriptscriptstyle \pm 0.6}$ &        
  $84.6$ & $67.4$ & $85.1$ & $68.8$ \\                                                                                                         
      6.0 & AIDO.Cell & $47.6 {\scriptscriptstyle \pm 0.3}$ & $58.2 {\scriptscriptstyle \pm 0.8}$ & $36.3 {\scriptscriptstyle \pm 1.9}$ & $80.3 {\scriptscriptstyle \pm 0.6}$ & $\mathbf{86.5}$ & $\underline{71.5}$ &             
  $\mathbf{86.7}$ & $\underline{72.1}$ \\                                                                                                      
      6.2 & scTab & $\underline{53.6} {\scriptscriptstyle \pm 0.2}$ & $\mathbf{68.3} {\scriptscriptstyle \pm 0.5}$ & $\mathbf{57.2} {\scriptscriptstyle \pm 1.6}$ & $91.1 {\scriptscriptstyle \pm 0.4}$ & $80.8$ & $59.0$ &        
  $81.7$ & $62.2$ \\                                                                                                                           
      6.2 & Transcriptformer [TF-Sapiens data] & $52.3 {\scriptscriptstyle \pm 0.1}$ & $\underline{66.0} {\scriptscriptstyle \pm 0.8}$ & $\underline{47.3} {\scriptscriptstyle \pm 3.1}$ & $82.8 {\scriptscriptstyle \pm 0.7}$     
   & $83.0$ & $67.1$ & $82.8$ & $66.3$ \\                                                                                                      
      7.8 & (Lib+Log)Norm+2000HVG $\diamondsuit$ & $50.4 {\scriptscriptstyle \pm 0.2}$ & $62.8 {\scriptscriptstyle \pm 0.5}$ & $42.9 {\scriptscriptstyle \pm 0.9}$ & $\underline{93.6} {\scriptscriptstyle \pm 0.2}$ & $79.1$      
  & $62.6$ & $81.5$ & $67.3$ \\                                                                                                                
      7.9 & PCA [DiverseRNA-1.4M] $\star$ & $50.0 {\scriptscriptstyle \pm 0.1}$ & $55.3 {\scriptscriptstyle \pm 0.9}$ & $36.6 {\scriptscriptstyle \pm 1.8}$ & $91.2 {\scriptscriptstyle \pm 0.4}$ & $82.8$ & $66.1$ & $83.9$ &     
   $67.4$ \\                                                                                                                                   
      9.9 & scVI [DiverseRNA-1.4M] $\star$ & $50.4 {\scriptscriptstyle \pm 0.2}$ & $56.8 {\scriptscriptstyle \pm 0.5}$ & $37.5 {\scriptscriptstyle \pm 1.4}$ & $80.5 {\scriptscriptstyle \pm 0.9}$ & $78.6$ & $61.4$ & $80.1$      
  & $64.7$ \\

  \end{tabularx}
  \label{tab:all-datasets-mean}
\end{table*}

%% file: figures/activations.tex
\begin{figure*}[ht]
    \centering % Center the figure on the page
    
    % --- First Figure ---
    \begin{subfigure}[b]{0.48\textwidth}
        \includegraphics[width=\textwidth]{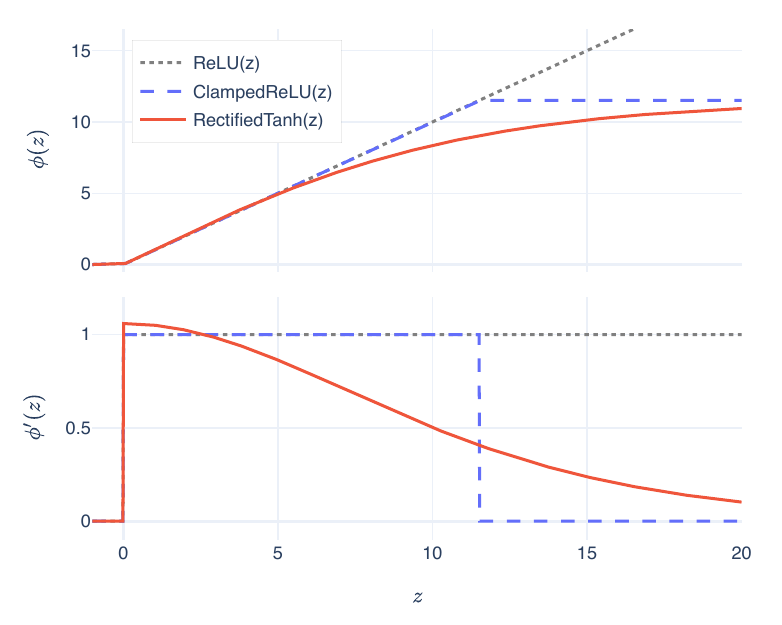}
        \caption{Count activation functions and their derivatives on logits $z$.}
    \end{subfigure}%
    \hfill
    \begin{subfigure}[b]{0.48\textwidth}
        \includegraphics[width=\textwidth]{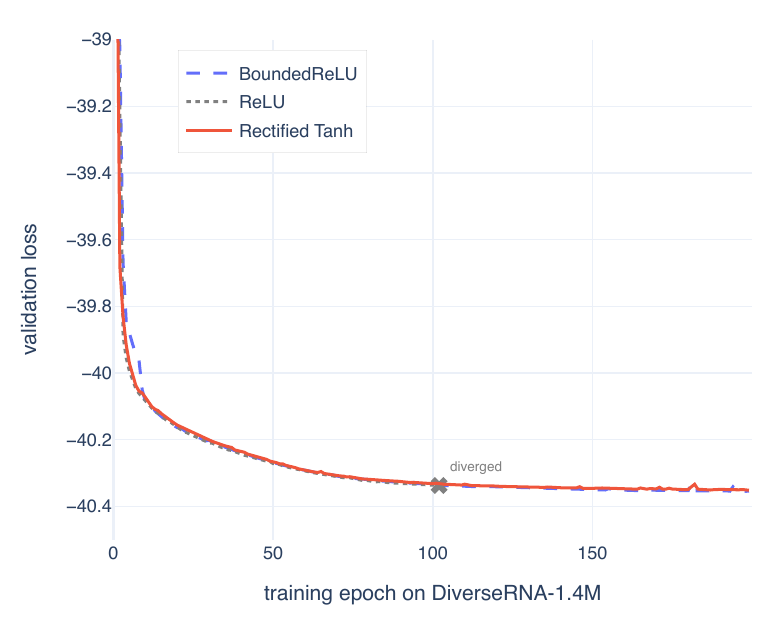}
        \caption{Model validation loss per epoch.}
    \end{subfigure}
    
    \caption{Comparing TxFM decoder count activation functions for models trained in ablation (f). A pure ReLU activation diverges to NaN midway through training. Clamping ReLU prevents divergence and achieves comparable loss, but has lower zero-shot performance in representing perturbations compared to our rectified tanh activation function (\autoref{tab:txfm_ablations}).}
    \label{fig:activations}
\end{figure*}

%% file: tables/appendix_model_choice.tex
\begin{table*}[tb]
\caption{\textbf{Backbone choice for existing models}. Average score across all benchmarking tasks in \cite{bendidi2024benchmarking} on RPE1, HEPG2, and Jurkat datasets for different backbone choices of existing FMs with multiple pretrained backbones. Grayed backbones are the best performing backbones used in \autoref{tab:main_overall_results}.}
\label{tab:existing_models_choice}
\centering
\begingroup\scriptsize  

\setlength{\tabcolsep}{6pt}
\renewcommand{\arraystretch}{1.15}

% ---------- Row 1 ----------
\begin{subtable}[t]{0.47\linewidth}
  \centering
  \begin{tabularx}{\linewidth}{@{}l *{3}{C}@{}}
    \toprule
    Backbone & RPE1 & HEPG2 & Jurkat \\ \midrule
    GenePT ADA & 39.29 & 35.45 & 32.07 \\
    \rowcolor{tablegray} GenePT Large & 39.82 & 34.77 & 32.55 \\
    \bottomrule
  \end{tabularx}
  \subcaption{\textbf{GenePT} ablation of best performing backbone.}
\end{subtable}\hfill
\begin{subtable}[t]{0.47\linewidth}
  \centering
  \begin{tabularx}{\linewidth}{@{}l *{3}{C}@{}}
    \toprule
    Backbone & RPE1 & HEPG2 & Jurkat \\ \midrule
    AIDO.cell 3M & 28.03 & 29.21 & 25.87 \\
    AIDO.cell 10M & 31.01 & 31.84 & 26.47 \\
    \rowcolor{tablegray} AIDO.cell 100M & 37.45 & 34.52 & 31.53 \\
    \bottomrule
  \end{tabularx}
  \subcaption{\textbf{AIDO.cell} ablation of best performing backbone.}
\end{subtable}

\vspace{0.8\baselineskip}

% ---------- Row 2 ----------
\begin{subtable}[t]{0.47\linewidth}
  \centering
  \begin{tabularx}{\linewidth}{@{}l *{3}{C}@{}}
    \toprule
    Backbone & RPE1 & HEPG2 & Jurkat \\ \midrule
    scPrint Small & 25.70 & 22.03 & 24.15 \\
    scPrint Medium & 30.84 & 24.22 & 26.58 \\
    \rowcolor{tablegray} scPrint Large & 31.62 & 23.80 & 27.04 \\
    \bottomrule
  \end{tabularx}
  \subcaption{\textbf{scPrint} ablation of best performing backbone.}
\end{subtable}\hfill
\begin{subtable}[t]{0.47\linewidth}
  \centering
  \begin{tabularx}{\linewidth}{@{}l *{3}{C}@{}}
    \toprule
    Backbone & RPE1 & HEPG2 & Jurkat \\ \midrule
    Transcriptformer Exemplar & 34.53 & 31.54 & 29.96 \\
    Transcriptformer Metazoa & 34.52 & 31.51 & 30.00 \\
    \rowcolor{tablegray} Transcriptformer Sapiens & 34.88 & 31.55 & 29.92 \\
    \bottomrule
  \end{tabularx}
  \subcaption{\textbf{TranscriptFormer} ablation of best performing backbone.}
\end{subtable}
\endgroup         
\end{table*}

%% file: tables/rpe1_main_results.tex
\begin{table*}[tb]
  \centering
  \setlength{\tabcolsep}{8pt}
  \tiny
  % first col left-aligned, then six centered-X (Y) columns
  \caption{Overview of model performance across different settings and perturbational downstream tasks on the scores for \textbf{RPE1} dataset. Metrics are mean $\pm$ std across 5 random seeds.}
  
  \begin{tabularx}{\textwidth}{l|YYYYYYY}
    \toprule
    \textbf{Models}
      & \textbf{ilisi$\uparrow$}
      & \textbf{lin$\uparrow$}
      & \textbf{knn$\uparrow$}
      & \textbf{p.cst$\uparrow$}
      & \textbf{bmdb$\uparrow$}
      & \textbf{inv$\uparrow$}
      & \textbf{avg$\uparrow$}\\
    \midrule
    %––– Subgroup 2 –––
    \multicolumn{1}{l|}{\textbf{Transductive SSL models}} & \multicolumn{7}{l}{} \\
      PCA & 0.63\tiny$\pm$0.00 & 32.03\tiny$\pm$0.05 & 14.54\tiny$\pm$0.09 & 58.79\tiny$\pm$0.14 & 46.03\tiny$\pm$0.00 & 41.17\tiny$\pm$0.71 & 42.64 \\
      scVI & 0.69\tiny$\pm$0.00 & 19.68\tiny$\pm$0.02 & 12.69\tiny$\pm$0.09 & 38.27\tiny$\pm$0.15 & 41.86\tiny$\pm$0.00 & 43.93\tiny$\pm$0.72 & 37.65 \\
      TxFM-B finetuned (ours) & 0.70\tiny$\pm$0.00 & 42.10\tiny$\pm$0.01 & 17.82\tiny$\pm$0.07 & 48.73\tiny$\pm$0.28 & 47.27\tiny$\pm$0.00 & 42.64\tiny$\pm$0.71 & 44.85 \\
    \addlinespace
    \midrule
    %––– Subgroup 1 –––
    \multicolumn{1}{l|}{\textbf{Count data baselines}} & \multicolumn{7}{l}{} \\
      Random label shuffle & 0.70\tiny$\pm$0.00 & 0.85\tiny$\pm$0.01 & 8.20\tiny$\pm$0.01 & 0.00\tiny$\pm$0.00 & 11.17\tiny$\pm$0.00 & 25.34\tiny$\pm$0.80 & 19.40 \\
      Raw data & 0.61\tiny$\pm$0.00 & 36.68\tiny$\pm$0.06 & 13.92\tiny$\pm$0.02 & 36.23\tiny$\pm$0.16 & 44.06\tiny$\pm$0.00 & 59.63\tiny$\pm$0.58 & 41.93 \\
      (Lib+Log)Norm & 0.61\tiny$\pm$0.00 & 30.07\tiny$\pm$0.03 & 14.74\tiny$\pm$0.08 & 37.60\tiny$\pm$0.13 & 42.37\tiny$\pm$0.00 & 62.18\tiny$\pm$0.67 & 41.37 \\
      (Lib+Log)Norm+5k HVG & 0.64\tiny$\pm$0.00 & 26.66\tiny$\pm$0.03 & 14.17\tiny$\pm$0.05 & 41.61\tiny$\pm$0.25 & 43.24\tiny$\pm$0.00 & 52.02\tiny$\pm$0.67 & 40.41 \\
      (Lib+Log)Norm+1024HVG & 0.69\tiny$\pm$0.00 & 23.67\tiny$\pm$0.02 & 14.02\tiny$\pm$0.05 & 36.73\tiny$\pm$0.32 & 43.45\tiny$\pm$0.00 & 41.25\tiny$\pm$0.68 & 38.06 \\
    \addlinespace
    %––– Subgroup 5 –––
    \multicolumn{1}{l}{\textbf{ChatGPT Embeddings}} & \multicolumn{7}{l}{} \\
      GenePT-Large & 0.73\tiny$\pm$0.00 & 27.49\tiny$\pm$0.03 & 12.14\tiny$\pm$0.09 & 43.46\tiny$\pm$0.25 & 44.35\tiny$\pm$0.00 & 37.55\tiny$\pm$0.75 & 39.82 \\
    \addlinespace
    %––– Subgroup 4 –––
    \multicolumn{1}{l|}{\textbf{Pretrained FMs}} & \multicolumn{7}{l}{} \\
      Geneformer [CELLxGENE] & 0.71\tiny$\pm$0.00 & 4.08\tiny$\pm$0.01 & 8.22\tiny$\pm$0.02 & 0.00\tiny$\pm$0.00 & 11.04\tiny$\pm$0.00 & 25.80\tiny$\pm$0.77 & 20.15 \\
      GeneJEPA [Tahoe-100M] & 0.70\tiny$\pm$0.00 & 4.68\tiny$\pm$0.00 & 8.37\tiny$\pm$0.03 & 0.45\tiny$\pm$0.00 & 22.19\tiny$\pm$0.00 & 36.17\tiny$\pm$0.25 & 23.74 \\
      scTab [CELLxGENE] & 0.70\tiny$\pm$0.00 & 6.69\tiny$\pm$0.01 & 8.76\tiny$\pm$0.03 & 1.73\tiny$\pm$0.03 & 31.60\tiny$\pm$0.00 & 35.33\tiny$\pm$0.70 & 25.80 \\
      CellPLM & 0.71\tiny$\pm$0.00 & 7.32\tiny$\pm$0.00 &  9.50\tiny$\pm$0.06 & 2.18\tiny$\pm$0.01 & 23.74\tiny$\pm$0.00 & 36.89\tiny$\pm$0.78 & 25.20 \\
      UCE [CELLxGENE] & 0.70\tiny$\pm$0.00 & 7.42\tiny$\pm$0.00 & 9.34\tiny$\pm$0.05 & 10.24\tiny$\pm$0.10 & 29.36\tiny$\pm$0.00 & 36.74\tiny$\pm$0.73 & 27.34 \\
      scCello [CELLxGENE] & 0.71\tiny$\pm$0.00 & 14.04\tiny$\pm$0.01 & 10.52\tiny$\pm$0.10 & 1.37\tiny$\pm$0.00 & 31.03\tiny$\pm$0.00 & 37.16\tiny$\pm$0.85 & 27.53 \\
      scPrint-L [CELLxGENE] & 0.70\tiny$\pm$0.00 & 16.26\tiny$\pm$0.02 & 11.38\tiny$\pm$0.04 & 16.38\tiny$\pm$0.19 & 37.59\tiny$\pm$0.00 & 37.44\tiny$\pm$0.76 & 31.62 \\
      scGPT [CELLxGENE] & 0.71\tiny$\pm$0.00 & 12.58\tiny$\pm$0.01 & 10.34\tiny$\pm$0.03 & 18.26\tiny$\pm$0.11 & 34.19\tiny$\pm$0.00 & 37.55\tiny$\pm$0.74 & 30.67 \\
      Tahoe-x1 [Tahoe-100M] & 0.71\tiny$\pm$0.00 & 17.16\tiny$\pm$0.00 & 11.59\tiny$\pm$0.01 & 22.64\tiny$\pm$0.39 & 37.23\tiny$\pm$0.00 & 39.16\tiny$\pm$0.24 & 33.15 \\
      TranscriptFormer-Sapiens [CELLxGENE] & 0.71\tiny$\pm$0.00 & 18.65\tiny$\pm$0.01 & 10.90\tiny$\pm$0.05 & 34.22\tiny$\pm$0.19 & 35.08\tiny$\pm$0.00 & 39.40\tiny$\pm$0.72 & 34.88 \\
      AIDO.Cell-100M [CELLxGENE] & 0.71\tiny$\pm$0.00 & 27.66\tiny$\pm$0.02 & 12.88\tiny$\pm$0.02 & 31.62\tiny$\pm$0.40 & 42.61\tiny$\pm$0.00 & 38.54\tiny$\pm$0.71 & 37.45 \\
      Cell2Sentence & 0.70\tiny$\pm$0.00 & 26.73\tiny$\pm$0.00 & 13.10\tiny$\pm$0.00 & 40.09\tiny$\pm$0.20 & 43.38\tiny$\pm$0.00 & 39.83\tiny$\pm$0.29 & 39.01 \\
      TxFM-B [TF-Sapiens data] (ours) & 0.71\tiny$\pm$0.00 & 27.54\tiny$\pm$0.00 & 12.55\tiny$\pm$0.07 & 36.48\tiny$\pm$0.39 & 42.11\tiny$\pm$0.00 & 41.47\tiny$\pm$0.70 & 38.57 \\
      STATE-SE [CxG + Tahoe-100M + scBC] & 0.69\tiny$\pm$0.00 & 29.05\tiny$\pm$0.00 & 13.31\tiny$\pm$0.10 & 40.80\tiny$\pm$0.20 & 44.59\tiny$\pm$0.00 & 44.97\tiny$\pm$0.22 & 40.38 \\
      
    \addlinespace
    %––– Subgroup 3 –––
    \multicolumn{1}{l|}{\textbf{Fit on our public train set}} & \multicolumn{7}{l}{} \\
      PCA [DiverseRNA-1.4M] & 0.69\tiny$\pm$0.00 & 16.10\tiny$\pm$0.01 & 12.65\tiny$\pm$0.08 & 29.59\tiny$\pm$0.29 & 40.01\tiny$\pm$0.00 & 46.38\tiny$\pm$0.66 & 35.72 \\
      scVI [DiverseRNA-1.4M] & 0.71\tiny$\pm$0.00 & 22.10\tiny$\pm$0.01 & 13.08\tiny$\pm$0.05 & 26.25\tiny$\pm$0.12 & 39.07\tiny$\pm$0.00 & 39.52\tiny$\pm$0.77 & 35.18 \\
      
      Linear AE [DiverseRNA-1.4M] & 0.72\tiny$\pm$0.00 & 24.24\tiny$\pm$0.03 & 15.37\tiny$\pm$0.23 & 29.92\tiny$\pm$0.05 & 44.87\tiny$\pm$0.00 & 42.35\tiny$\pm$0.51 & 38.21 \\
      Ridge PCA [DiverseRNA-1.4M] & 0.71\tiny$\pm$0.00 & 28.21\tiny$\pm$0.07 & 16.06\tiny$\pm$0.29 & 34.10\tiny$\pm$0.11 & 45.94\tiny$\pm$0.00 & 41.79\tiny$\pm$0.81 & 39.61 \\
      
      % TxFM-S [DiverseRNA-1.4M] (ours) & 0.71\tiny$\pm$0.00 & 35.11\tiny$\pm$0.01 & 16.04\tiny$\pm$0.04 & 42.10\tiny$\pm$0.62 & 46.21\tiny$\pm$0.00 & 41.06\tiny$\pm$0.30 & 41.94 \\
      TxFM-B [DiverseRNA-1.4M] (ours) & 0.70\tiny$\pm$0.00 & 37.61\tiny$\pm$0.02 & 15.13\tiny$\pm$0.04 & 42.71\tiny$\pm$0.61 & 45.33\tiny$\pm$0.00 & 41.54\tiny$\pm$0.73 & 42.17 \\
    \bottomrule
  \end{tabularx}
  \label{tab:main_rpe1_results}
\end{table*}

%% file: tables/hepg2_main_results.tex
\begin{table*}[tb]
  \centering
  \setlength{\tabcolsep}{8pt}
  \tiny
  \caption{Overview of model performance across different settings and perturbational downstream tasks on the scores for \textbf{HEPG2} dataset. Metrics are mean $\pm$ std across 5 random seeds.}
  
  % first col left-aligned, then six centered-X (Y) columns
  \begin{tabularx}{\textwidth}{l|YYYYYYY}
    \toprule
    \textbf{Models}
      & \textbf{ilisi$\uparrow$}
      & \textbf{lin$\uparrow$}
      & \textbf{knn$\uparrow$}
      & \textbf{p.cst$\uparrow$}
      & \textbf{bmdb$\uparrow$}
      & \textbf{inv$\uparrow$}
      & \textbf{avg$\uparrow$}\\
    \midrule
    %––– Subgroup 2 –––
    \multicolumn{1}{l|}{\textbf{Transductive SSL models}} & \multicolumn{7}{l}{} \\
      PCA & 0.63\tiny$\pm$0.00 & 23.90\tiny$\pm$0.09 & 12.81\tiny$\pm$0.05 & 43.56\tiny$\pm$0.10 & 45.57\tiny$\pm$0.00 & 42.54\tiny$\pm$1.57 & 38.63 \\
      scVI & 0.69\tiny$\pm$0.00 & 12.61\tiny$\pm$0.04 & 11.47\tiny$\pm$0.14 & 27.84\tiny$\pm$0.07 & 45.22\tiny$\pm$0.00 & 47.03\tiny$\pm$2.27 & 35.63 \\
      TxFM-B finetuned (ours) & 0.70\tiny$\pm$0.00 & 29.27\tiny$\pm$0.04 & 15.81\tiny$\pm$0.05 & 38.40\tiny$\pm$0.19 & 46.24\tiny$\pm$0.00 & 44.55\tiny$\pm$3.18 & 40.78 \\
    \addlinespace
    \midrule
    %––– Subgroup 1 –––
    \multicolumn{1}{l|}{\textbf{Count data baselines}} & \multicolumn{7}{l}{} \\
      Random label shuffle & 0.70\tiny$\pm$0.00 & 0.66\tiny$\pm$0.01 & 5.45\tiny$\pm$0.05 & 0.00\tiny$\pm$0.00 & 12.95\tiny$\pm$0.00 & 28.14\tiny$\pm$2.58 & 19.68 \\
      Raw data & 0.60\tiny$\pm$0.00 & 23.21\tiny$\pm$0.74 & 9.75\tiny$\pm$0.06 & 2.60\tiny$\pm$0.12 & 32.15\tiny$\pm$0.00 & 56.38\tiny$\pm$2.01 & 30.81 \\
      (Lib+Log)Norm & 0.65\tiny$\pm$0.00 & 21.51\tiny$\pm$0.05 & 11.93\tiny$\pm$0.05 & 10.49\tiny$\pm$0.22 & 35.32\tiny$\pm$0.00 & 60.09\tiny$\pm$1.68 & 34.12 \\
      (Lib+Log)Norm+5k HVG & 0.67\tiny$\pm$0.00 & 19.22\tiny$\pm$0.05 & 11.38\tiny$\pm$0.05 & 8.09\tiny$\pm$0.13 & 34.75\tiny$\pm$0.00 & 50.54\tiny$\pm$2.00 & 31.88 \\
      (Lib+Log)Norm+1024HVG & 0.69\tiny$\pm$0.00 & 15.04\tiny$\pm$0.04 & 10.74\tiny$\pm$0.03 & 24.13\tiny$\pm$0.08 & 44.06\tiny$\pm$0.00 & 44.46\tiny$\pm$2.16 & 34.65 \\
    \addlinespace
    \multicolumn{1}{l}{\textbf{ChatGPT Embeddings}} & \multicolumn{7}{l}{} \\
      GenePT-Large & 0.73\tiny$\pm$0.00 & 18.65\tiny$\pm$0.06 & 8.28\tiny$\pm$0.03 & 24.78\tiny$\pm$0.05 & 42.01\tiny$\pm$0.00 & 40.91\tiny$\pm$2.40 & 34.77 \\
    \addlinespace    
    %––– Subgroup 4 –––
    \multicolumn{1}{l|}{\textbf{Pretrained FMs}} & \multicolumn{7}{l}{} \\
      Geneformer [CELLxGENE] & 0.71\tiny$\pm$0.00 & 2.21\tiny$\pm$0.00 & 5.44\tiny$\pm$0.02 & 0.04\tiny$\pm$0.00 & 11.30\tiny$\pm$0.00 & 28.41\tiny$\pm$2.45 & 19.89 \\
      GeneJEPA [Tahoe-100M] & 0.70\tiny$\pm$0.00 & 2.64\tiny$\pm$0.00 & 5.74\tiny$\pm$0.00 & 0.41\tiny$\pm$0.00 & 23.68\tiny$\pm$0.00 & 38.86\tiny$\pm$1.82 & 23.64 \\
      scTab [CELLxGENE] & 0.70\tiny$\pm$0.00 & 4.59\tiny$\pm$0.02 & 6.67\tiny$\pm$0.02 & 1.93\tiny$\pm$0.03 & 32.89\tiny$\pm$0.00 & 38.80\tiny$\pm$2.37 & 25.91 \\
      CellPLM & 0.71\tiny$\pm$0.00 & 5.12\tiny$\pm$0.02 & 7.52\tiny$\pm$0.03 & 3.68\tiny$\pm$0.07 & 23.03\tiny$\pm$0.00 & 40.68\tiny$\pm$2.29 & 25.25 \\
      UCE [CELLxGENE] & 0.70\tiny$\pm$0.00 & 5.23\tiny$\pm$0.01 & 7.20\tiny$\pm$0.03 & 6.31\tiny$\pm$0.01 & 31.32\tiny$\pm$0.00 & 40.47\tiny$\pm$2.31 & 26.92 \\
      scCello [CELLxGENE] & 0.71\tiny$\pm$0.00 & 8.50\tiny$\pm$0.03 & 7.44\tiny$\pm$0.03 & 6.95\tiny$\pm$0.03 & 36.77\tiny$\pm$0.00 & 40.14\tiny$\pm$2.38 & 28.50 \\
      scPrint-L [CELLxGENE] & 0.71\tiny$\pm$0.00 & 2.96\tiny$\pm$0.00 & 5.93\tiny$\pm$0.03 & 0.16\tiny$\pm$0.00 & 26.89\tiny$\pm$0.00 & 35.57\tiny$\pm$2.47 & 23.80 \\
      scGPT [CELLxGENE] & 0.70\tiny$\pm$0.00 & 7.21\tiny$\pm$0.01 & 8.51\tiny$\pm$0.04 & 11.46\tiny$\pm$0.03 & 36.69\tiny$\pm$0.00 & 41.30\tiny$\pm$2.27 & 29.35 \\
      Tahoe-x1 [Tahoe-100M] & 0.70\tiny$\pm$0.00 & 9.53\tiny$\pm$0.02 & 9.83\tiny$\pm$0.05 & 15.39\tiny$\pm$0.04 & 39.30\tiny$\pm$3.28 & 43.25\tiny$\pm$1.51 & 31.36 \\
      TranscriptFormer-Sapiens [CELLxGENE] & 0.70\tiny$\pm$0.00 & 12.16\tiny$\pm$0.02 & 8.94\tiny$\pm$0.04 & 19.33\tiny$\pm$0.03 & 35.14\tiny$\pm$0.00 & 42.77\tiny$\pm$2.26 & 31.55 \\
      AIDO.Cell-100M [CELLxGENE] & 0.71\tiny$\pm$0.00 & 16.98\tiny$\pm$0.03 & 10.26\tiny$\pm$0.04 & 25.05\tiny$\pm$0.05 & 41.42\tiny$\pm$0.00 & 42.28\tiny$\pm$2.27 & 34.52 \\
      Cell2Sentence & 0.70\tiny$\pm$0.00 & 16.12\tiny$\pm$0.03 & 9.99\tiny$\pm$0.08 & 24.52\tiny$\pm$0.00 & 42.24\tiny$\pm$0.00 & 43.44\tiny$\pm$1.85 & 34.49 \\
      TxFM-B [TF-Sapiens data] (ours) & 0.70\tiny$\pm$0.00 & 17.81\tiny$\pm$0.04 & 10.59\tiny$\pm$0.08 & 25.28\tiny$\pm$0.19 & 43.31\tiny$\pm$0.00 & 43.87\tiny$\pm$2.48 & 35.30 \\
      STATE-SE [CxG + Tahoe-100M + scBC] & 0.69\tiny$\pm$0.00 & 17.20\tiny$\pm$0.00 & 10.79\tiny$\pm$0.05 & 23.13\tiny$\pm$0.08 & 43.94\tiny$\pm$0.00 & 45.15\tiny$\pm$1.77 & 35.10 \\
    \addlinespace
    \multicolumn{1}{l|}{\textbf{Fit on our public train set}} & \multicolumn{7}{l}{} \\
      PCA [DiverseRNA-1.4M] & 0.69\tiny$\pm$0.00 & 11.11\tiny$\pm$0.05 & 11.50\tiny$\pm$0.06 & 20.53\tiny$\pm$0.13 & 43.34\tiny$\pm$0.00 & 48.21\tiny$\pm$2.10 & 34.00 \\
      scVI [DiverseRNA-1.4M] & 0.70\tiny$\pm$0.00 & 14.16\tiny$\pm$0.03 & 10.83\tiny$\pm$0.00 & 17.28\tiny$\pm$0.05 & 40.00\tiny$\pm$0.00 & 43.17\tiny$\pm$2.25 & 32.69 \\

      Linear AE [DiverseRNA-1.4M] & 0.72\tiny$\pm$0.00 & 16.59\tiny$\pm$0.07 & 11.82\tiny$\pm$0.03 & 20.45\tiny$\pm$0.31 & 45.74\tiny$\pm$0.00 & 45.63\tiny$\pm$0.16 & 35.44 \\
      Ridge PCA [DiverseRNA-1.4M] & 0.71\tiny$\pm$0.00 & 20.23\tiny$\pm$0.09 & 12.43\tiny$\pm$0.11 & 22.87\tiny$\pm$0.22 & 46.92\tiny$\pm$0.00 & 46.16\tiny$\pm$0.24 & 36.76 \\
      
      % TxFM-S [DiverseRNA-1.4M] (ours) & 0.70\tiny$\pm$0.00 & 23.08\tiny$\pm$0.07 & 13.11\tiny$\pm$0.05 & 29.52\tiny$\pm$0.02 & 45.88\tiny$\pm$0.00 & 44.56\tiny$\pm$1.88 & 37.82 \\
      TxFM-B [DiverseRNA-1.4M] (ours) & 0.70\tiny$\pm$0.00 & 25.93\tiny$\pm$0.07 & 13.14\tiny$\pm$0.06 & 32.47\tiny$\pm$0.21 & 45.26\tiny$\pm$0.00 & 44.40\tiny$\pm$2.21 & 38.63 \\
    \bottomrule
  \end{tabularx}
  % scCello was trained on cell type ontologies, so its biased for cell type related tasks
  \label{tab:main_hepg2_results}
\end{table*}

%% file: tables/jurkat_main_results.tex
\begin{table*}[tb]
  \centering
  \setlength{\tabcolsep}{8pt}
  \tiny
  \caption{Overview of model performance across different settings and perturbational downstream tasks on the scores for \textbf{Jurkat} dataset. Metrics are mean $\pm$ std across 5 random seeds.}
  
  % first col left-aligned, then six centered-X (Y) columns
  \begin{tabularx}{\textwidth}{l|YYYYYYY}
    \toprule
    \textbf{Models}
      & \textbf{ilisi$\uparrow$}
      & \textbf{lin$\uparrow$}
      & \textbf{knn$\uparrow$}
      & \textbf{p.cst$\uparrow$}
      & \textbf{bmdb$\uparrow$}
      & \textbf{inv$\uparrow$}
      & \textbf{avg$\uparrow$}\\
    \midrule
    %––– Subgroup 2 –––
    \multicolumn{1}{l|}{\textbf{Transductive SSL models}} & \multicolumn{7}{l}{} \\
      PCA & 0.66\tiny$\pm$0.00 & 15.59\tiny$\pm$0.02 & 10.37\tiny$\pm$0.03 & 36.27\tiny$\pm$0.10 & 41.13\tiny$\pm$0.00 & 39.70\tiny$\pm$0.11 & 34.89 \\
      scVI & 0.69\tiny$\pm$0.00 & 9.95\tiny$\pm$0.01 & 9.92\tiny$\pm$0.08 & 23.26\tiny$\pm$0.10 & 40.50\tiny$\pm$0.00 & 42.39\tiny$\pm$0.07 & 32.62 \\
      TxFM-B finetuned (ours) & 0.69\tiny$\pm$0.00 & 25.64\tiny$\pm$0.03 & 11.99\tiny$\pm$0.01 & 36.56\tiny$\pm$0.10 & 42.64\tiny$\pm$0.00 & 41.52\tiny$\pm$0.10 & 38.01 \\
    \addlinespace
    \midrule
    %––– Subgroup 1 –––
    \multicolumn{1}{l|}{\textbf{Count data baselines}} & \multicolumn{7}{l}{} \\
      Random label shuffle & 0.70\tiny$\pm$0.00 & 0.54\tiny$\pm$0.00 & 7.46\tiny$\pm$0.04 & 0.04\tiny$\pm$0.00 & 11.59\tiny$\pm$0.00 & 24.77\tiny$\pm$0.08 & 19.20 \\
      Raw data & 0.63\tiny$\pm$0.00 & 20.47\tiny$\pm$0.05 & 10.36\tiny$\pm$0.04 & 11.55\tiny$\pm$0.02 & 35.25\tiny$\pm$0.00 & 57.28\tiny$\pm$0.14 & 33.06 \\
      (Lib+Log)Norm & 0.62\tiny$\pm$0.00 & 16.73\tiny$\pm$0.02 & 11.07\tiny$\pm$0.04 & 14.69\tiny$\pm$0.07 & 34.71\tiny$\pm$0.00 & 60.50\tiny$\pm$0.19 & 33.37 \\
      (Lib+Log)Norm+5k HVG & 0.64\tiny$\pm$0.00 & 13.07\tiny$\pm$0.01 & 10.76\tiny$\pm$0.01 & 20.95\tiny$\pm$0.06 & 37.29\tiny$\pm$0.00 & 50.61\tiny$\pm$0.13 & 32.94 \\
      (Lib+Log)Norm+1024HVG & 0.68\tiny$\pm$0.00 & 9.67\tiny$\pm$0.01 & 10.38\tiny$\pm$0.03 & 16.95\tiny$\pm$0.10 & 38.68\tiny$\pm$0.00 & 39.44\tiny$\pm$0.09 & 30.67 \\
    \addlinespace    
    %––– Subgroup 5 –––
    \multicolumn{1}{l}{\textbf{ChatGPT Embeddings}} & \multicolumn{7}{l}{} \\
      GenePT-Large & 0.74\tiny$\pm$0.00 & 14.30\tiny$\pm$0.02 & 8.58\tiny$\pm$0.01 & 24.47\tiny$\pm$0.16 & 38.11\tiny$\pm$0.00 & 35.74\tiny$\pm$0.48 & 32.55 \\
    \addlinespace
    %––– Subgroup 4 –––
    \multicolumn{1}{l|}{\textbf{Pretrained FMs}} & \multicolumn{7}{l}{} \\
      Geneformer [CELLxGENE] & 0.71\tiny$\pm$0.00 & 2.97\tiny$\pm$0.00 & 7.44\tiny$\pm$0.03 & 0.00\tiny$\pm$0.00 & 10.60\tiny$\pm$0.00 & 25.00\tiny$\pm$0.32 & 19.63 \\
      GeneJEPA [Tahoe-100M] & 0.70\tiny$\pm$0.00 & 3.14\tiny$\pm$0.01 & 7.59\tiny$\pm$0.04 & 0.04\tiny$\pm$0.00 & 14.96\tiny$\pm$0.00 & 33.56\tiny$\pm$0.21 & 21.66 \\
      scTab [CELLxGENE] & 0.70\tiny$\pm$0.00 & 3.70\tiny$\pm$0.01 & 7.72\tiny$\pm$0.04 & 0.12\tiny$\pm$0.00 & 21.12\tiny$\pm$0.00 & 32.79\tiny$\pm$0.14 & 22.69 \\
      CellPLM & 0.71\tiny$\pm$0.00 & 4.27\tiny$\pm$0.00 & 8.19\tiny$\pm$0.03 & 1.79\tiny$\pm$0.07 & 23.99\tiny$\pm$0.00 & 34.81\tiny$\pm$0.17 & 24.06 \\
      UCE [CELLxGENE] & 0.71\tiny$\pm$0.00 & 4.97\tiny$\pm$0.00 & 7.97\tiny$\pm$0.02 & 2.95\tiny$\pm$0.01 & 28.61\tiny$\pm$0.00 & 34.87\tiny$\pm$0.16 & 25.07 \\
      scCello [CELLxGENE] & 0.70\tiny$\pm$0.00 & 7.26\tiny$\pm$0.01 & 8.69\tiny$\pm$0.04 & 0.37\tiny$\pm$0.00 & 28.45\tiny$\pm$0.00 & 34.77\tiny$\pm$0.08 & 25.09 \\
      scPrint-L [CELLxGENE] & 0.70\tiny$\pm$0.00 & 8.43\tiny$\pm$0.01 & 9.24\tiny$\pm$.01 & 5.13\tiny$\pm$0.06 & 33.34\tiny$\pm$0.00 & 35.48\tiny$\pm$0.13 & 27.04 \\
      scGPT [CELLxGENE] & 0.71\tiny$\pm$0.00 & 7.66\tiny$\pm$0.01 & 8.61\tiny$\pm$0.05 & 5.53\tiny$\pm$0.17 & 32.45\tiny$\pm$0.00 & 35.70\tiny$\pm$0.12 & 26.86 \\
      Tahoe-x1 [Tahoe-100M] & 0.71\tiny$\pm$0.00 & 9.61\tiny$\pm$0.01 & 9.40\tiny$\pm$0.00 & 10.15\tiny$\pm$0.00 & 34.02\tiny$\pm$0.00 & 36.37\tiny$\pm$0.15 & 28.47 \\
      TranscriptFormer-Sapiens [CELLxGENE] & 0.70\tiny$\pm$0.00 & 10.77\tiny$\pm$0.01 & 8.68\tiny$\pm$0.04 & 20.40\tiny$\pm$0.04 & 31.38\tiny$\pm$0.00 & 37.30\tiny$\pm$0.14 & 29.92 \\
      AIDO.Cell-100M [CELLxGENE] & 0.71\tiny$\pm$0.00 & 15.07\tiny$\pm$0.00 & 9.46\tiny$\pm$.05 & 19.54\tiny$\pm$0.07 & 37.08\tiny$\pm$0.00 & 36.69\tiny$\pm$0.12 & 31.53 \\
      Cell2Sentence & 0.71\tiny$\pm$0.00 & 13.93\tiny$\pm$0.00 & 9.39\tiny$\pm$0.04 & 26.21\tiny$\pm$0.00 & 39.11\tiny$\pm$0.00 & 37.21\tiny$\pm$0.08 & 32.82 \\
      TxFM-B [TF-Sapiens data] (ours) & 0.71\tiny$\pm$0.00 & 16.52\tiny$\pm$0.01 & 9.82\tiny$\pm$0.05 & 22.59\tiny$\pm$0.13 & 37.97\tiny$\pm$0.00 & 39.77\tiny$\pm$0.11 & 32.98 \\
      STATE-SE [CxG + Tahoe-100M + scBC] & 0.69\tiny$\pm$0.00 & 17.15\tiny$\pm$0.00 & 9.92\tiny$\pm$0.04 & 24.45\tiny$\pm$0.00 & 39.44\tiny$\pm$0.00 & 42.65\tiny$\pm$0.01 & 33.81 \\
    \addlinespace
    \multicolumn{1}{l|}{\textbf{Fit on our public train set}} & \multicolumn{7}{l}{} \\
      PCA [DiverseRNA-1.4M] & 0.69\tiny$\pm$0.00 & 7.60\tiny$\pm$0.00 & 10.04\tiny$\pm$0.05 & 16.83\tiny$\pm$0.03 & 37.44\tiny$\pm$0.00 & 44.56\tiny$\pm$0.09 & 30.91 \\
      scVI [DiverseRNA-1.4M] & 0.71\tiny$\pm$0.00 & 12.97\tiny$\pm$0.01 & 10.43\tiny$\pm$0.02 & 16.18\tiny$\pm$0.09 & 36.16\tiny$\pm$0.00 & 37.36\tiny$\pm$0.11 & 30.68 \\

      Linear AE [DiverseRNA-1.4M] & 0.71\tiny$\pm$0.00 & 14.13\tiny$\pm$0.05 & 11.13\tiny$\pm$0.14 & 17.45\tiny$\pm$0.25 & 40.18\tiny$\pm$0.00 & 40.23\tiny$\pm$0.36 & 32.49 \\
      
      Ridge PCA [DiverseRNA-1.4M] & 0.71\tiny$\pm$0.00 & 16.53\tiny$\pm$0.09 & 11.94\tiny$\pm$0.11 & 20.84\tiny$\pm$0.19 & 42.08\tiny$\pm$0.00 & 40.26\tiny$\pm$0.53 & 33.85 \\

      % TxFM-S [DiverseRNA-1.4M] (ours) & 0.70\tiny$\pm$0.00 & 22.18\tiny$\pm$0.00 & 11.60\tiny$\pm$0.04 & 28.51\tiny$\pm$0.11 & 42.56\tiny$\pm$0.00 & 38.89\tiny$\pm$0.02 & 35.75 \\
      TxFM-B [DiverseRNA-1.4M] (ours) & 0.69\tiny$\pm$0.00 & 23.79\tiny$\pm$0.01 & 11.59\tiny$\pm$0.03 & 31.27\tiny$\pm$0.17 & 42.08\tiny$\pm$0.00 & 40.42\tiny$\pm$0.11 & 36.52 \\
    \bottomrule
  \end{tabularx}
  \label{tab:main_jurkat_results}
\end{table*}

%% file: figures/pca_ablation.tex
\begin{figure*}
    \centering
    \includegraphics[width=\linewidth]{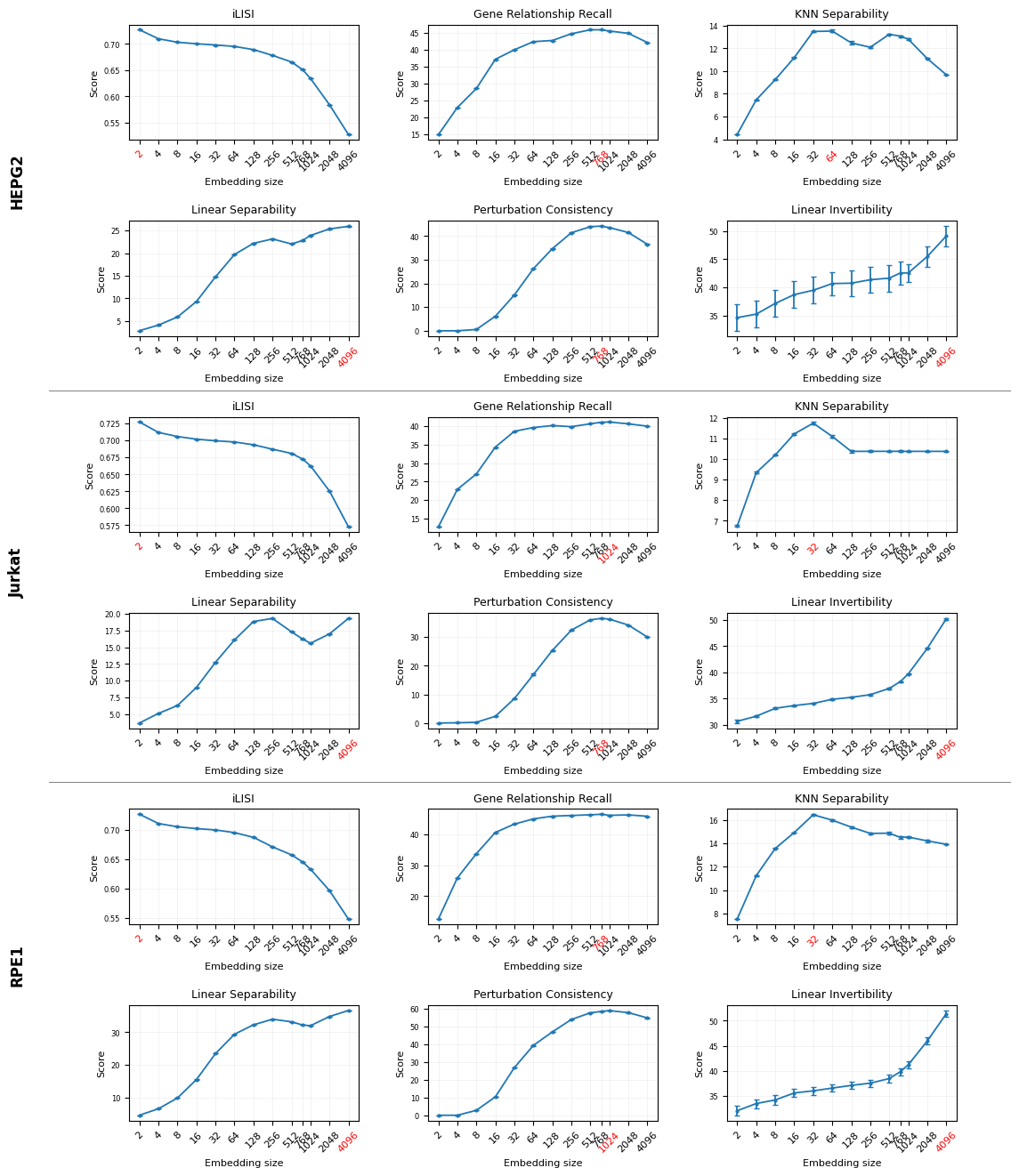}
    \caption{\textbf{Dimension size ablation for PCA.} \cite{bendidi2024benchmarking} benchmarking results on RPE1, HEPG2, and Jurkat datasets after applying PCA on raw counts with different numbers of principal components. Red is optimal size for each downstream task.}
    \label{fig:pca_ablation}
\end{figure*}

%% file: checklist.tex
\newpage
\section{NeurIPS Paper Checklist}

\begin{enumerate}

\item {\bf Claims}
    \item[] Question: Do the main claims made in the abstract and introduction accurately reflect the paper's contributions and scope?
    \item[] Answer: \answerYes{} % Replace by \answerYes{}, \answerNo{}, or \answerNA{}.
    \item[] Justification: The benchmarking analysis (\S~\ref{sec:perts}, \S~\ref{sec:bmdb}) and comprehensive ablation studies (\S~\ref{sec:ablations})---representing 10,000s of hours of H100 GPU training and inference compute, comprising many training evaluations along different data and architecture training splits---accurately reflect the paper's claims (state-of-the-art inductive gene perturbation representation learning and gene relationship recall from model parameters) as being the result of careful SSL dataset curation and a novel SSL model architecture design (a transcriptomics MAE with transformer encoder and MLP decoder off CLS token).

\item {\bf Limitations}
    \item[] Question: Does the paper discuss the limitations of the work performed by the authors?
    \item[] Answer: \answerYes{} % Replace by \answerYes{}, \answerNo{}, or \answerNA{}.
    \item[] Justification: The limitations section appears immediately after the conclusion in \S~\ref{sec:conclusion}.

\item {\bf Theory assumptions and proofs}
    \item[] Question: For each theoretical result, does the paper provide the full set of assumptions and a complete (and correct) proof?
    \item[] Answer: \answerNA{} % Replace by \answerYes{}, \answerNo{}, or \answerNA{}.
    \item[] Justification: This paper is empirical in nature.

    \item {\bf Experimental result reproducibility}
    \item[] Question: Does the paper fully disclose all the information needed to reproduce the main experimental results of the paper to the extent that it affects the main claims and/or conclusions of the paper (regardless of whether the code and data are provided or not)?
    \item[] Answer: \answerYes{} % Replace by \answerYes{}, \answerNo{}, or \answerNA{}.
    \item[] Justification: The training dataset is curated from fully public data sources and the evaluation code is from previously published work: \cite{bendidi2024benchmarking} for perturbation evaluation, and \cite{celik2024building} for gene relationship evaluation. The new TxFM model architecture is fully described and detailed. We also commit to releasing additional code resources upon acceptance.

\item {\bf Open access to data and code}
    \item[] Question: Does the paper provide open access to the data and code, with sufficient instructions to faithfully reproduce the main experimental results, as described in supplemental material?
    \item[] Answer: \answerYes{} % Replace by \answerYes{}, \answerNo{}, or \answerNA{}.
    \item[] Justification: As described above, training data is public, evaluation data is public, and evaluation code is also public. 

\item {\bf Experimental setting/details}
    \item[] Question: Does the paper specify all the training and test details (e.g., data splits, hyperparameters, how they were chosen, type of optimizer) necessary to understand the results?
    \item[] Answer: \answerYes{} % Replace by \answerYes{}, \answerNo{}, or \answerNA{}.
    \item[] Justification: This is fully described in both the paper text and in the appendix throughout.

\item {\bf Experiment statistical significance}
    \item[] Question: Does the paper report error bars suitably and correctly defined or other appropriate information about the statistical significance of the experiments?
    \item[] Answer: \answerYes{} % Replace by \answerYes{}, \answerNo{}, or \answerNA{}.
    \item[] Justification: Error bars are included in primary results either in the appendix or main text (e.g., \autoref{tab:bmdb}).

\item {\bf Experiments compute resources}
    \item[] Question: For each experiment, does the paper provide sufficient information on the computer resources (type of compute workers, memory, time of execution) needed to reproduce the experiments?
    \item[] Answer: \answerYes{} % Replace by \answerYes{}, \answerNo{}, or \answerNA{}.
    \item[] Justification: Each newly trained model is clearly documented with its compute requirements, see \S~\ref{sec:appendix_hyperparameters}.
    
\item {\bf Code of ethics}
    \item[] Question: Does the research conducted in the paper conform, in every respect, with the NeurIPS Code of Ethics \url{https://neurips.cc/public/EthicsGuidelines}?
    \item[] Answer: \answerYes{} % Replace by \answerYes{}, \answerNo{}, or \answerNA{}.
    \item[] Justification: The datasets and models in this work are used correctly and documented thoroughly throughout the work. No ethical dilemmas are present in this work.

\item {\bf Broader impacts}
    \item[] Question: Does the paper discuss both potential positive societal impacts and negative societal impacts of the work performed?
    \item[] Answer: \answerYes{} % Replace by \answerYes{}, \answerNo{}, or \answerNA{}.
    \item[] Justification: This work may positively impact biomedical research by improving representation learning for transcriptomic data, enabling more effective analysis of perturbational screens, gene-gene relationships, and potential discovery of disease mechanisms. These advances could support target discovery and more efficient biological hypothesis generation. Potential negative impacts are limited but include overinterpretation of learned gene relationships as causal mechanisms, biased conclusions from training data skewed toward oncology and public atlas datasets, and inappropriate use of representations in clinical or therapeutic decision-making without validation. We mitigate these risks by evaluating only on public research datasets, clearly framing the model as a research tool rather than a clinical system, reporting limitations around dataset bias and generalization, and planning to release code/checkpoints to support reproducibility and independent evaluation.
    
\item {\bf Safeguards}
    \item[] Question: Does the paper describe safeguards that have been put in place for responsible release of data or models that have a high risk for misuse (e.g., pre-trained language models, image generators, or scraped datasets)?
    \item[] Answer: \answerNA{} % Replace by \answerYes{}, \answerNo{}, or \answerNA{}.
    \item[] Justification: The data and many transcriptomics models are already publicly available.

\item {\bf Licenses for existing assets}
    \item[] Question: Are the creators or original owners of assets (e.g., code, data, models), used in the paper, properly credited and are the license and terms of use explicitly mentioned and properly respected?
    \item[] Answer: \answerYes{} % Replace by \answerYes{}, \answerNo{}, or \answerNA{}.
    \item[] Justification: We cite each data source used in this work, and all are already available on \url{https://cellxgene.cziscience.com/datasets}

\item {\bf New assets}
    \item[] Question: Are new assets introduced in the paper well documented and is the documentation provided alongside the assets?
    \item[] Answer: \answerYes{} % Replace by \answerYes{}, \answerNo{}, or \answerNA{}.
    \item[] Justification: We will release code / model checkpoint upon publication with clear and appropriate documentation.

\item {\bf Crowdsourcing and research with human subjects}
    \item[] Question: For crowdsourcing experiments and research with human subjects, does the paper include the full text of instructions given to participants and screenshots, if applicable, as well as details about compensation (if any)? 
    \item[] Answer: \answerNA{} % Replace by \answerYes{}, \answerNo{}, or \answerNA{}.
    \item[] Justification: The paper does not involve crowdsourcing nor research with human subjects.

\item {\bf Institutional review board (IRB) approvals or equivalent for research with human subjects}
    \item[] Question: Does the paper describe potential risks incurred by study participants, whether such risks were disclosed to the subjects, and whether Institutional Review Board (IRB) approvals (or an equivalent approval/review based on the requirements of your country or institution) were obtained?
    \item[] Answer: \answerNA{} % Replace by \answerYes{}, \answerNo{}, or \answerNA{}.
    \item[] Justification: The paper does not involve crowdsourcing nor research with human subjects.

\item {\bf Declaration of LLM usage}
    \item[] Question: Does the paper describe the usage of LLMs if it is an important, original, or non-standard component of the core methods in this research? Note that if the LLM is used only for writing, editing, or formatting purposes and does \emph{not} impact the core methodology, scientific rigor, or originality of the research, declaration is not required.
    %this research? 
    \item[] Answer: \answerNA{} % Replace by \answerYes{}, \answerNo{}, or \answerNA{}.
    \item[] Justification: The core method development in this research does not involve LLMs as any important, original, or non-standard components.

\end{enumerate}